\documentclass[journal]{template/IEEEtran}

\usepackage{bmpsize}
\usepackage{amsthm}
\usepackage[linesnumbered,ruled,vlined]{algorithm2e}
\usepackage{amsmath,amssymb,enumerate}
\usepackage{mathrsfs}
\usepackage{times}
\usepackage{multicol}
\usepackage{amsfonts}
\usepackage{textcomp}
\usepackage{balance}
\usepackage{multirow}
\usepackage{booktabs}
\usepackage{dblfloatfix} 
\usepackage{bbold}
\usepackage{makecell}
\usepackage{hhline} 
\usepackage{float} 
\usepackage{xparse}
\usepackage{lipsum}
\usepackage{soul} 
\usepackage{times}
\usepackage{cite}
\usepackage{balance}
\usepackage{dsfont} 
\usepackage{physics} 
\usepackage{setspace} 
\usepackage{accents} 

\usepackage{mathabx} 

\usepackage[dvipdfmx]{graphicx} 
\usepackage[T1]{fontenc}
\usepackage[usenames,dvipsnames,svgnames,table, xcdraw]{xcolor}
\usepackage[utf8]{inputenc}

\usepackage{hyperref}
\hypersetup{
  colorlinks=true,pdfpagemode=UseNone,citecolor=black,linkcolor=black,urlcolor=BrickRed
}

\usepackage{caption}
\usepackage{cleveref}
\usepackage{subcaption}

\SetCommentSty{mycommfont}

\SetKwInput{KwInput}{Input}                
\SetKwInput{KwOutput}{Output}              

\graphicspath{
    {graphics/}
}

\newtheoremstyle{break}
  {\topsep}{\topsep}%
  {\itshape}{}%
  {\bfseries}{}%
  {\newline}{}%

\newtheorem{theorem}{Theorem}

\newtheorem{remark}{Remark}

\theoremstyle{break}

\definecolor{note}{rgb}{0.3,0.7,0.25}
\definecolor{rephase}{rgb}{0.15,0.7,0.15}
\definecolor{bag}{rgb}{0.5,0.5,0.0}

\newcommand{\lc}{LiDAR-camera~}

\newcommand{\lidar}{LiDAR~}

\newcommand{\lidarN}{LiDAR}
\newcommand{\lidars}{LiDARs~}

\newcommand{\velodyne}{\textit{32-Beam Velodyne ULTRA Puck LiDAR~}}
\newcommand{\velodyneN}{\textit{32-Beam Velodyne ULTRA Puck LiDAR}}

\newcommand{\dom}{\mathrm{dom}}

\newcommand{\squeezeup}{\vspace{-4mm}} 
\newcommand{\overbar}[1]{\mkern 1.5mu\overline{\mkern-1.5mu#1\mkern-1.5mu}\mkern 1.5mu}

\newcommand{\comment}[1]{}

\def\realnumbers{\mathbb{R}}
\def\real{\mathbb{R}}
\def\reals{\mathbb{R}}

\newcommand{\tp}{\mathcal{TP}}

\newcommand{\ep}{\mathcal{EP}}

\DeclareDocumentCommand{\RI}{ O{} O{} }{\mathcal{RI}_{#1}^{#2}}

\newcommand\inv[1]{#1\raisebox{1.15ex}{$\scriptscriptstyle-\!1$}}

\DeclareDocumentCommand{\td}{ O{} }{\tilde{#1}}
\newcommand{\transpose}{\mathsf{T}}

\newcommand{\SO}{\mathrm{SO}}
\newcommand{\se}{\mathfrak{se}}
\newcommand{\SE}{\mathrm{SE}}

\newcommand{\Exp}{\mathrm{Exp}}
\newcommand{\Log}{\mathrm{Log}}

\DeclareDocumentCommand{\asin}{ O{} }{\sin^{-1}(#1)}
\DeclareDocumentCommand{\acos}{ O{} }{\cos^{-1}(#1)}
\DeclareDocumentCommand{\atan}{ O{} }{\tan^{-1}(#1)}

\DeclareDocumentCommand{\vector}{ O{} }{\mathrm{vec}(#1)}
\DeclareDocumentCommand{\zeros}{ O{} }{\textbf{0}_{#1}}
\DeclareDocumentCommand{\pre}{ O{} O{} }{{}_{#1}^{#2}}

\DeclareMathOperator*{\argmin}{arg\,min}

\newcommand{\Lcal}{\mathcal{L}}
\newcommand{\Mcal}{\mathcal{M}}

\newcommand{\Rcal}{\mathcal{R}}

\newcommand{\Vcal}{\mathcal{V}}
\newcommand{\Xcal}{\mathcal{X}}

\DeclareDocumentCommand{\A}{ O{} O{} }{\textbf{A}_{#1}^{#2}}
\DeclareDocumentCommand{\H}{ O{} O{} }{\textbf{H}_{#1}^{#2}}
\DeclareDocumentCommand{\I}{ O{} O{} }{\textbf{I}_{#1}^{#2}}
\DeclareDocumentCommand{\L}{ O{} O{} }{\textbf{L}_{#1}^{#2}}
\DeclareDocumentCommand{\M}{ O{} O{} }{\textbf{M}_{#1}^{#2}}
\DeclareDocumentCommand{\N}{ O{} O{} }{\textbf{N}_{#1}^{#2}}
\DeclareDocumentCommand{\O}{ O{} O{} }{\textbf{O}_{#1}^{#2}}
\DeclareDocumentCommand{\P}{ O{} O{} }{\textbf{P}_{#1}^{#2}}
\DeclareDocumentCommand{\Q}{ O{} O{} }{\textbf{Q}_{#1}^{#2}}
\DeclareDocumentCommand{\R}{ O{} O{} }{\textbf{R}_{#1}^{#2}}
\DeclareDocumentCommand{\T}{ O{} O{} }{\textbf{T}_{#1}^{#2}}
\DeclareDocumentCommand{\U}{ O{} O{} }{\textbf{U}_{#1}^{#2}}
\DeclareDocumentCommand{\V}{ O{} O{} }{\textbf{V}_{#1}^{#2}}
\DeclareDocumentCommand{\X}{ O{} O{} }{\textbf{X}_{#1}^{#2}}
\DeclareDocumentCommand{\Y}{ O{} O{} }{\textbf{Y}_{#1}^{#2}}
\DeclareDocumentCommand{\Z}{ O{} O{} }{\textbf{Z}_{#1}^{#2}}

\DeclareDocumentCommand{\e}{ O{} O{} }{\textbf{e}_{#1}^{#2}}
\DeclareDocumentCommand{\n}{ O{} O{} }{\textbf{n}_{#1}^{#2}}
\DeclareDocumentCommand{\o}{ O{} O{} }{\textbf{o}_{#1}^{#2}}
\DeclareDocumentCommand{\t}{ O{} O{} }{\textbf{t}_{#1}^{#2}}
\DeclareDocumentCommand{\p}{ O{} O{} }{\textbf{p}_{#1}^{#2}}
\DeclareDocumentCommand{\q}{ O{} O{} }{\textbf{q}_{#1}^{#2}}
\DeclareDocumentCommand{\r}{ O{} O{} }{\textbf{r}_{#1}^{#2}}
\DeclareDocumentCommand{\u}{ O{} O{} }{\textbf{u}_{#1}^{#2}}
\DeclareDocumentCommand{\v}{ O{} O{} }{\textbf{v}_{#1}^{#2}}
\DeclareDocumentCommand{\x}{ O{} O{} }{\textbf{x}_{#1}^{#2}}

\title{Optimal Target Shape for LiDAR Pose Estimation}

\author{Jiunn-Kai Huang, William Clark, and Jessy W. Grizzle
\thanks{Jiunn-Kai Huang, William Clark, and J. Grizzle, are with the Robotics
Institute, University of Michigan, Ann Arbor, MI 48109, USA. \texttt{\{bjhuang, wiclark, grizzle\}@umich.edu}.} }

\begin{document}

\maketitle
\pagestyle{plain}

\begin{abstract}
    Targets are essential in problems such as object tracking in cluttered or
    textureless environments, camera (and multi-sensor) calibration tasks, and simultaneous localization and mapping (SLAM). Target shapes for
    these tasks typically are symmetric (square, rectangular, or circular) and work
    well for structured, dense sensor data such as pixel arrays (i.e., image).
    However, symmetric shapes lead to pose ambiguity when using sparse sensor data
    such as LiDAR point clouds and suffer from the quantization uncertainty of the
    LiDAR. This paper introduces the concept of optimizing target shape to remove pose ambiguity for LiDAR point clouds. A target is designed to induce large gradients at edge points under rotation and translation relative to the LiDAR to ameliorate the
    quantization uncertainty associated with point cloud sparseness. Moreover, given a target shape, we present a means that leverages the target's geometry to
    estimate the target's vertices while globally estimating the pose. Both the
    simulation and the experimental results (verified by a motion capture system)
    confirm that by using the optimal shape and the global solver, we
    achieve centimeter error in translation and a few degrees in rotation even
    when a partially illuminated target is placed 30 meters away. All the implementations and
    datasets are available at
    \href{https://github.com/UMich-BipedLab/global_pose_estimation_for_optimal_shape}{https://github.com/UMich-BipedLab/global\_pose\_estimation\_for\_optimal\_shape}.
\end{abstract}


%


\vspace{-2mm}
\section{Introduction}
\label{sec:OptimalIntro}
Targets have been widely employed as fiducial markers \cite{huang2021lidartag,
huang2020lidartaglonger, olson2011apriltag, wagner2003artoolkit,fiala2005artag,
degol2017chromatag, atcheson2010caltag, fiala2005comparing, wang2020lftag} and for
target-based sensor calibration \cite{huang2020intinsic, huang2020improvements,
liao2018extrinsic, zhou2018automatic, gong20133d, dhall2017lidar,
verma2019automatic, jiao2019novel, kim2019extrinsic, guindel2017automatic,
mishra2020extrinsic, xue2019automatic}. Fiducial markers
(artificial landmarks or targets) help robots estimate their pose by estimating
the target pose and are applied to simultaneous localization and mapping (SLAM)
systems for robot state estimation and loop closures. Additionally, it can 
facilitate human-robot interaction, allowing humans to give commands to a robot by
showing an appropriate marker. Extrinsic target-based calibration between sensors
(cameras, Light Detection and Ranging (LiDAR) sensors, Inertial Measurement Units
(IMU), etc) is crucial for modern autonomous navigation\cite{huang2021efficient}. Particularly in
target-based \lc calibration, one seeks to estimate a set of corresponding
features of a target (e.g., edge lines, normal vectors, vertices, or plane
equations) in the \lidarN's point cloud and the camera's image plane. 

\begin{figure}[!t]%
    \centering
    \begin{subfigure}{1\columnwidth}
        \centering
        \includegraphics[width=1\columnwidth, trim={0 0 0 0},clip]{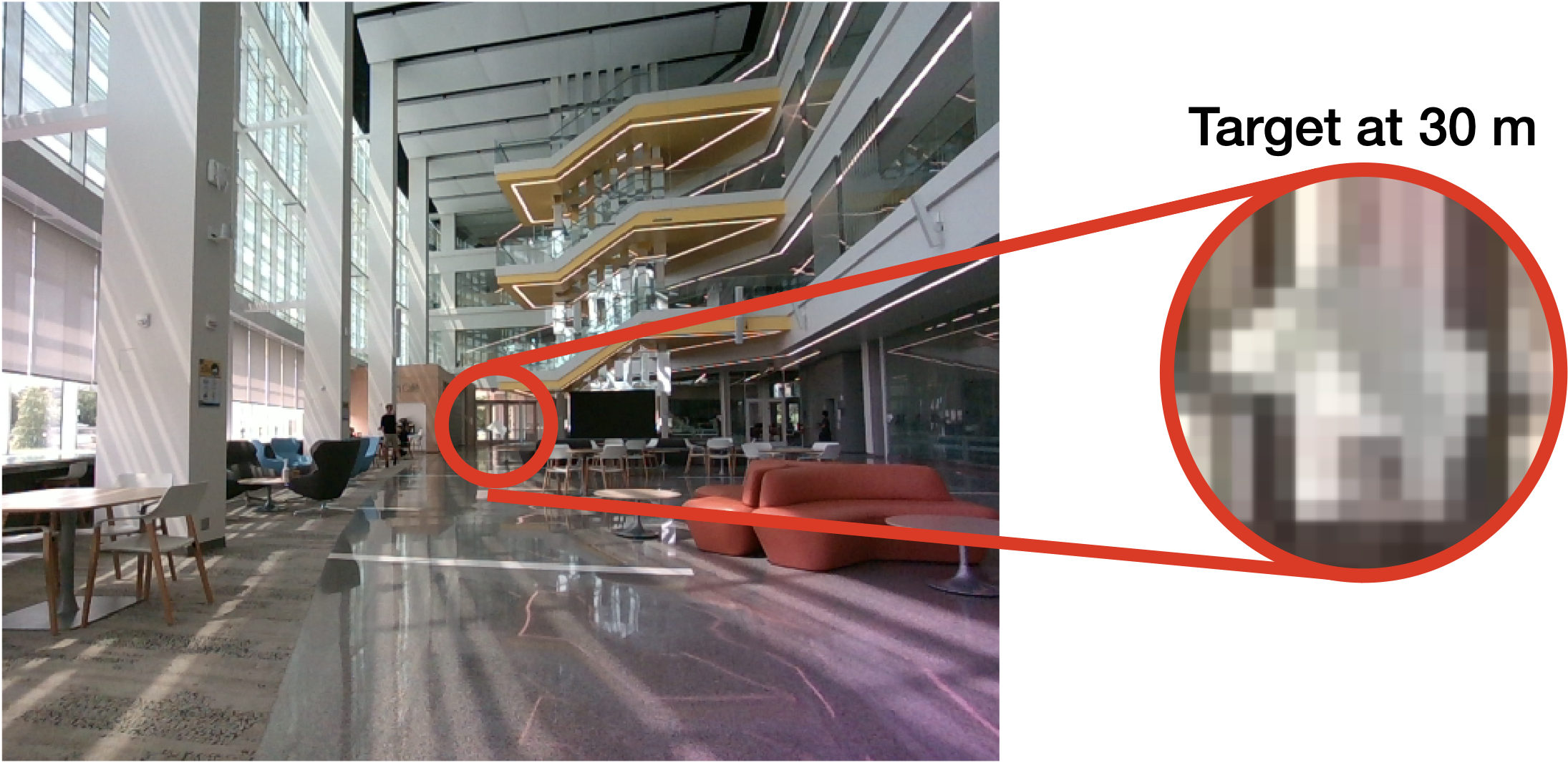}
    \end{subfigure}\vspace{15pt}
    \begin{subfigure}{0.55\columnwidth}
        \centering
        \includegraphics[height=0.8\columnwidth, trim={0 0 0 0},clip]{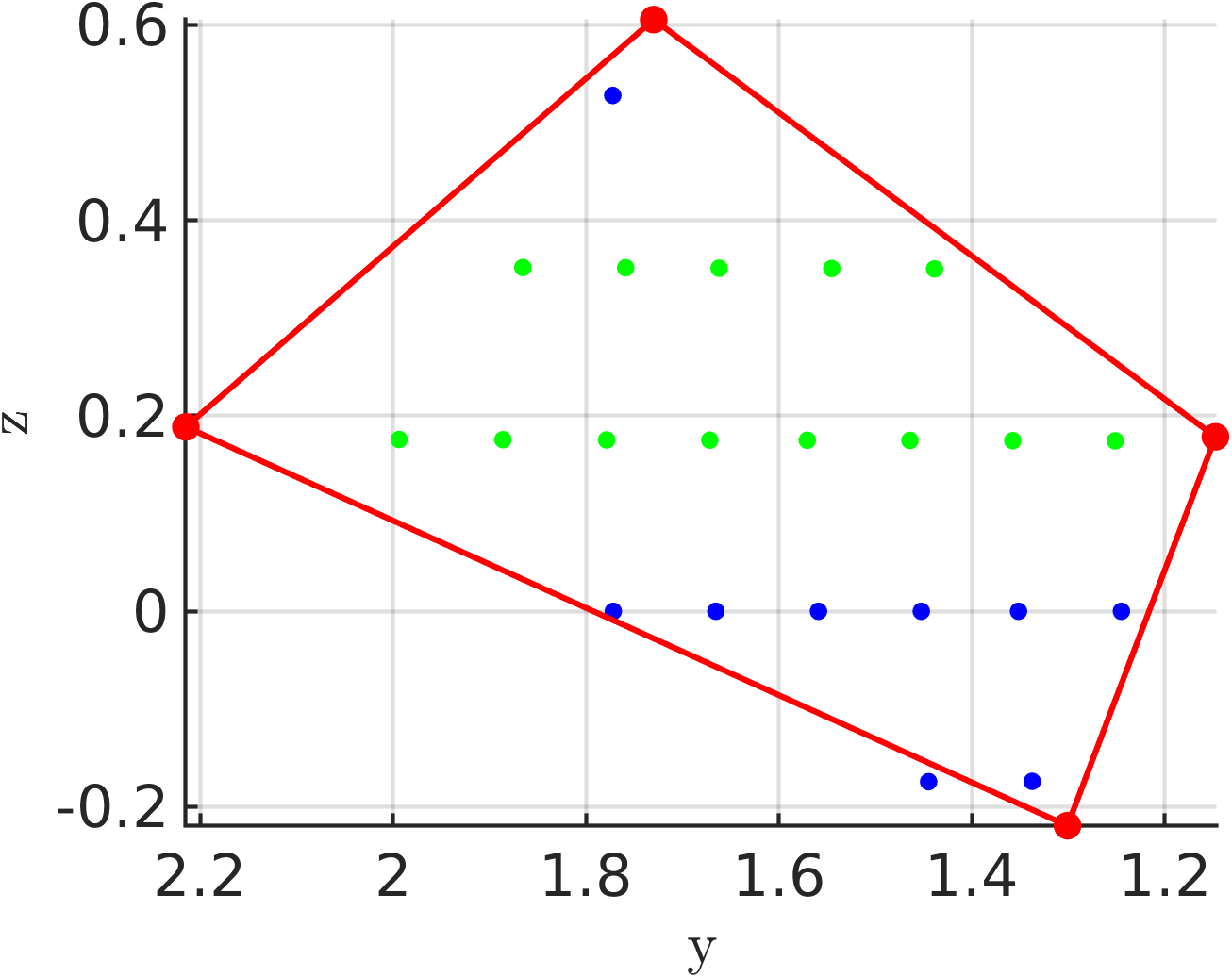}
    \end{subfigure}%
    \begin{subfigure}{0.48\columnwidth}
        \centering
        \includegraphics[height=0.95\columnwidth, trim={0 0 0 0},clip]{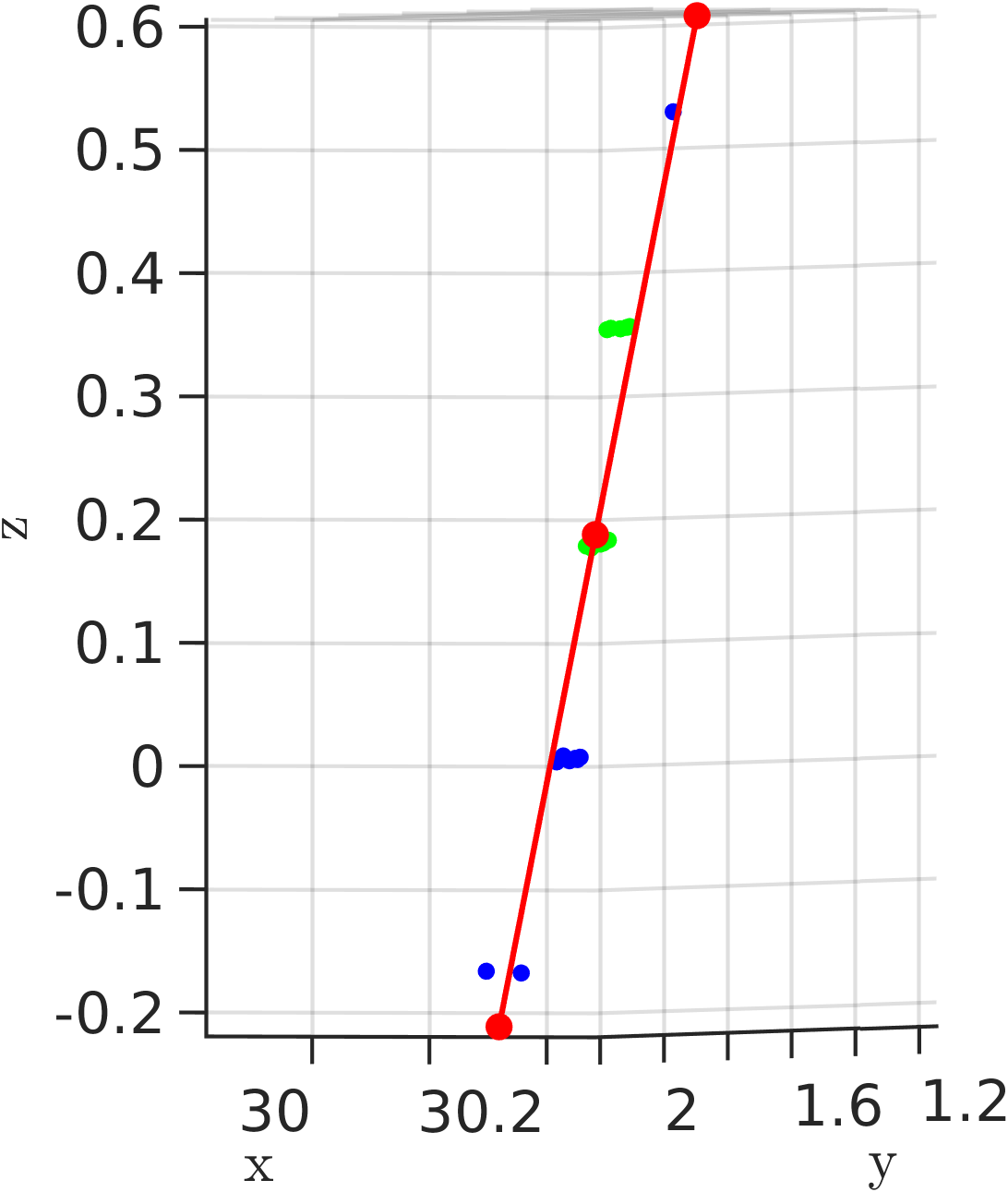}
    \end{subfigure}
    \caption[]{Illustration of the vertex and pose estimation using the proposed optimal
        target shape placed 30 meters away from a \lidar
        in the atrium of the Ford Robotics Building at the University of Michigan.
        The red frame is the proposed optimal shape that induces large gradients at
        edge points under translation and rotation. The pose and vertices of the target
        are jointly estimated by a global solver that uses known target geometry. The bottom two figures
        show the front view and a side view of the fitting results, respectively.
        The blue and the green dots are the \lidar returns on the target. Only the
        blue dots are used for pose estimation while to demonstrate robustness of the approach, the green dots are considered missing. If one were to apply \textit{RANSAC} --- a commonly used method --- to regress the target boundaries and subsequently estimate the vertices by line
        intersections, the method would fail due to the sparsity of inliers.}
    \label{fig:OptimalFirstImg}%
    \squeezeup
\end{figure}

The targets applied in these critical tasks are typically symmetric (square, diamond,
rectangle, or circle). A single symmetric target, such as a square\cite{huang2021lidartag,
olson2011apriltag}, leads to an ambiguous pose. This can be solved by adding an observable pattern to the target or by assuring that several asymmetrically-placed symmetric targets can be observed in a single scene. Furthermore,
estimating the pose or features of a target of injudicious design suffers from the
quantization uncertainty of a sensor, especially for \lidar sensors. A high-end
\lidarN, such as \velodyneN, still has roughly six centimeters of quantization error
at 10 meters, and 18 centimeters at 30 meters. The quantization uncertainty in the 
\lidar point cloud leads to rotation errors greater than 15 degrees for
targets farther away than 15 meters.

In this paper, we propose the concept of optimizing target shape to ameliorate
problems caused by quantization uncertainty and sparsity of the \lidar image of a
target. Specifically, we propose that a ``good target shape'' is one that possesses
large gradients at edge points when the target undergoes rotations or translations.
Moreover, we present a means that exploits target geometry to extract target vertices
while estimating pose. The pose estimation problem is formulated so that an existing
Semidefinite programming (SDP) global solver can be modified to globally and
efficiently compute the target's pose. Figure~\ref{fig:OptimalFirstImg} shows the
obtained pose estimation of a partially illuminated target placed 30 meters away and
having only nine returns (blue dots) from a \velodyneN, and three \lidar rings on the
target after the green dots are considered missing.

\vspace{-2mm}
\subsection{Contributions}
In particular, this work presents the following contributions: 
\begin{enumerate}
    \item We propose the concept of target shape optimization for estimating pose and vertices from \lidar
        point clouds. Specifically, we design a target so that its edge points induced by \lidar rings are ``highly'' sensitive to translation and rotation. This attenuates the effects of quantization uncertainty and sparsity of a target's
        \lidar image. The resulting shape is asymmetric to
        remove pose ambiguity.
    \item We present a means that uses target shape to jointly estimate target vertices
        and pose. Because the cost function of the proposed method can be formulated
        as an SDP, the target's pose and vertices can be globally estimated with an open-source solver \cite{briales2017convex}.
    \item We utilize an open-source \lidar simulator \cite{githubFileLiDARSimulator} to provide ground truth of the poses and
        vertices. In the simulation, we validate that the optimal shape with the
        global solver achieves centimeter error in translation and a few degrees of
        error in rotation when the targets are at a distance of 30 meters and
        partially illuminated. In addition, we conduct experimental evaluations where
        the ground truth data are provided by a motion capture system, and achieve results similar to the simulation. 
    \item We open-source all the related software for this work, including the
        generation of the optimal shape, our means for pose estimation, and the simulated/experimental datasets; see
        \cite{githubFileOptimalShapeGeneration, 
        githubFilePoseEstimationForOptimalShape}. 
\end{enumerate}

\section{Related Work}
\label{sec:OptimalRelatedWork}
To the best of our knowledge, there is no existing work on target shape design for \lidar point clouds. The closet publication on target shape design is \cite{muralikrishnan2017relative}, which evaluated the relative range error
 of dense terrestrial laser scanners using a plate, a sphere, and a novel dual-sphere-plate target. We therefore review instead some techniques to improve the pose estimation of fiducial markers and to assist in extracting features of calibration targets.  

\vspace{-2mm}
\subsection{Fiducial Markers}
\label{sec:RelatedFiducialMarkers}
Fiducial markers for cameras were originally developed and used for augmented reality applications
\cite{wagner2003artoolkit,fiala2005artag} and have been widely used for object
detection and tracking, and pose estimation \cite{klopschitz2007automatic}. Due to
their uniqueness and fast detection rate, they are also often used to improve
Simultaneous Localization And Mapping (SLAM) systems \cite{degol2018improved}. CCTag
\cite{calvet2016detection} adopts a set of rings (circular target) to enhance pose
estimation from blurry images. ChromaTag \cite{degol2017chromatag} proposes color
gradients on a squared target to speed up the detection process and obtain more
accurate pose estimation. More recently, LFTag \cite{wang2020lftag} has taken
advantage of topological markers, a kind of uncommon topological pattern, on a
squared target to improve pose estimation at a longer distance. However, all the
mentioned fiducial markers only work on cameras.

In our prior work on \lidar
\cite{huang2021lidartag}, we proposed the first fiducial marker for \lidar point
clouds, which can be perceived by both \lidars and cameras. 
We achieved millimeter error in translation and a few degrees in rotation.
However, due to the quantization error of the \lidarN, the performance of the
pose estimation (especially in-plane rotation) was noticeably degraded when the target was farther than 12 meters.
Thus, this work proposes the concept of target shape design to specifically
address the quantization uncertainty present in \lidar returns and push the range of pose estimation
to more than 30 meters. In passing, we note that symmetric targets, such as a square or hexagon, suffer from rotational ambiguity. Hence, our designed target will not be symmetric.

\vspace{-2mm}
\subsection{Target-Based \lc Calibration}
\label{sec:RelatedCalibration}
\lc calibration \cite{huang2020improvements, liao2018extrinsic, zhou2018automatic,
gong20133d, dhall2017lidar, verma2019automatic, jiao2019novel, kim2019extrinsic,
guindel2017automatic, mishra2020extrinsic, xue2019automatic} requires feature
correspondences from the image pixels and the \lidar point cloud. However, the
representations and inherent properties of camera images and \lidar point clouds are
distinct. An image (pixel arrays) is dense and very structured, with the pixels
arranged in a uniform (planar) grid, and each image has a fixed number of data
points. On the other hand, each scan of a \lidar returns a 3D point cloud consisting of a sparse set of
$(x,y,z)$ coordinates with associated intensities. In particular, \lidar
returns are not uniformly distributed in angle or distance
\cite[III-A]{huang2020lidartaglonger}. Target-based \lc calibration utilizes targets
to identify and estimate the corresponding features, such as vertices, 2D/3D edge
lines, normal vectors, or the plane equations of the targets. References
\cite{huang2020improvements, huang2020intinsic, liao2018extrinsic,
zhou2018automatic} have noted that placing the targets so that the rings of the
\lidar ran parallel to its edges led to ambiguity in the vertical position due to the
spacing of the rings and thus was detrimental to vertex or feature estimation.
References \cite{verma2019automatic, mishra2020extrinsic} utilize \textit{RANSAC}
\cite{fischler1981random} and plane fitting to remove the outliers of the \lidar
returns, while \cite{zhou2018automatic} proposes a ``denoising process'' for \lidar returns
around the target boundaries before applying \textit{RANSAC} to extract features.
When estimating the target vertices, the later references separate the edge points into groups and
then apply the \textit{RANSAC} algorithm. However, regressing the line equation of
edge points will fail when there are not enough edge points or inliers, as shown in
Fig.~\ref{fig:OptimalFirstImg}. Additionally, no target geometry information is used
while estimating the features.

The remainder of this paper is organized as follows.
Section~\ref{sec:ShapeSensitivity} formulates the design of an optimal target shape
 for \lidar point clouds. The extraction of the target vertices while
globally estimating the pose is discussed in Sec.~\ref{sec:GlobalPoseForOptimal}.
The simulation and experimental results are presented in
Sec.~\ref{sec:OptimalShapeSimulation} and Sec.~\ref{sec:OptimalShapeExperiment}.
Finally, Sec.~\ref{sec:OptimalShapeConclusion} concludes the paper and provides
suggestions for further work.

%

\section{Optimal Shape For Sparse LiDAR Point Clouds}
\label{sec:ShapeSensitivity}
\begin{figure}[t]
\centering
\includegraphics[width=1\columnwidth]{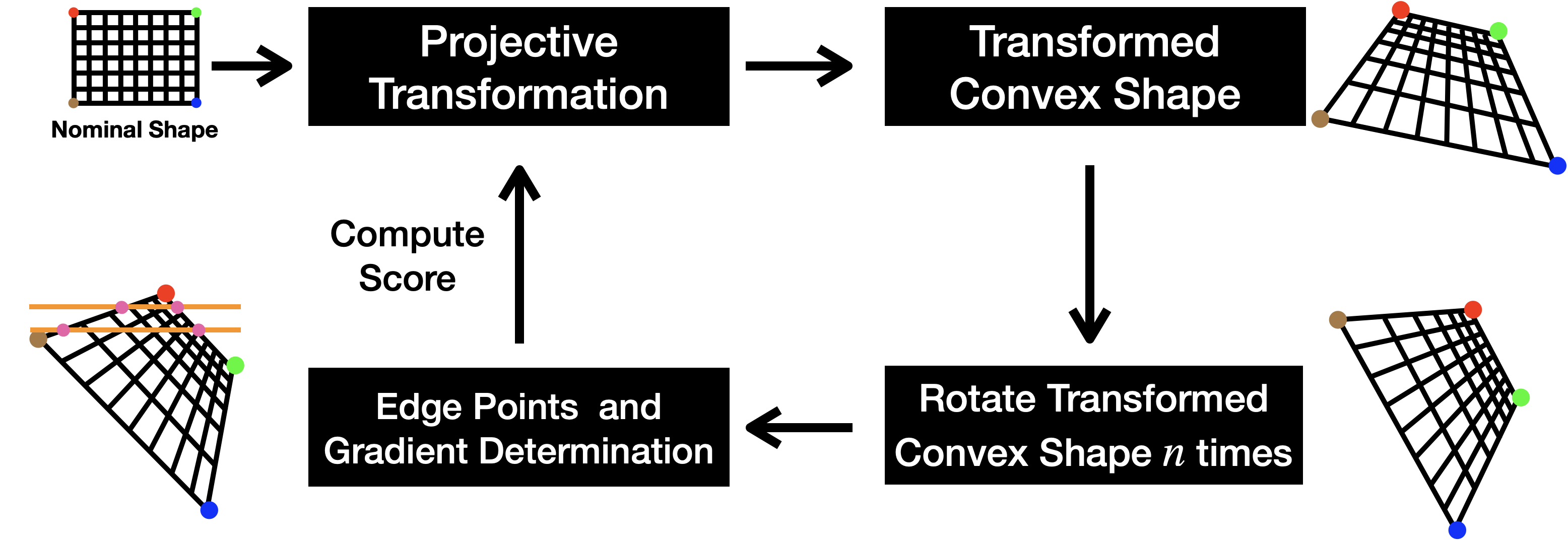}
\caption{Optimization process for determining an optimal target shape. Projective transformations
    applied to a nominal quadrilateral generate candidate convex quadrilaterals (targets). Edge points are intersections of \lidar rings with the target boundaries. The objective is to maximize the gradient of edge points under actions of $\SE(2)$ applied to the target. To enhance robustness, the gradients are computed for $n$-discrete rotations of the quadrilateral under partial illumination, and the score is the worst case.}
\label{fig:sensitivity-process}
\end{figure}

In this section, we propose a mathematical formulation of target shape design. The
main idea is for target translation and rotation to result in large gradients at edge
points defined by the \lidar returns. Fig.~\ref{fig:sensitivity-process} summarizes a
high-level optimization process.

\subsection{Convex Shape Generation}
We apply projective transformations on a nominal convex quadrilateral in 3D to
generate planar candidate targets. We will see that applying a projective transformation rather than working with a collection of vertices makes it easier to ensure convexity of the target and to generate a cost function that is invariant under scaling and translations.

Let $\Vcal := \{X_i| X_i := (x_i, y_i, 1)|x_i,
y_i\in\reals\}_{i=1}^4$ denote the 3D vertices $X_i$ of a nominal convex
quadrilateral, such as a square. Given $\P$, a projective transformation defined by a non-singular $3\times3$ matrix\cite[p.33]{hartley2000multiple}, let
$\widetilde{X}_i$ denote the new vertices transformed by $\P$: 
\begin{equation}
    \label{eq:projective}
    \widetilde{X}_i = 
    \begin{bmatrix}
        x_i^\prime\\y_i^\prime\\\lambda_i^\prime
    \end{bmatrix} = 
    \P X_i =
    \begin{bmatrix}
        p_{11}  & p_{12} & p_{13} \\
        p_{21}  & p_{22} & p_{23} \\
        p_{31}  & p_{32} & \upsilon 
    \end{bmatrix}
    \begin{bmatrix}
        x_i\\y_i\\1 
    \end{bmatrix}.
\end{equation}
The resulting vertices $\widetilde{\Vcal}:=\{\widetilde{X}_i\}_{i=1}^4$
lie in the projective space $\mathbb{P}^2$ \cite[p.26]{hartley2000multiple}. Let
$\Vcal^\prime$ be the corresponding transformed vertices in the Cartesian space
$(\reals^2)$ \cite[p.27]{hartley2000multiple} and $\Pi:\mathbb{P}^2\rightarrow
\reals^2$ be the mapping function 
\begin{equation}
\label{eq:CandidateTargetFamily}
  \Pi(\widetilde{\Vcal}):=\Pi(\P(\Vcal))):=\left\{X_i^\prime\bigg|X_i^\prime :=
\left(\frac{x_i^\prime}{\lambda_i^\prime}, \frac{y_i^\prime}{\lambda_i^\prime}\right)\right\}_{i=1}^4. 
 \end{equation}
To summarize, given a nominal convex quadrilateral, $\Vcal$, and a projective transformation, $\P$, we construct a new quadrilateral via
\begin{equation}
    \mathcal{V} \longmapsto \P\mathcal{V}=:\widetilde{\Vcal} \text{ and } \Vcal^\prime:= \Pi(\widetilde{\Vcal}).
\end{equation}

\begin{remark}
From here on, we will abuse notation
and pass from Cartesian coordinates to homogeneous coordinates without noting the
distinction. 
\end{remark}
It is important to note that for any desired quadrilateral, there exists a projective transformation yielding $\Vcal^\prime$ from $\Vcal$. Hence, our procedure for generating candidate targets is without loss of generality.

\begin{figure}[t]%
    \centering
    \begin{subfigure}{0.48\columnwidth}
        \centering
        \includegraphics[width=1\columnwidth, trim={0 0 0 0},clip]{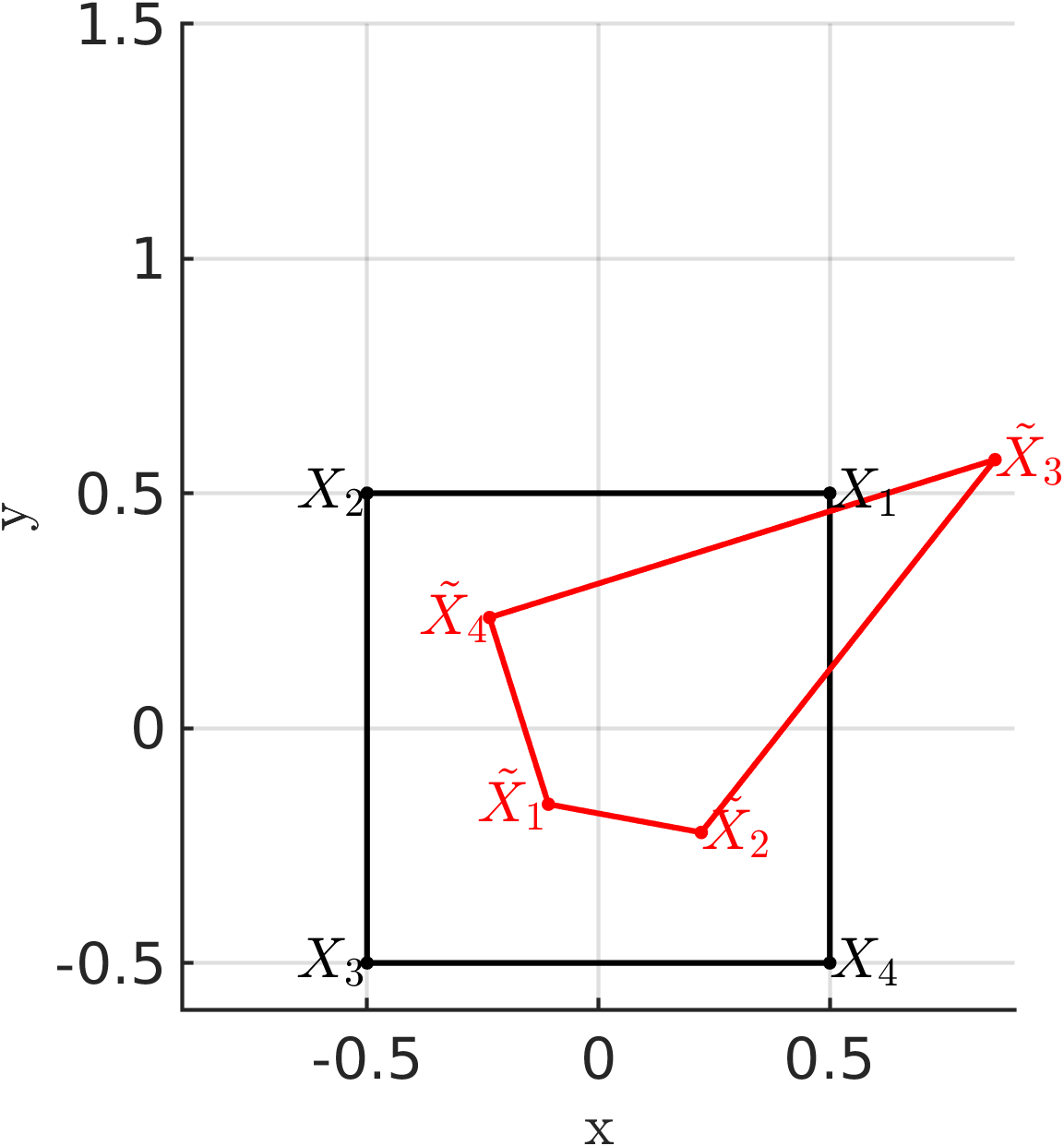}
        \caption{}
        \label{fig:convex}%
    \end{subfigure}%
    \begin{subfigure}{0.48\columnwidth}
        \centering
        \includegraphics[width=1\columnwidth, trim={0 0 0 0},clip]{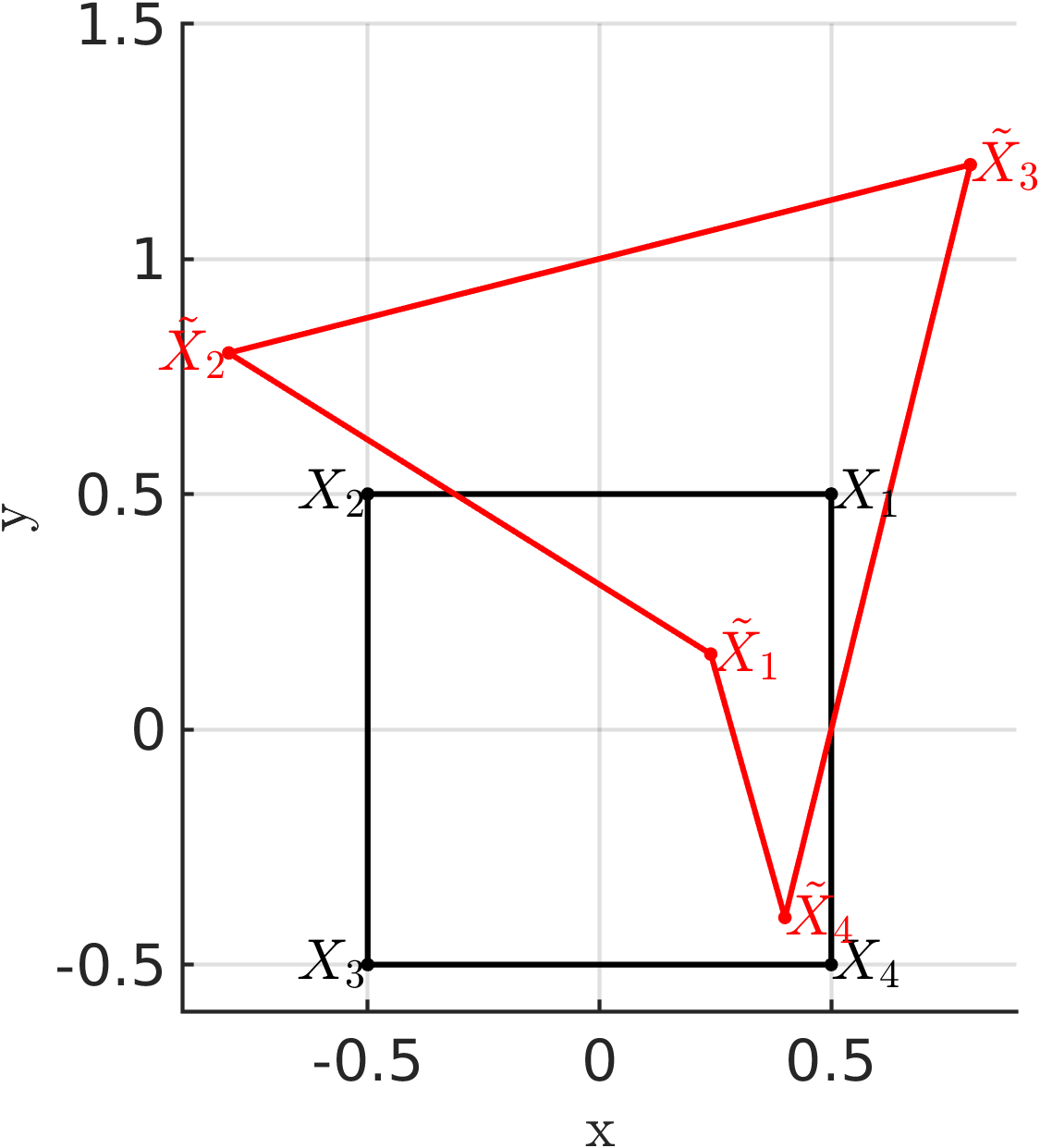}
        \caption{}
        \label{fig:non-convex}%
    \end{subfigure}%
    \caption[]{Equation \eqref{eq:convexityConstraints} versus convexity. 
        The red indicates the resulting shapes transformed by projective transformations
        from the same nominal shape (black). The left shows a case where
        \eqref{eq:convexityConstraints} is satisfied and thus the transformed shape is
        convex; otherwise, it is non-convex, as shown on the right, where 
        $ p_{31}x_3 + p_{32}y_3 + \upsilon =-0.5$.}%
    \label{fig:convexity-comparison}%
\end{figure}

\begin{theorem}[\hspace{1sp}{\cite[p.274]{cederberg2013course}}]
    \label{thm:uniquessOfProjective}
    There exists a unique projective transformation that maps any four points, no
    three of which are collinear, to any four points, no three of which are collinear.
\end{theorem}
While Theorem~\ref{thm:uniquessOfProjective} can be used to construct an arbitrary quadrilateral, convexity need not be conserved as shown in Fig.~\ref{fig:convexity-comparison}. We need the following condition to ensure that the resulting polygon is convex:
\begin{theorem}[\hspace{1sp}{\cite[p.39]{boyd2004convex}}]
    \label{thm:convexity}
    Let $\Omega:\realnumbers^{3}\rightarrow\realnumbers^2$ by  $\Omega(x,y,\lambda) = (x/\lambda,y/\lambda)$. If $\dom \Omega =
    \realnumbers^2 \times \realnumbers_{++},$ where $\realnumbers_{++} =
    \{x\in\realnumbers|x>0\}$ and the set $C\subseteq\dom\Omega$ is convex, then its image 
    \begin{equation}
        \Omega(C) = \{\Omega(x)|x\in C\}
    \end{equation}
    is also convex.
\end{theorem} 
For the domain of our projective transformation to be $\realnumbers^2 \times \realnumbers_{++}$, and hence for the candidate target to be automatically convex, the following linear inequality should be imposed in \eqref{eq:projective},
\begin{equation}
    \label{eq:convexityConstraints}
    p_{31}x_i + p_{32}y_i + \upsilon > 0.
\end{equation}

Because we consider two candidate targets to be equivalent if one can be obtained from the other by translation and
scaling, we are led to decompose the projective transformation as follows\cite{hartley2003multiple,
decomposition},
\begin{align}
    \P &= \H[S]\H[SH]\H[SC]\H[E] \nonumber\\
    \label{eq:decomposition}
       &=
\begin{bmatrix}s\mathbf{R} & \mathbf{t} \\ \mathbf{0}^\top & \mathbf{1}\end{bmatrix}
 \begin{bmatrix} 1 & k & 0 \\ 0 & 1 & 0 \\ 0 & 0 & 1\end{bmatrix}
 \begin{bmatrix} \lambda & 0 & 0\\ 0 & 1/\lambda & 0 \\ 0 & 0 & 1\end{bmatrix}
 \begin{bmatrix}\mathbf{I} & \mathbf{0} \\ \mathbf{v}^\top & \mathbf{\upsilon}\end{bmatrix},
\end{align}
where $\H[S], \H[SH], \H[SC], \H[E]$ are similarity, shearing, scaling and elation
transformations, respectively; see\cite{hartley2003multiple} for more details. By
setting $s=1, \t =(0,0)$ in \eqref{eq:decomposition}, the degree of freedom (DoF) of the projective transformation drops from eight to five. Our family of candidate targets is now given by \eqref{eq:CandidateTargetFamily} with $\P$
satisfying \eqref{eq:convexityConstraints} and \eqref{eq:decomposition}.

In summary, we can describe candidate convex target shapes via projective transformations while reducing the number of degrees of freedom to five.

\begin{remark}
  The same design process could be run with an $N$-gon, for $N \ge 3$. If $N$ is too large, the target will have at least one very short edge which will be impossible to discern in a point cloud. In between, it's a tradeoff between having adequate area to collect \lidar returns, non-parallel edges to minimize pose ambiguity, and long enough edges to pick them out of a point cloud. We used a $4$-gon as a reasonable starting point. Investigating $N$ equals 3 and 5 would be interesting as well.
\end{remark}

\subsection{Edge Points Determination}
\label{sec:EdgePointsDetermination}
\begin{figure}[t]
\centering
\includegraphics[width=0.85\columnwidth]{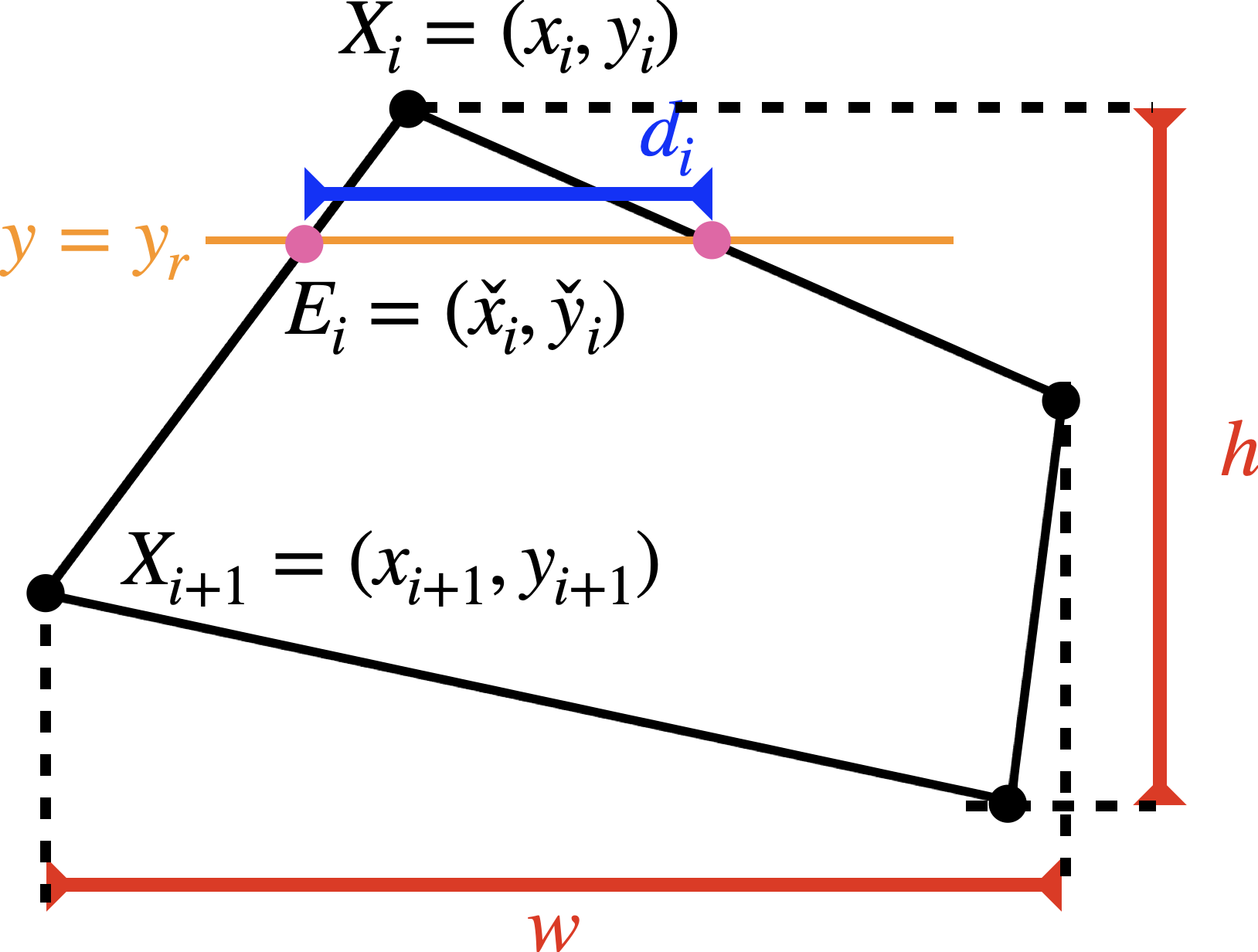}
\caption{The pink dots are the edge points determined by
    a \lidar ring (orange line) intersecting with the edge line connecting two vertices
($X_i$ and $X_{i+1}$). The distance between the two edge points on the same \lidar ring is
$d_i$.
The height and width of the shape is $w$ and $h$, respectively.
}
\label{fig:edge-points}
\end{figure}

As mentioned in \eqref{eq:CandidateTargetFamily}, $\Vcal^\prime$ are the 2D vertices of a candidate target. 
An edge point $E_i$ is defined by the intersection point of a \lidar ring and the
line connecting two vertices of the polygon as shown in Fig.~\ref{fig:edge-points}.
If we let $\mathcal{S}$ be the boundary of the quadrilateral with vertices $\mathcal{V}'$, the collection of edge points detected by the \lidar is the set $\ep:=\{E_i\}_{i=1}^M$, where $M$ is the number of edge points and is given by the intersection of $\mathcal{S}$ with the \lidar rings $\Lcal{\Rcal}$, i.e.
\begin{equation}
    \mathcal{S} = \partial\mathrm{conv}(\mathcal{V}'), \quad
    \left\{ E_i\right\}_{i=1}^M = \mathcal{S}\cap \Lcal{\Rcal}.
\end{equation}
When the \lidar rings are horizontal $(y=y_r)$,
an edge point $E_i = (\widecheck{x}_i, \widecheck{y}_i)$ can always be computed in closed form,
\begin{equation}
\label{eq:edge-points} 
\widecheck{x}_i = x_i + \frac{x_{i+1}-x_i}{y_{i+1}-y_i}(y_r-y_i)\text{~~and~~} \widecheck{y}_i=y_r.
\end{equation}

\subsection{Shape Sensitivity}
\label{sec:ShapeSensitivityComputation}
\begin{figure}[t]
\centering
\includegraphics[width=1\columnwidth]{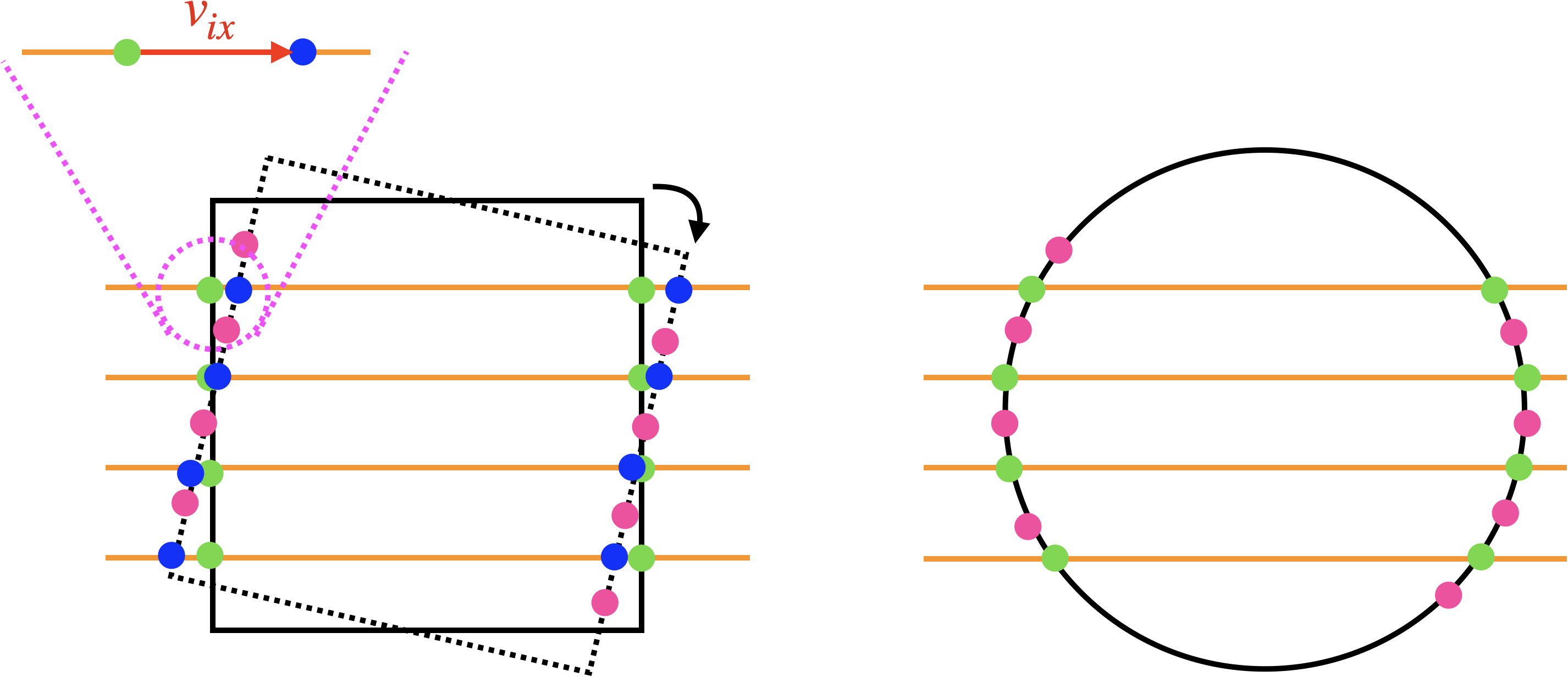}
\caption{Shape sensitivity under rotation. The sensitivity of a shape is defined 
    by the gradient ($v_{ix}$) of each edge point as it moves along the ring lines (orange) under rotations and translations of the shape.  
    The green dots are the original edge points and the pink dots are the edge
    points on the rotated shape. The same process can be done for translation.
    The left shows a gradient $v_{ix}$
of a square where the dotted square has been rotated from the nominal positioned square. The right
shows that the gradient of an edge point on a circle is zero under rotation about its center. The gradient would be non-zero for translations.
}
\label{fig:sensitivity-comparison}

\end{figure}

From experience gained in LiDARTag~[1], we observed that the pose estimation suffers the most from in-plane
    rotation. Therefore, we compute the shape sensitivity in $\SE(2)$.
The sensitivity of a polygon is defined as the gradient of the edge points with respect to rigid-body transformations of the polygon, with the \lidar rings held constant, as
shown in Fig.~\ref{fig:sensitivity-comparison}. Hence, the sensitivity captures
the horizontal movement of an edge point after the shape is rotated and translated. 

For a transformation in the Special Euclidean group $\H\in \SE(2)$, let $E_i^\prime$, denote the transformed edge point by
\begin{equation}
    E_i^\prime := \H\circ E_i =
   \begin{bmatrix}
       \cos\theta &-\sin\theta &t_x\\
       \sin\theta & \cos\theta &t_y\\
       0 & 0 & 1
   \end{bmatrix} 
   \begin{bmatrix}
       \widecheck{x}_i\\\widecheck{y}_i\\1 
  \end{bmatrix},
  \label{eq:SE2action}
\end{equation}
where $\theta, t_x, t_y$ are the rotation angle, the translation on x-axis, and the
translation on y-axis, respectively. Denote $\Exp(\kappa,\omega, u, v)$ as the
exponential map that maps from the Lie algebra $\mathfrak{se}(2)$ to the continuous
Lie group $\SE(2)$, 
\begin{align}
    \label{eq:lie-algebra}
    &\Exp(\kappa,\omega,u,v)=\mathrm{expm}\left(\kappa \begin{bmatrix}
            0 & -\omega & u\\
            \omega & 0 & v\\
            0&0&0
    \end{bmatrix}\right) \nonumber\\ 
    &= 
    \begin{bmatrix}
        \cos(\omega \kappa) &  -\sin(\omega \kappa) & \frac{1}{\omega}(v\cos(\omega \kappa)+u\sin(\omega \kappa)-v)\\
        \sin(\omega \kappa) &   \cos(\omega \kappa) & \frac{1}{\omega}(v\sin(\omega \kappa)-u\cos(\omega \kappa)+u)\\
        0&0&1
    \end{bmatrix},
\end{align}
where $(\omega, u, v)$ parameterize the unit sphere $S^2\subset\mathbb{R}^3$, $\mathrm{expm}$ is the usual matrix exponential, and $\kappa$ is a dummy variable
for differentiation. Comparing $\H$ in \eqref{eq:SE2action} and
\eqref{eq:lie-algebra} leads to
\begin{equation}
    \begin{cases}
        \label{eq:coefficients}
        \theta = \omega \kappa\\
        t_x = \frac{1}{\omega}(v\cos(\omega \kappa)+u\sin(\omega \kappa)-v)\\
        t_y = \frac{1}{\omega}(v\sin(\omega \kappa)-u\cos(\omega \kappa)+u).
    \end{cases}
\end{equation}

For each triple of values $(\omega, u, v)$, the action of \eqref{eq:lie-algebra} on a candidate target quadrilateral results a path $p_i(\kappa):=p_i(\Exp(\kappa\omega, \kappa u,\kappa v))$ being traced out by the edge points along a \lidar ring. Using  
 \eqref{eq:coefficients} to differentiate the path
$p_i$ at the identity of $SE(2)$
produces an action of
$\mathfrak{se}_2$,
\begin{equation}
    \label{eq:differentiated-path}
        v_i(\omega,u,v) := \left.\frac{d}{dt}\right\vert_{\kappa=0}p_i(\Exp(\kappa\omega, \kappa u,
        \kappa v)), ~(\omega, u, v)\in \mathfrak{se_2}.
\end{equation}
From \eqref{eq:edge-points}, \eqref{eq:lie-algebra}, \eqref{eq:coefficients}, and \eqref{eq:differentiated-path}, the gradient of the edge point with respect to the
\lidar ring is
\begin{equation}
\label{eq:gradient_x}
       \begin{aligned}
           v_{i_x} = &\omega\left( \frac{(x_i-x_{i+1})(x_iy_{i+1}-y_ix_{i+1}+x_{i+1}y_r-x_iy_r)}{(y_i-y_{i+1})^2}-y_r\right)\\
                    &~+ u + \left(\frac{x_{i+1}-x_i}{y_{i+1}-y_i}\right)v.
       \end{aligned}
\end{equation}
Notice that $v_{i_y} = 0$ because the $y$-coordinates of $\H(\Vcal^\prime)\cap \Lcal{\Rcal}$ remain fixed.

Finally, we define the sensitivity $\Mcal$ of the polygon
\begin{equation}
    \label{eq:sensitivity-cost}
       \Mcal(\Vcal, \Lcal{\Rcal}, \omega, u, v) :=\frac{1}{h} \sum_{i=1}^M v_{i_x}^2,
\end{equation}
where $M$ is the number of edge points, and $h$, defined in Fig.~\ref{fig:edge-points} is included because gradients in \eqref{eq:gradient_x} scale with vertical height.

\subsection{Initial Cost Function}
\label{sec:InitialCost}
The candidate target's sensitivity defined in \eqref{eq:sensitivity-cost} does not take into account the discrete nature of the \lidar returns on a given ring. Let $d_i$ denote the distance between two edge points on the $i$-th \lidar ring. We want to encourage targets that have $d_i$ larger than the spatial quantization of the \lidar returns. We can do this in two ways, by scaling \eqref{eq:sensitivity-cost} by $w$ and by including a term in the cost of the form $\sum_{i=1}^K d_i$ ($d_i$ is larger for wider targets). The resulting score of shape ($\Psi$)
becomes 
\begin{equation}
    \label{eq:final-cost}
        \Psi = w \Mcal  + \mu\sum_{i=1}^K d_i,
\end{equation}
where $w$ is the width of the polygon, $K$ is the number of rings illuminating the polygon, and $\mu$ is a weight trading off the two terms in the cost function.


\vspace{-2mm}
\section{Robust Shape for Real LiDAR Sensors}
In previous section (Sec.~\ref{sec:InitialCost}), we added an extra term to
\eqref{eq:final-cost} to account for the discrete measurements in azimuth direction of
the \lidarN.
Additionally, \lidars have different ring densities at different elevation angles.
For example, \velodyne has dense ring density between $-5^\circ$ and $3^\circ$, and
has sparse ring density from $-25^\circ$ to $-5^\circ$ and from $3^\circ$ to
$15^\circ$ \cite{velodyneUltraPuck}. A target could be partially illuminated in the
sparse region, as shown in Fig.~\ref{fig:partially-illuminated}. Therefore, assuming
edge points are uniformly distributed is not practical and using a distribution of edge
points that is similar to reality is critical while maximizing
\eqref{eq:final-cost}. Additionally, \lidar rings from mobile robots are not always parallel to
the ground plane. We account for non-horizontal \lidar
rings by rotating the candidate target.

\vspace{-2mm}
\subsection{Partial Illumination of Target}
\label{sec:OptimalOcclusion}
\begin{figure}[!t]%
    \centering
    \begin{subfigure}{0.49\columnwidth}
        \centering
        \includegraphics[height=0.65\columnwidth, trim={0cm 0cm 0cm 0.cm},clip]{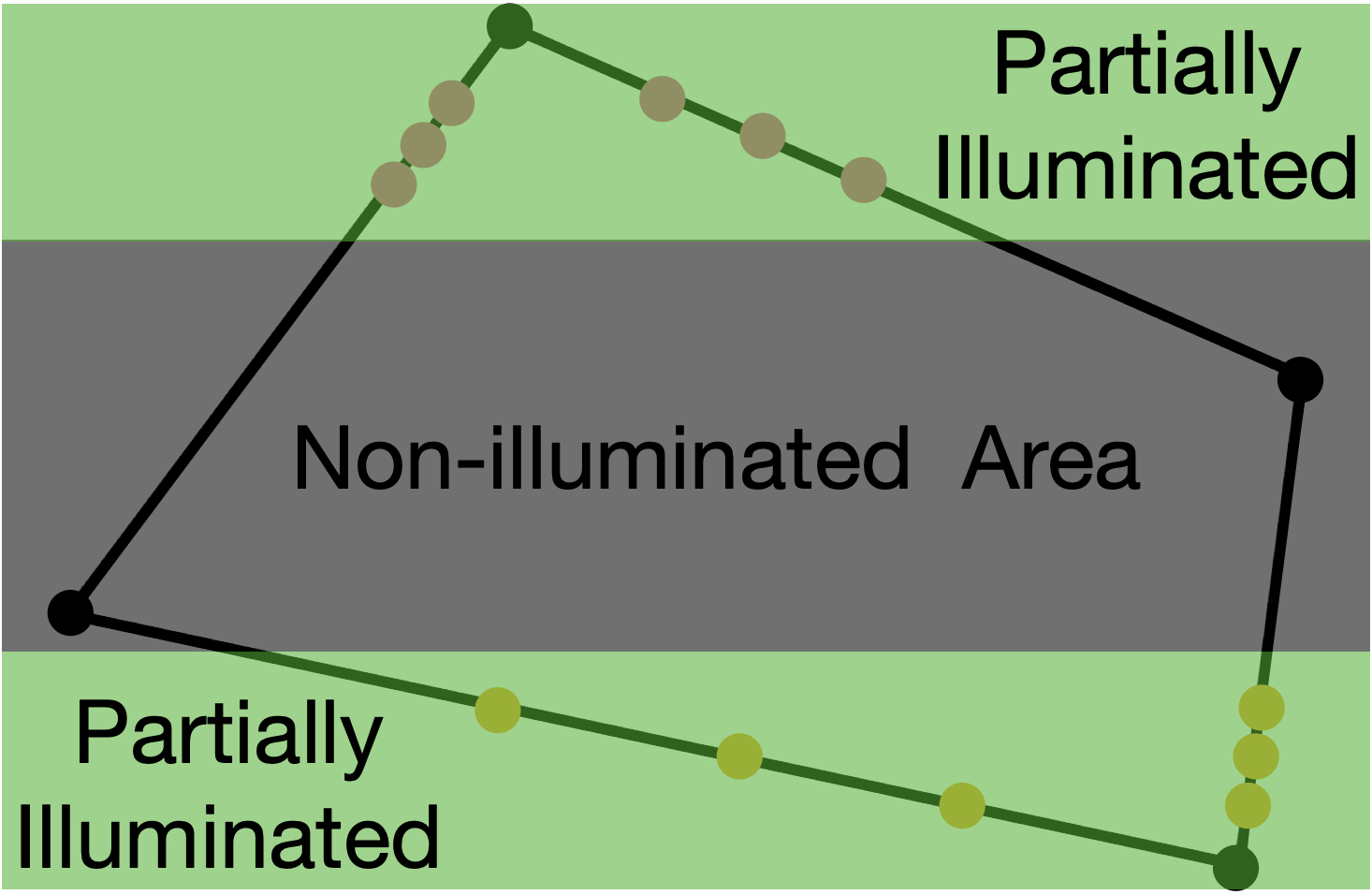}
    \caption[]{}%
\label{fig:partially-illuminated}
    \end{subfigure}%
    \begin{subfigure}{0.49\columnwidth}
        \centering
        \includegraphics[height=0.65\columnwidth, trim={0cm 0cm 0cm 0.cm},clip]{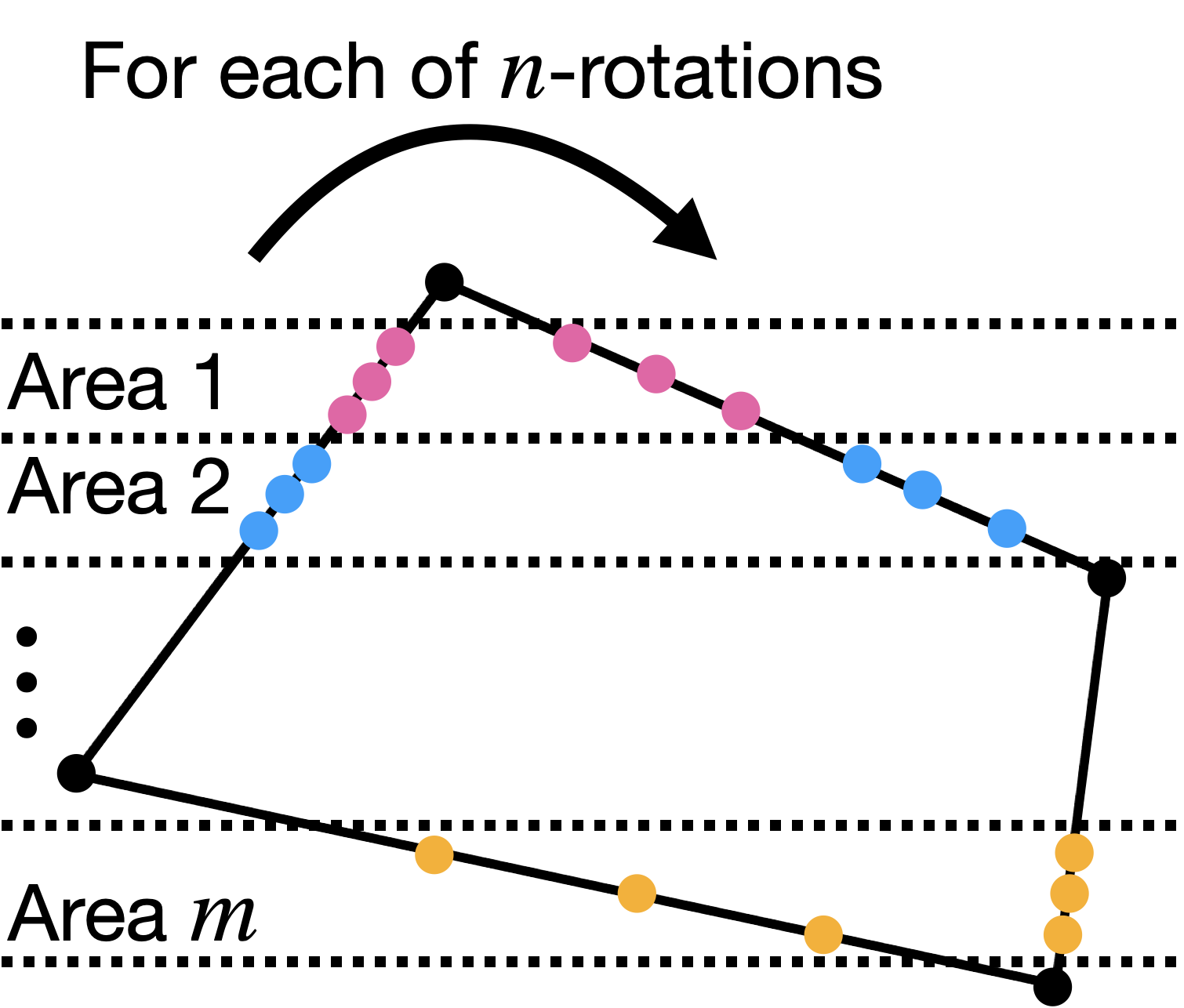}
    \caption[]{}%
\label{fig:partially-illuminated-edge-points}
    \end{subfigure}%
    \caption[]{The left shows a partially illuminated candidate shape. Because we also rotate the target when computing a score, we can without loss of generality use horizontal strips to partially occlude the target. The right shows
that the target is divided into $m$-areas of partial illumination, and that for each of $n$-rotations of the candidate target, a score is assigned to each subarea based on \eqref{eq:final-cost}. The final score of
the shape is the lowest among the $n\times m$ scores.
}%
\label{fig:stepsOfCostDecomposition}%
\squeezeup
\end{figure}

To have the shape being robust to illuminated area and the angle of the rings with respect to the target, we
first rotate the generated polygon $n$ times, and then divide the rotated polygon
into $m$ areas. Only one area is illuminated by \lidar rings at a time to determine
edge points and to compute the score \eqref{eq:final-cost}, $\Psi_{ij}$, for $1 \le i \le n$ and $1 \le j \le m$.
Figure~\ref{fig:partially-illuminated-edge-points} shows the edge points being determined for a partially illuminated target. Equation \eqref{eq:final-cost} is consequently evaluated
$n\times m$ times for illumination of the target and the lowest among the $n\times m$ scores is the final score of the candidate target shape. 

\vspace{-2mm}
\subsection{Optimization for the Optimal Shape}
To summarize, the resulting optimization problem depends on the projective
transformation parameters that are used to generate a convex polygon, edge points
illuminated  by horizontal \lidar rings lied on the rotated quadrilateral, the
transformation of the edge points in $\se_2$, and distances between two edge points
on the corresponding \lidar rings. Thus, the optimization problem is defined as:
\begin{equation}
    \label{eq:OptimalShapeOptimization}
        \P[][*] = \argmin_{\P}\min_{\omega, u, v}\max_{i, j}\{-\Psi_{ij}\}.
\end{equation}

\begin{figure}[t]
\centering
\includegraphics[width=1\columnwidth, trim={0cm 0cm 0cm 0.cm},clip]{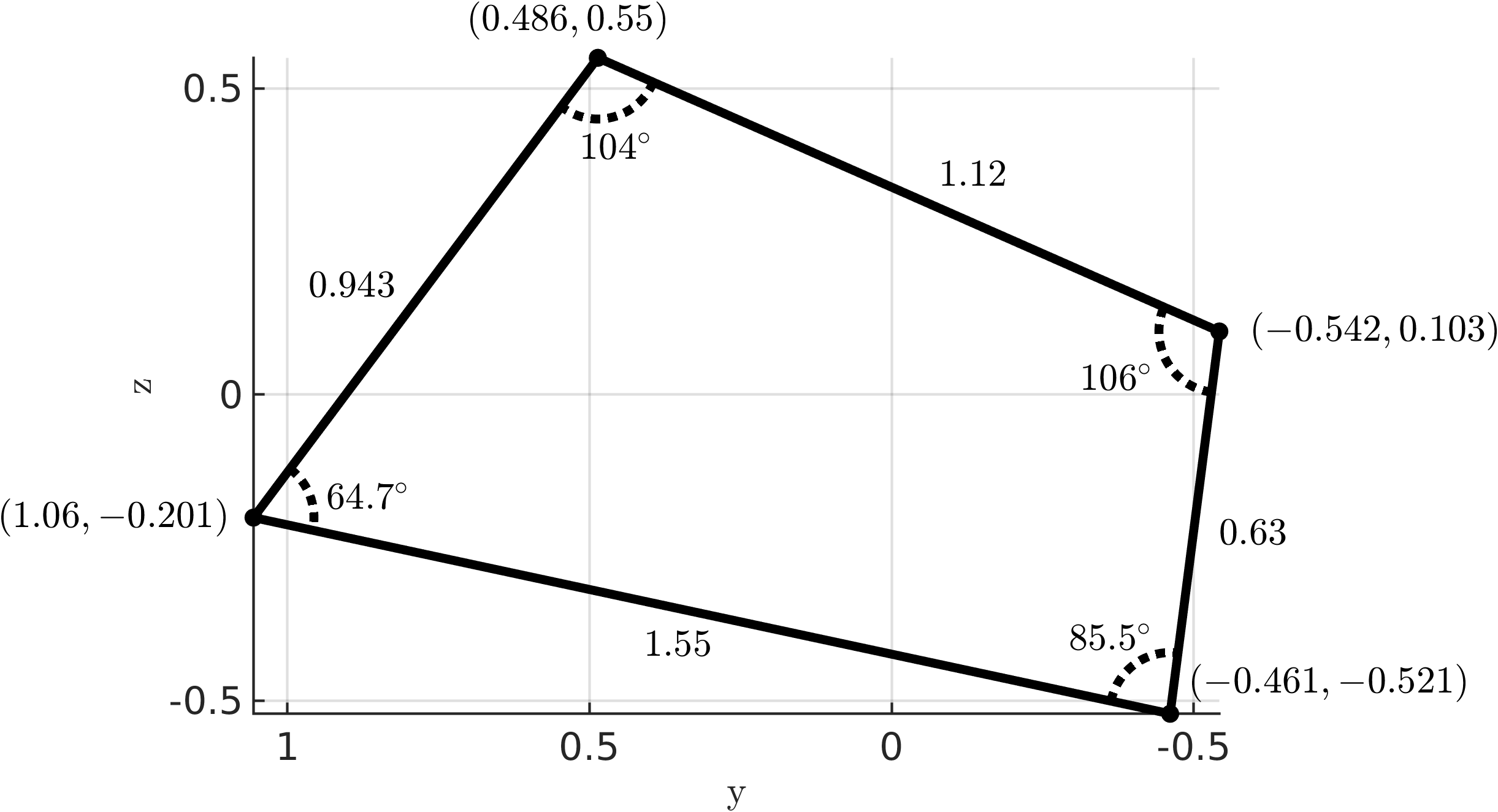}
\caption{The resulting optimal shape from \eqref{eq:OptimalShapeOptimization} in arbitrary units.
} \label{fig:optimal-shape}
\squeezeup
\end{figure}

The optimization problem \eqref{eq:OptimalShapeOptimization} was (locally) solved by
\texttt{fmincon} in MATLAB, after the optimization parameters were randomly
initialized. We rotated the generated polygon six times. Each rotated polygon was
divided into five areas, and four \lidar rings were used to illuminate one area at a
time. The unit sphere of unit vectors in $\se(2)$, mentioned in
Sec.~\ref{sec:ShapeSensitivityComputation}, was discretized into $25\times25$ faces (by normalizing the vectors to have unit length, we can reduce the dimension from three to two).
Once the generated polygon was rotated and illuminated, the sensitivity of the
resulting edges points were evaluated at each face on the unit sphere. The resulting
optimal shape is shown in Fig.~\ref{fig:optimal-shape}. One can observe that the resulting shape satisfies:
      \textbf{1)} it has sufficient area so as to collect \lidar returns; \textbf{2)} the length of the shortest side is still long enough to be identified through edge points; \textbf{3)} its asymmetric shape avoids the issue of pose ambiguity.

\begin{remark}
   The main point of this paper is that target shape can be used to enhance the estimation of target vertices and relative pose between a target and a \lidarN. We have proposed one algorithmic means to produce an ``optimal target shape''. 
    Different notions of cost will result in different shapes. 
\end{remark}

\vspace{-2mm}
\section{Global Pose and Feature Estimation} 
\label{sec:GlobalPoseForOptimal}
\begin{figure}[t]
\centering
\includegraphics[width=1\columnwidth, trim={0cm 0cm 0cm 0.cm},clip]{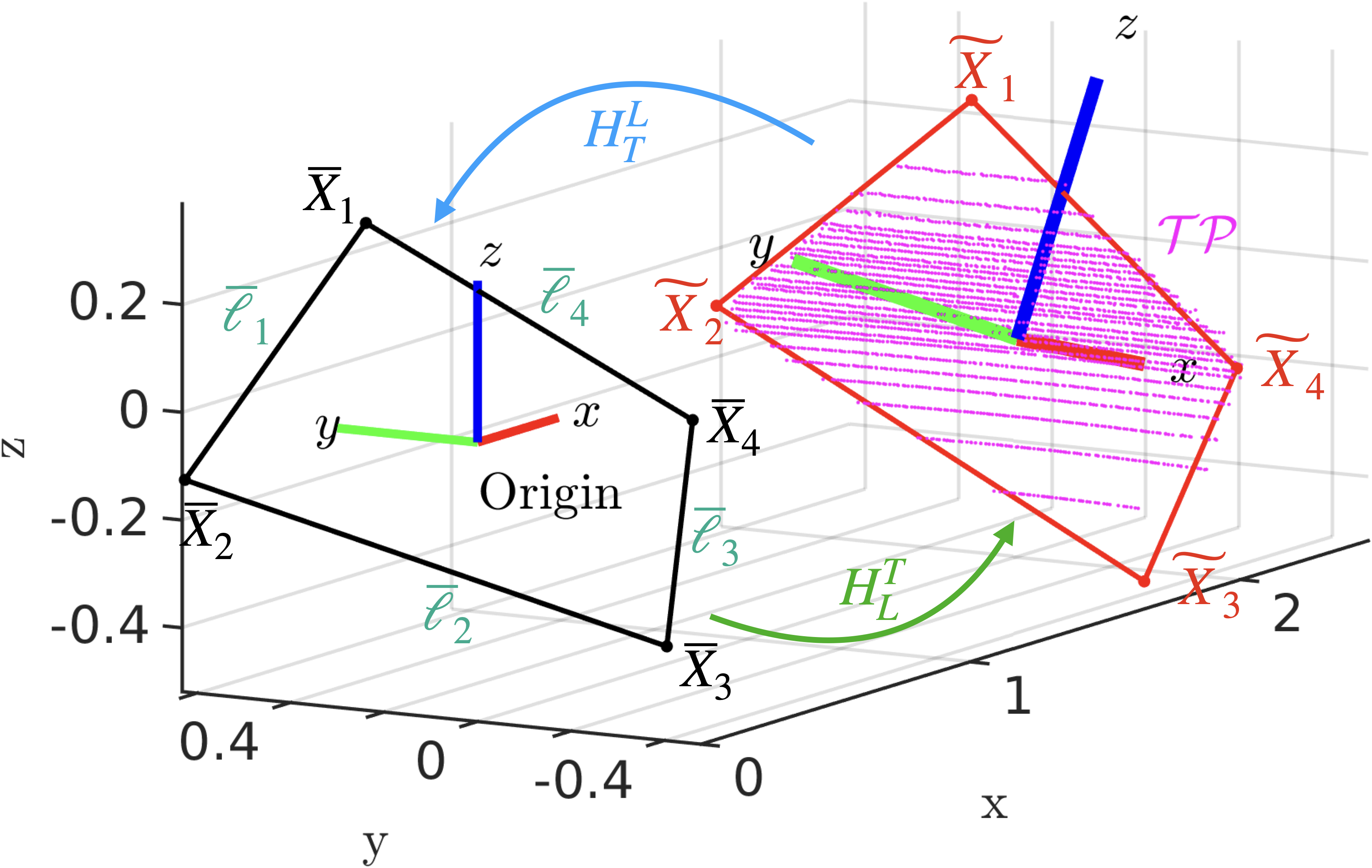}
\caption{Pose definition and illustration of template fitting.
    A coordinate frame for the template (target shown in black) is defined by aligning the plane of the template with the $y$-$z$ plane of the \lidar frame and also aligning the mean of its vertices with the origin of the \lidarN. 
    Let $H_T^L$ (blue arrow) be an estimate of the rigid-body transformation from target to \lidarN, projecting the edge points of the target back to the template. The
    estimated pose of the target is then given by the
    inverse transformation, $H_L^T$ (green arrow). The optimal $H_T^L$ is obtained by minimizing \eqref{eq:FittingError} (based on point-to-line distance).
    This figure also shows a fitting result of a target at 2 meters in the Ford
    Robotics Building. The red frame is the template re-projected onto the \lidar point cloud by $H_L^T$. 
} \label{fig:PoseDefinition}
\squeezeup
\end{figure}

In this section, we propose a means to use known target geometry to extract target
vertices while globally estimating relative pose between target and \lidarN. For a collection of \lidar returns $\tp := \{\Xcal_i\}_{i=i}^N$, let 
$\ep:=\{E_i\}_{i=1}^M\in\tp$ be the $M$ target edge points. Given the target geometry, we define a template with vertices
$\{\bar{X}_i\}_{i=1}^4$ located at the origin of the \lidar and aligned with the
$y$-$z$ plane as defined in Fig.~\ref{fig:PoseDefinition}. We also denote
$\overbar{\Lcal}:=\{\overbar{\ell}_i\}_{i=1}^4$ as the line equations of the adjacent
vertices of the template. We seek a rigid-body transformation from the template to
the target, $\H[L][T] \in \SE(3)$, that ``best fits'' the template onto the edge points. In practice, it is actually easier to project the
edge points $\ep$ back to the origin of the \lidar through the inverse of the
current estimate of transformation $\H[T][L]:=\inv{\left(\H[L][T]\right)}$ and
measure the error there. The action of $\H\in\SE(3)$ on $\reals^3$ is $\H\cdot\Xcal_i
= \R\Xcal_i+\t$, where $\R\in\SO(3)$ and $\t\in\reals^3$.

 \begin{figure*}[!b]%
    \centering
    \begin{subfigure}{0.48\columnwidth}
        \centering
        \includegraphics[width=1\columnwidth, trim={0 0 0 0},clip]{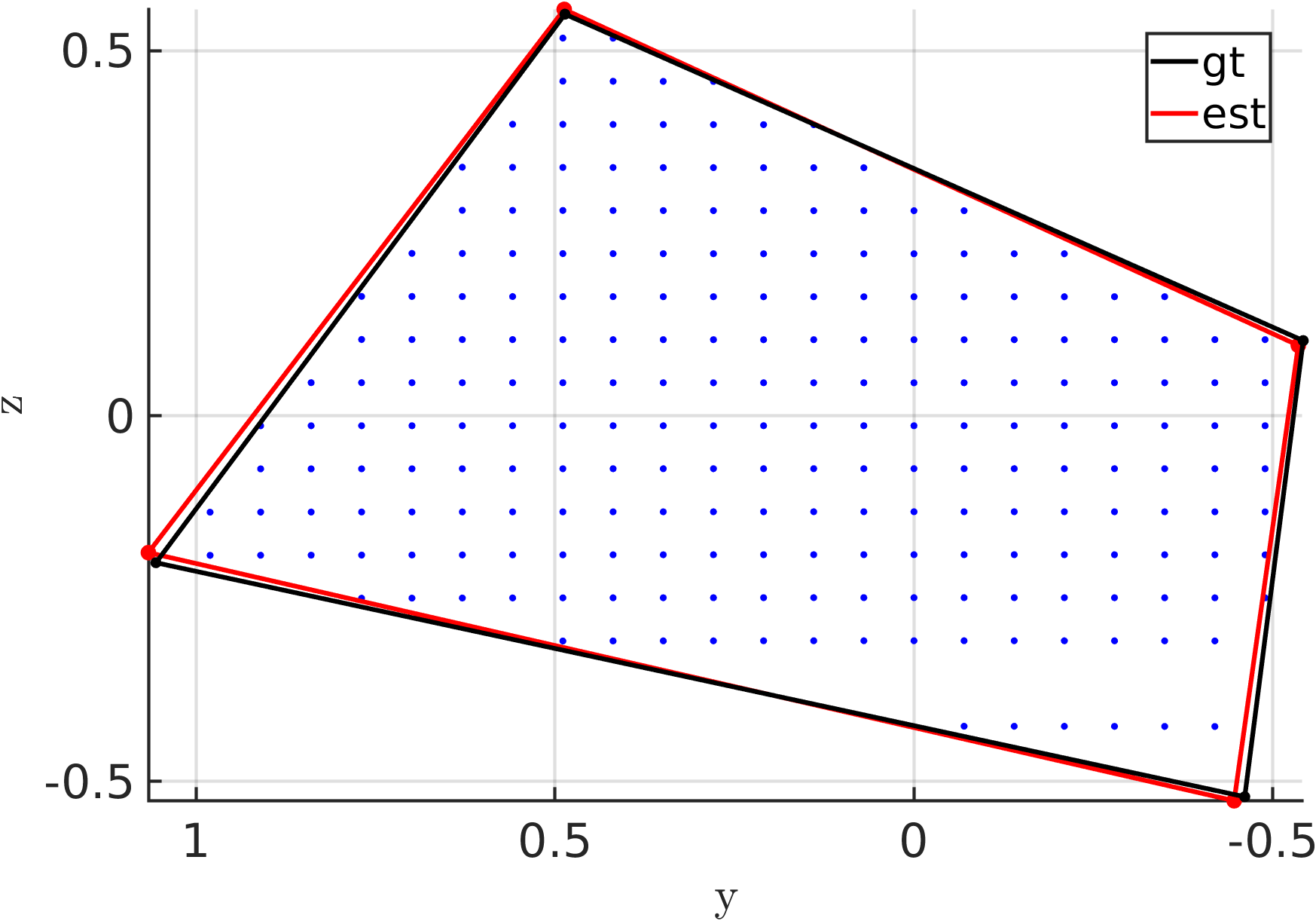}
    \end{subfigure}%
    \vspace{2pt}
    \begin{subfigure}{0.48\columnwidth}
        \centering
        \includegraphics[width=1\columnwidth, trim={0 0 0 0},clip]{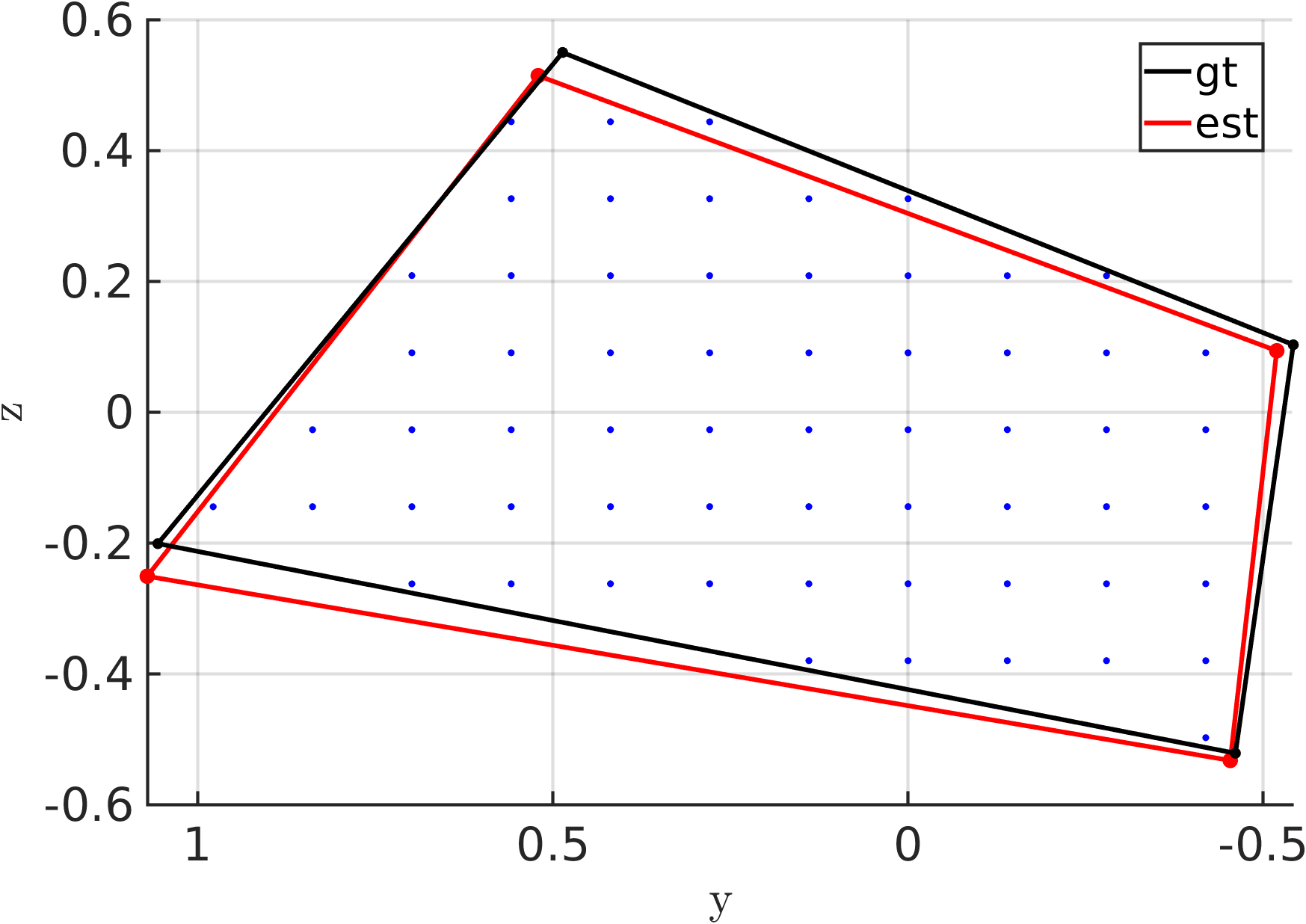}
    \end{subfigure}%
    \vspace{2pt}
    \begin{subfigure}{0.48\columnwidth}
        \centering
        \includegraphics[width=1\columnwidth, trim={0 0 0 0},clip]{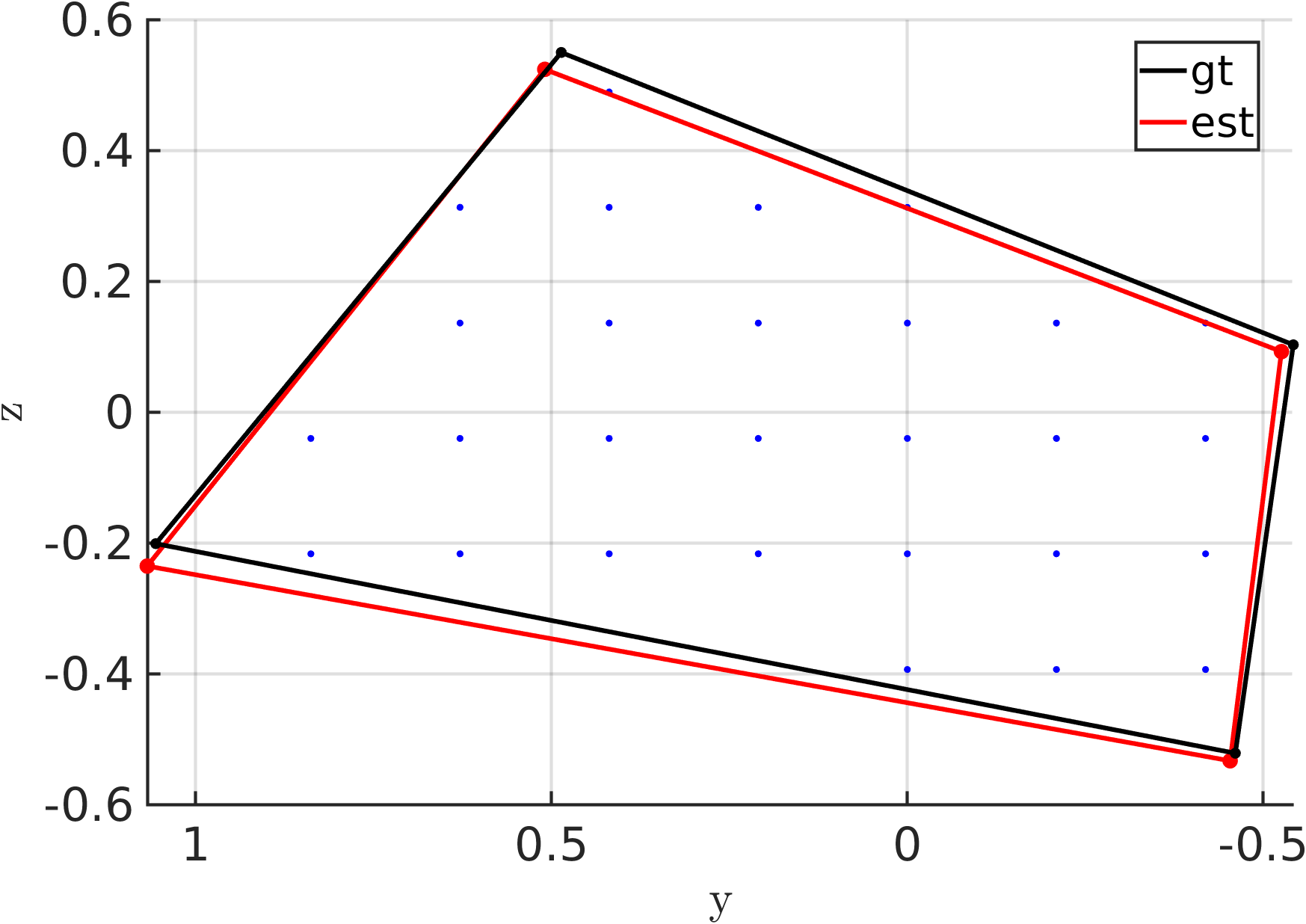}
    \end{subfigure}%
    \vspace{2pt}
    \begin{subfigure}{0.48\columnwidth}
        \centering
        \includegraphics[width=1\columnwidth, trim={0 0 0 0},clip]{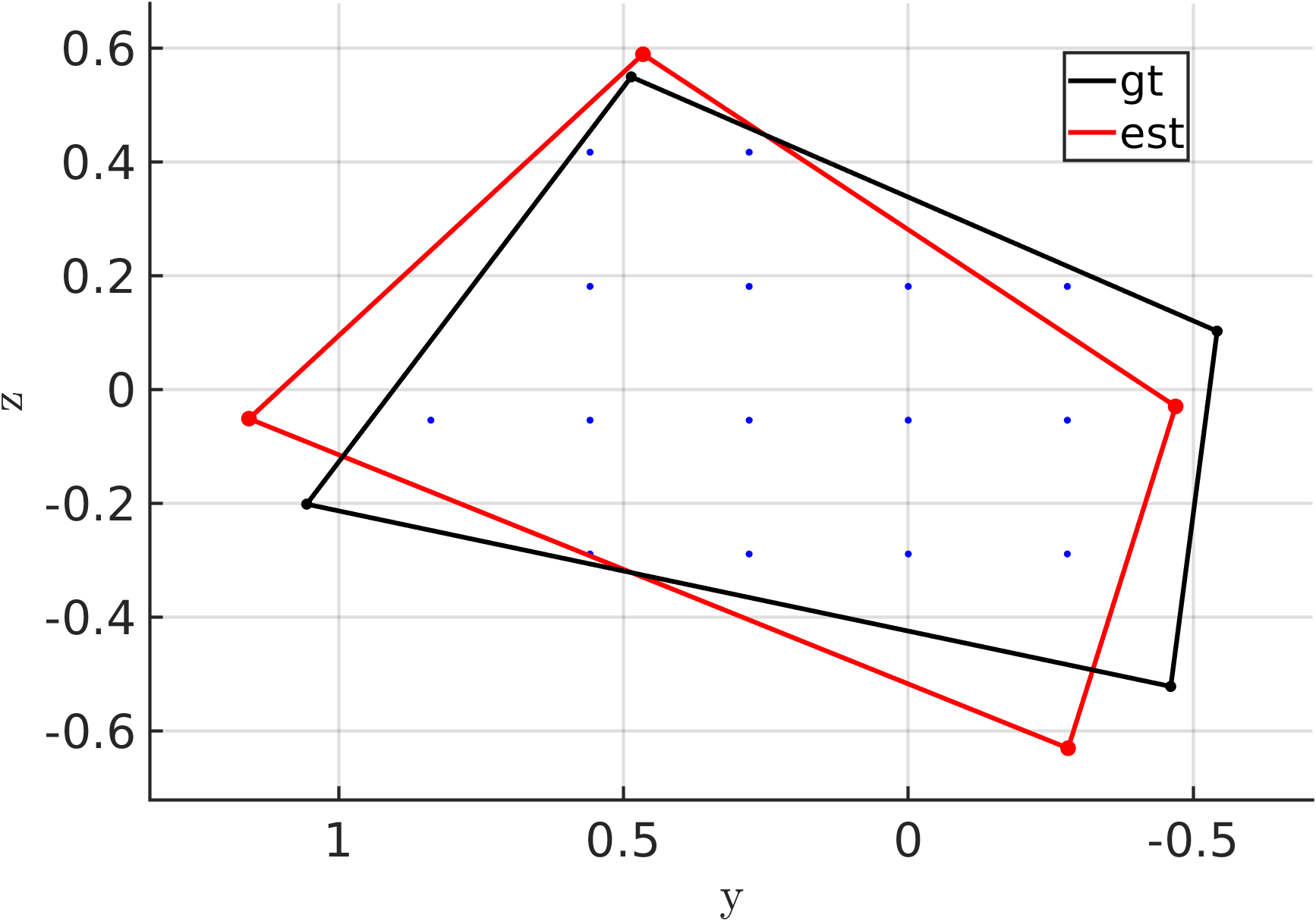}
    \end{subfigure}
    \vspace{2pt}
    \begin{subfigure}{0.48\columnwidth}
        \centering
        \includegraphics[width=1\columnwidth, trim={0 0 0 0},clip]{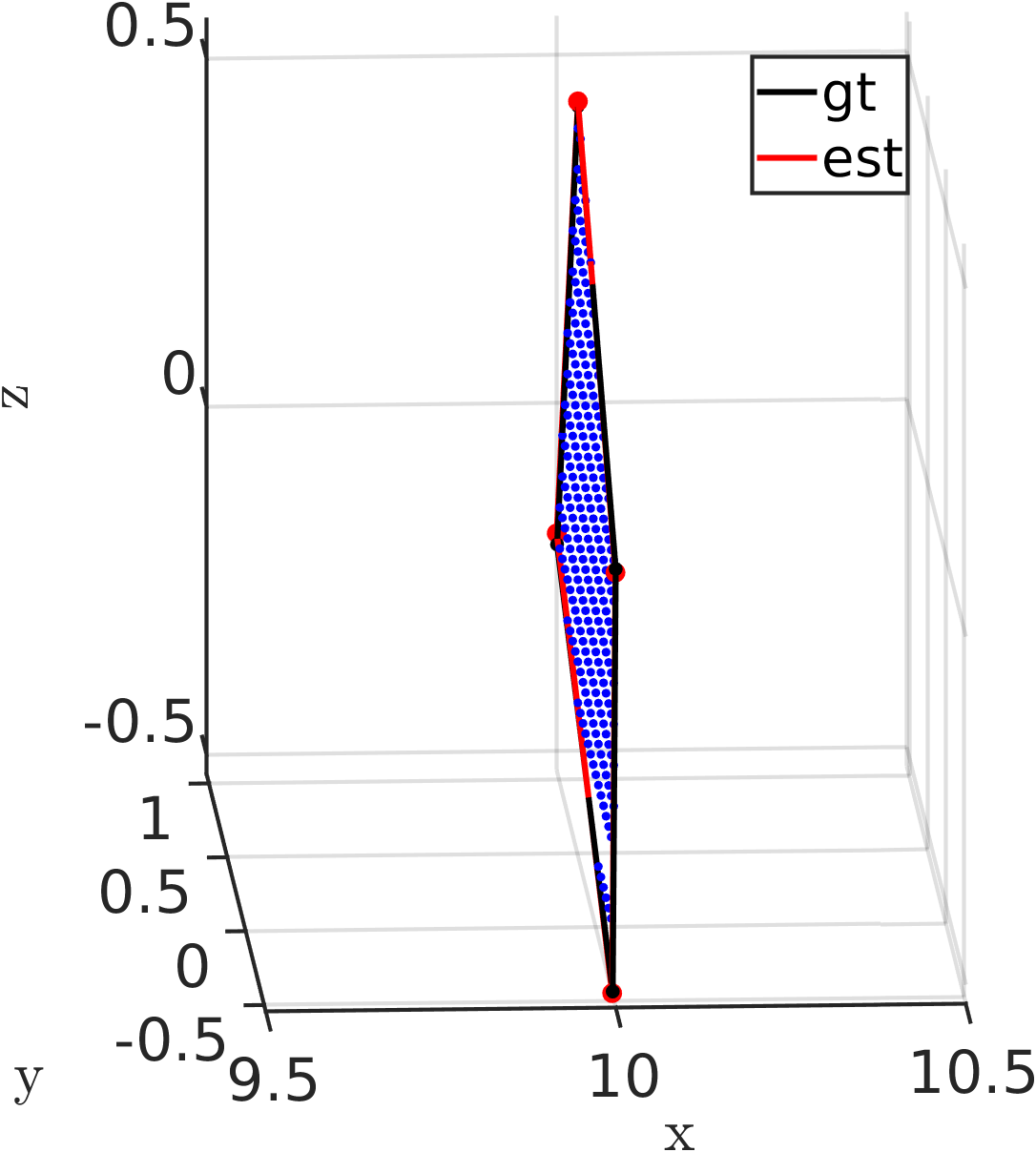}
    \end{subfigure}%
    \vspace{2pt}
    \begin{subfigure}{0.48\columnwidth}
        \centering
        \includegraphics[width=1\columnwidth, trim={0 0 0 0},clip]{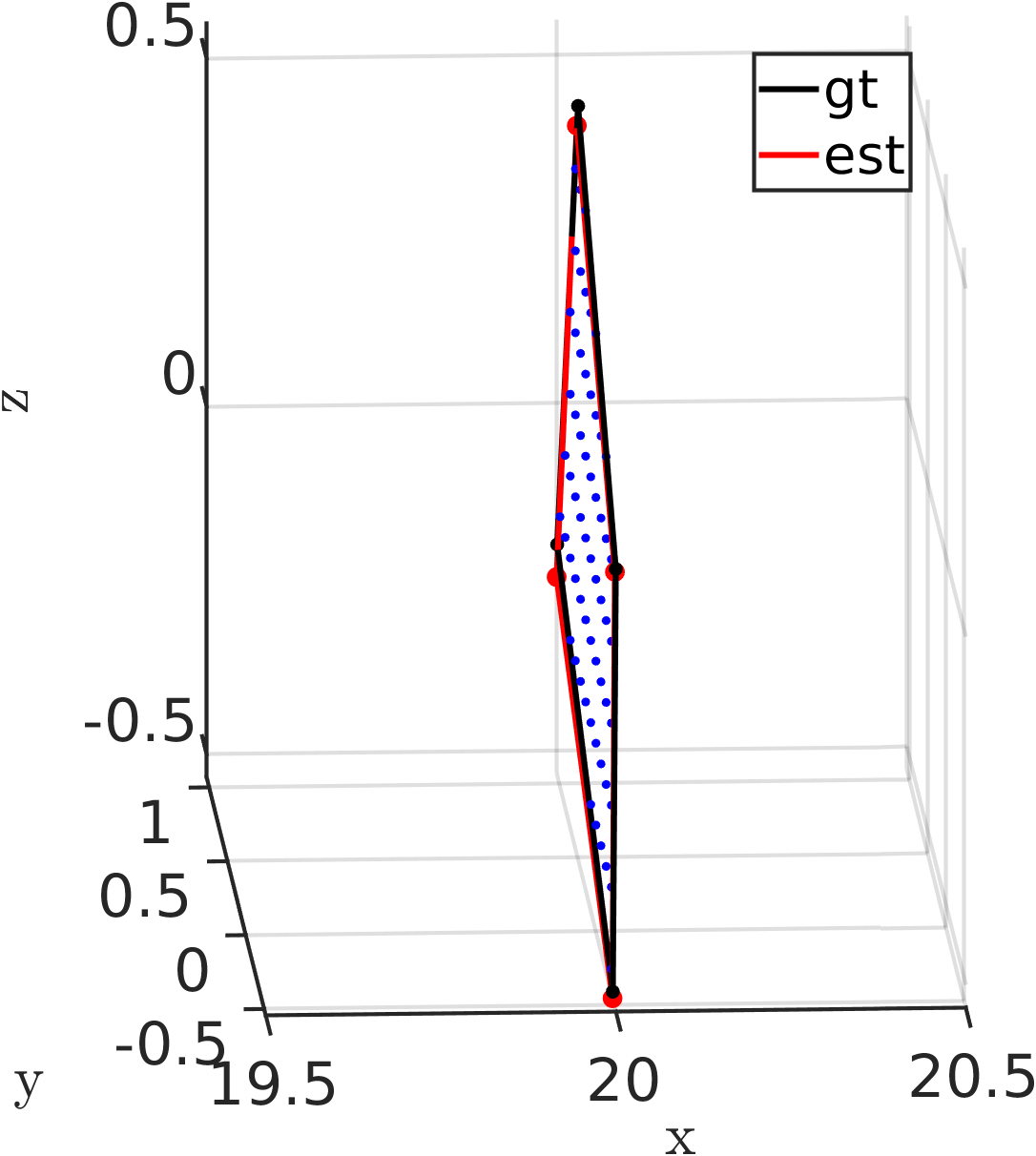}
    \end{subfigure}%
    \vspace{2pt}
    \begin{subfigure}{0.48\columnwidth}
        \centering
        \includegraphics[width=1\columnwidth, trim={0 0 0 0},clip]{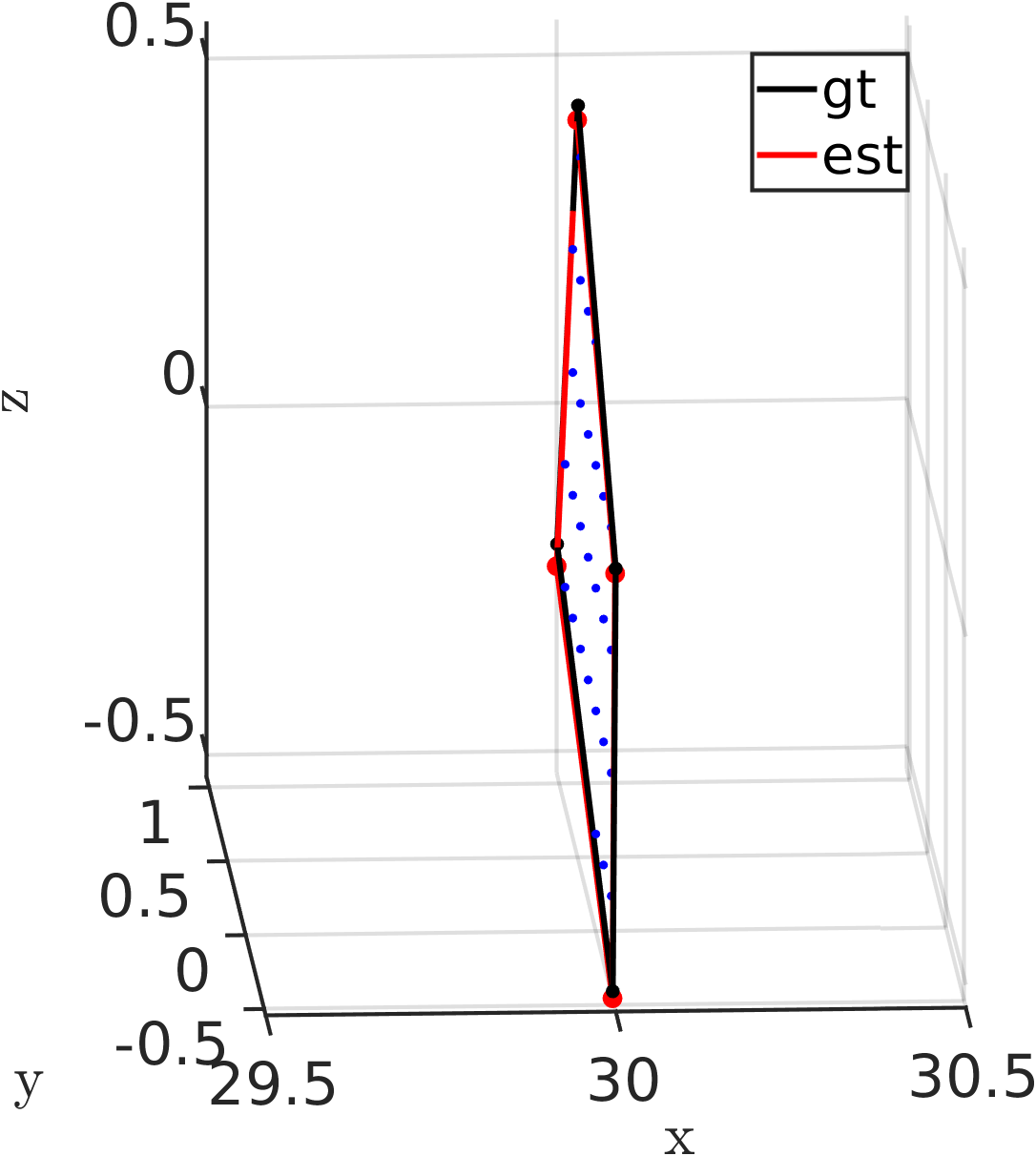}
    \end{subfigure}%
    \vspace{2pt}
    \begin{subfigure}{0.48\columnwidth}
        \centering
        \includegraphics[width=1\columnwidth, trim={0 0 0 0},clip]{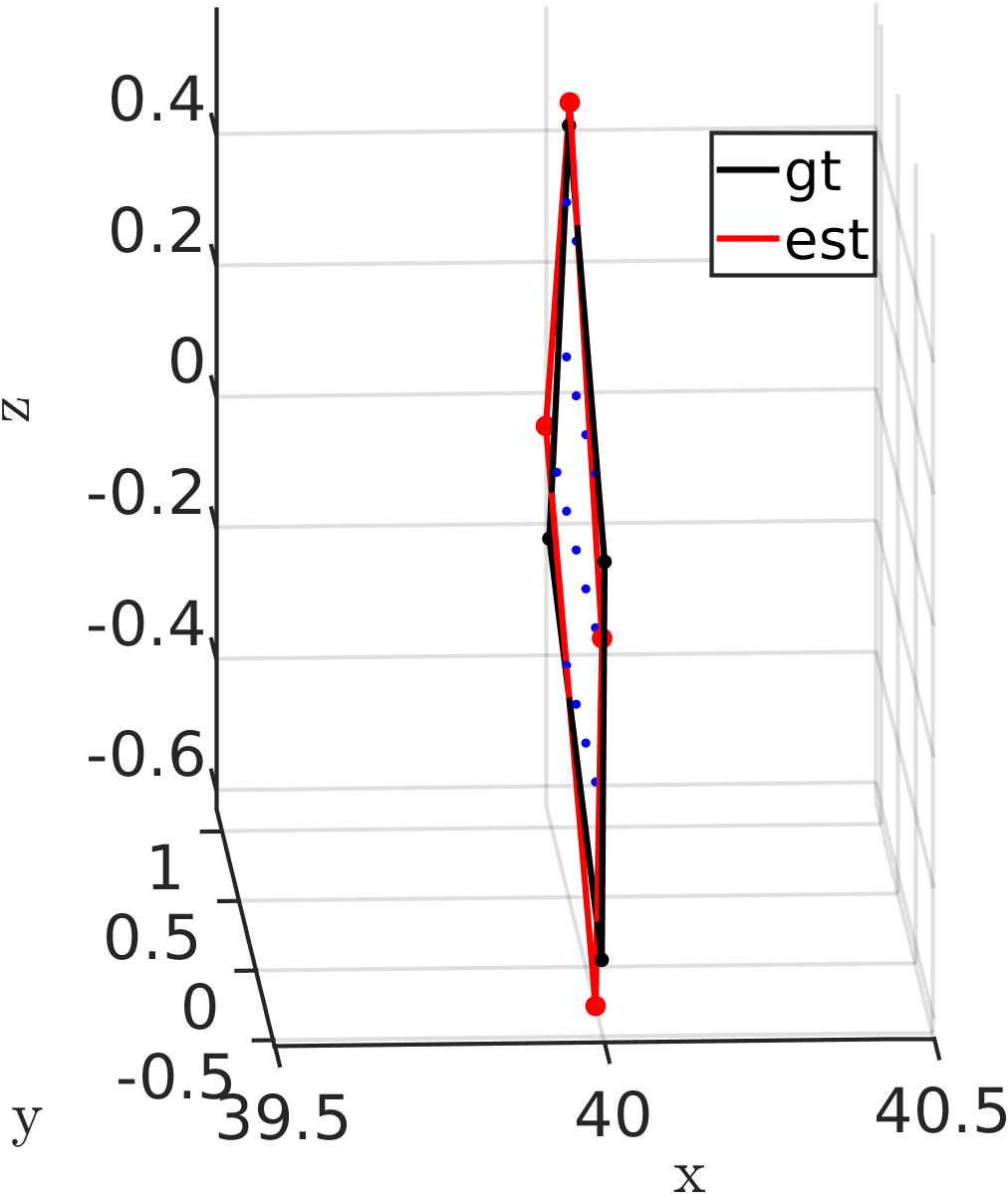}
    \end{subfigure}%
    \caption[]{Simulation results of the noise-free dataset of the pose estimation at various distances (10, 20, 30, 40 m). 
        \lidar returns (blue dots) on the target are provided by the \lidar simulator. 
        Black indicates the ground truth pose from the simulator, and red is the estimated pose and vertices. The top
   and bottom show the front view and a side view of the fitting results,
   respectively.
}%
    \label{fig:simResults}%
\end{figure*}

The cost $j_i$ of edge point $E_i\in\ep$ is defined as the point-to-line distance, 
\begin{equation}
        j_i(E_i; \overbar{\Lcal}) =
        \min_{\overbar{\ell}_i\in\overbar{\Lcal}}\|E_i-\overbar{\ell}_i\|_2^2
\end{equation}
where $\overbar{\Lcal}$ is the set of line equations for the target. Let
$\{\bar{E}_i\}_{i=1}^{M} := H_T^L(\ep)=\{H_T^L\cdot E_i\}_{i=1}^M$ denote the
projected points by $H_T^L$. The total fitting error is defined as
\begin{equation}
    \label{eq:FittingError}
    J(H_T^L(\ep)) := \sum_{i=1}^M j_i(\bar{E}_i)
\end{equation}


To minimize \eqref{eq:FittingError}, we adopt techniques that were used to globally
solve 3D registration or 3D SLAM~\cite{tron2015inclusion, carlone2015lagrangian,
olsson2008solving, briales2017convex} where the problem is formulated as a QCQP, and
the Lagrangian dual relaxation is used. The relaxed problem becomes a Semidefinite
Programming (SDP) and convex. The problem can thus be solved globally and efficiently
by off-the-shelf specialized solvers~\cite{grant2014cvx}. As shown
in~\cite{briales2017convex}, the dual relaxation is empirically always tight (the
duality gap is zero). 

Once we (globally) obtain $\H[T][L]$, the pose of the target is $\H[L][T] =
\inv{(\H[T][L])}$, and the estimated vertices are $\{\widetilde{X_i}\}_{i=i}^4 :=
\{\H[L][T]\cdot\bar{X}_i\}_{i=1}^4$. The edge-line equations, the normal vector, and
the plane equation of the target can be readily obtained from the vertices.


\begin{figure*}[!t]%
    \centering
    \begin{subfigure}{0.68\columnwidth}
        \centering
        \includegraphics[width=1\columnwidth, trim={0 0 0 0},clip]{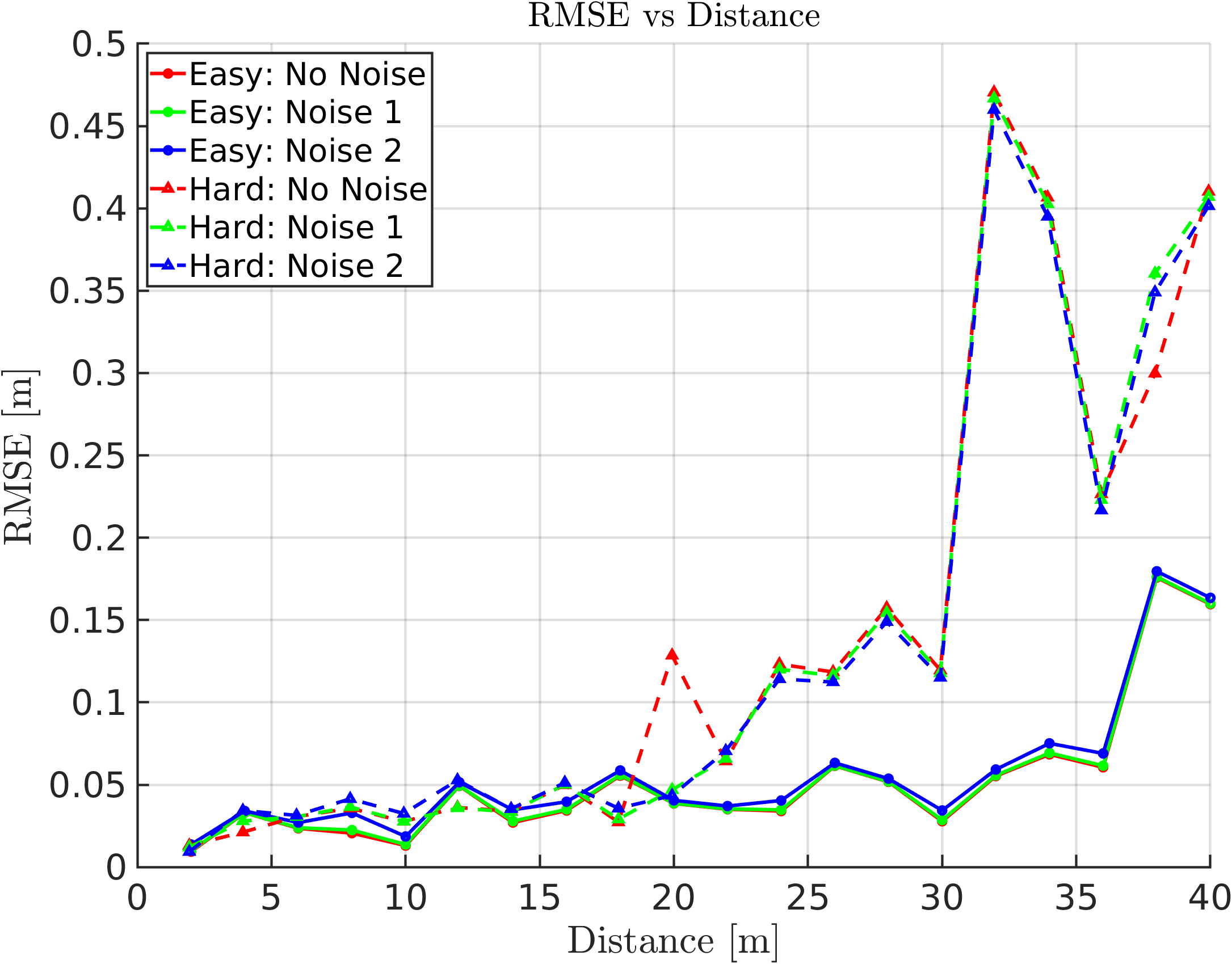}
    \end{subfigure}%
    \begin{subfigure}{0.68\columnwidth}
        \centering
        \includegraphics[width=1\columnwidth, trim={0 0 0 0},clip]{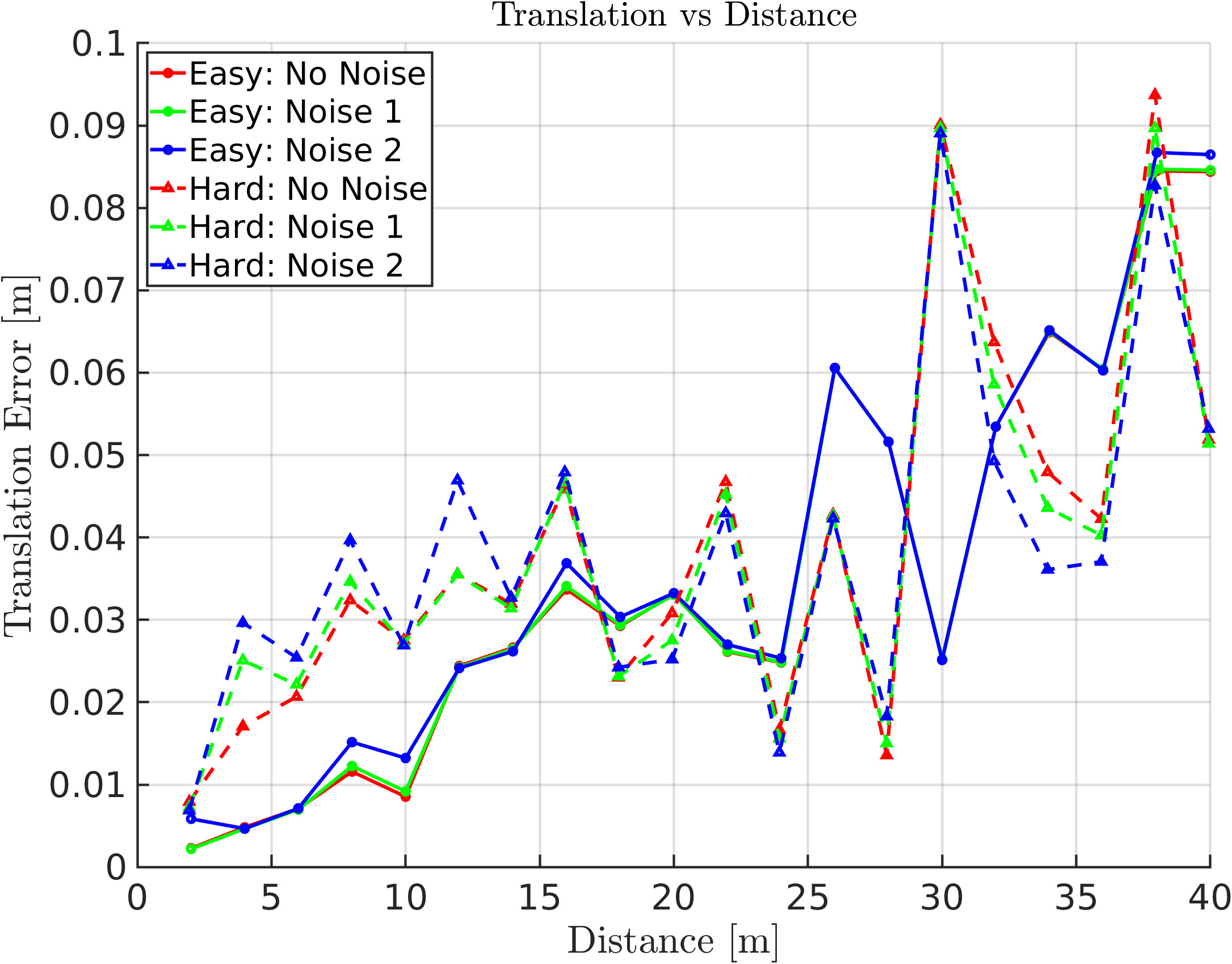}
    \end{subfigure}%
    \begin{subfigure}{0.68\columnwidth}
        \centering
        \includegraphics[width=1\columnwidth, trim={0 0 0 0},clip]{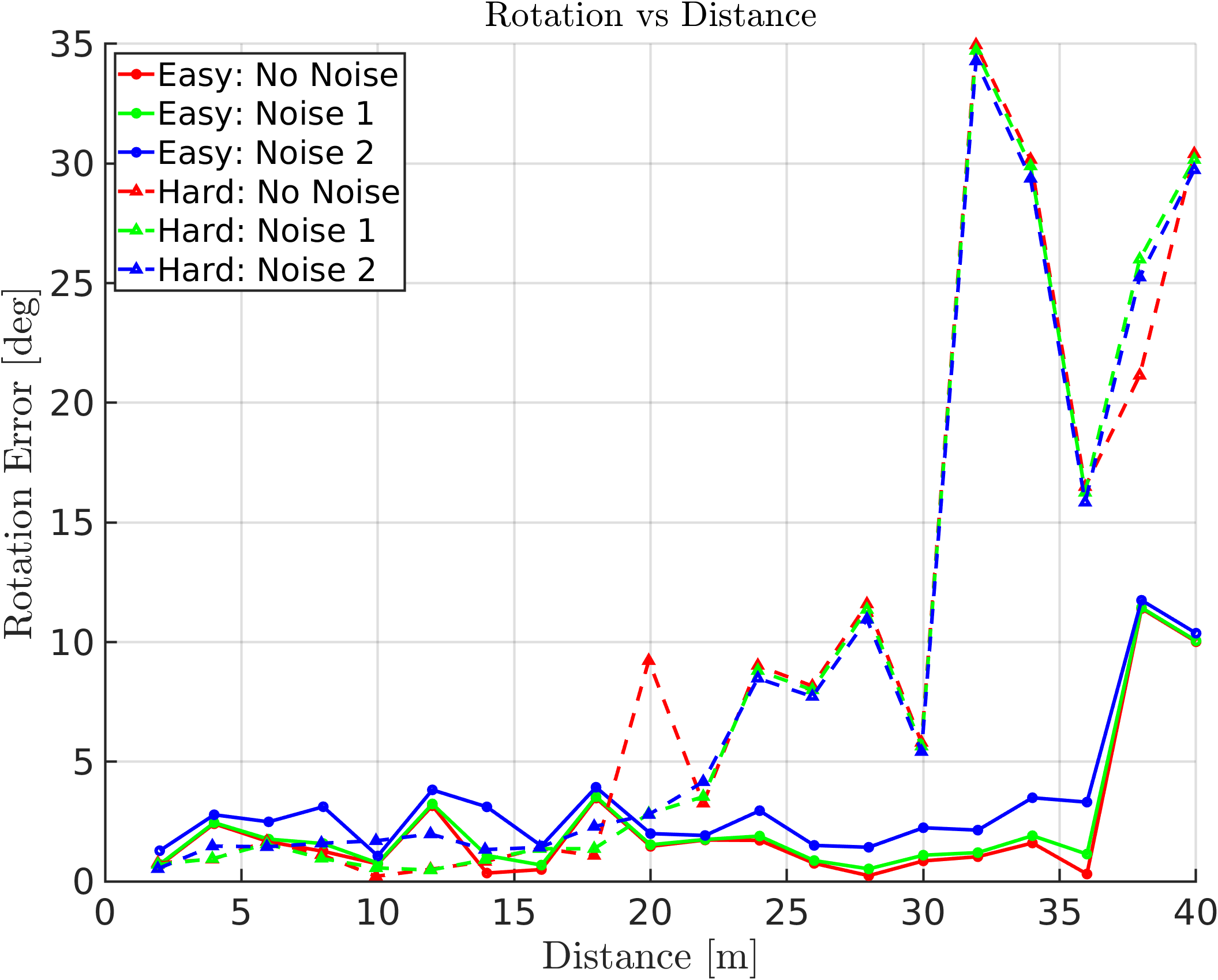}
    \end{subfigure}
    \begin{subfigure}{0.68\columnwidth}
        \centering
        \includegraphics[width=1\columnwidth, trim={0 0 0 0},clip]{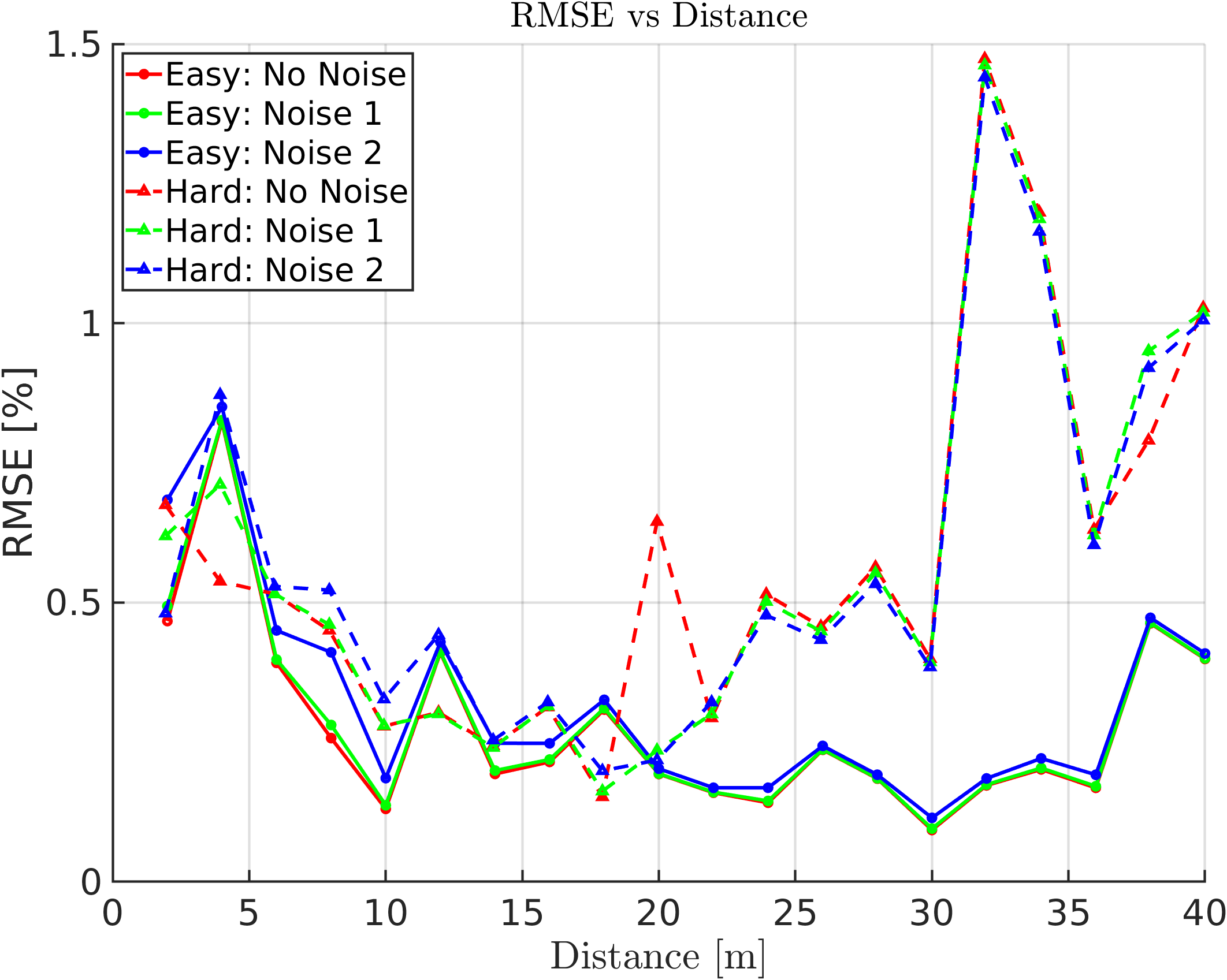}
    \end{subfigure}%
    \begin{subfigure}{0.68\columnwidth}
        \centering
        \includegraphics[width=1\columnwidth, trim={0 0 0 0},clip]{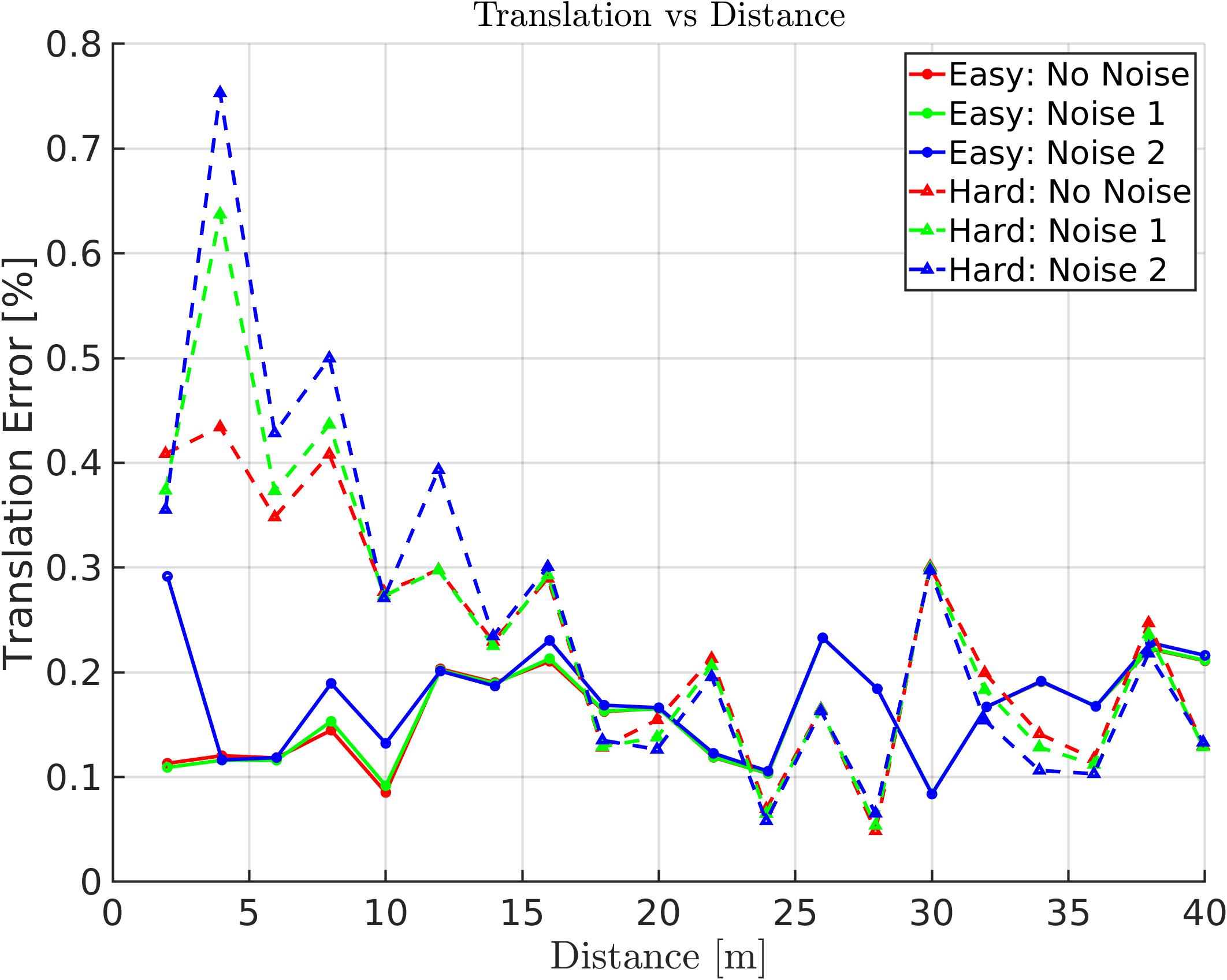}
    \end{subfigure}%
    \caption[]{Simulation results with a target placed at distances from 2
    to 40 meters in 2 m increments in the \lidar simulator. At each distance, the simulation data are collected
    with the target face-on to the \lidar as an easy case (solid line), and for the
    other, the target is 
rotated by the Euler angle (roll = $20^\circ$, pitch = $30^\circ$, yaw = $30^\circ$)
as a challenging case (dashed line). In addition, we induce two different levels of
noises to each dataset, as indicated by the different colors.}%
    \label{fig:simChart}%
\end{figure*}

\section{Simulation Results}
\label{sec:OptimalShapeSimulation}
Before carrying out experiments with the new target shape, we used a MATLAB-based
\lidar simulator introduced in \cite{huang2020intinsic} to extensively evaluate the
pose and vertex estimation of the optimal shape. Both quantitative and qualitative
results are provided. We do not compare against standard targets, such as unpatterned rectangles, diamonds, or circles, because their symmetry properties result in  pose ambiguity. At large distances, a pattern would not be discernible.

We simulate a \velodyneN, whose data sheet can be found
at\cite{velodyneUltraPuck}. A target is placed at distances from 2 to 40 meters in 2
m increments. At each distance, simulation data is collected with a target face-on to
the \lidar as an easy case, and another dataset with the target rotated by the Euler
angles (roll = $20^\circ$, pitch = $30^\circ$, yaw = $30^\circ$) under the ${XYZ}$
convention as a challenging case. In addition, we induce two different levels
of noise to each dataset to examine the robustness of the algorithm.

The results of vertex estimation are reported as the root-mean-square-error (RMSE):
\begin{equation}
\label{eq:e_RMSE}
    \text{RMSE} = \sqrt{\frac{1}{4}\sum_{i=1}^4\|\widetilde{X}_i - X_i\|_2^2},
\end{equation}
where $\widetilde{X}_i$ is the estimated vertex and $X_i$ is the ground truth vertex
from the simulator. The pose on $\SE(3)$ is evaluated on translation $e_t$ in
$\real^3$ and rotation $e_r$ on $\SO(3)$, separately. In particular, $e_t$ and $e_r$
are computed by 
\begin{equation}
\label{eq:PoseError}
    e_t := \|\t - \widetilde{\t}\| \text{~~and~~}
    e_r := \|\Log(\R\widetilde{\R}^\transpose)\|, 
\end{equation}
where $\|\cdot\|$ is the Euclidean norm, $\widetilde{\cdot}$ is the estimated
quantity, $\R$ and $\t$ are the ground truth rotation and translation, respectively,
and $\Log(\cdot)$ is the logarithm map in the Lie group $\SO(3)$. Additionally, we
also report the percentage of the RMSE and translation error, which are computed by
each quantity divided by the centroid of the target. The quantization error $e_q$
is the distance between two adjacent points on the same ring and can be approximated
by the azimuth resolution ($0.4^\circ$) of the \lidar times the target distance. 

Qualitative results are shown in Figure~\ref{fig:simResults}. The complete
quantitative results of the distances and noise levels are shown as line charts in
Fig.~\ref{fig:simChart}. Table~\ref{tab:SimNoiseFreeResults} shows a subset of
quantitative results of the pose and vertex estimation using the noise-free dataset. For vertex estimation, we achieve
        less than 1\% error in most cases. The
    translation errors are less than the quantization error $e_q$. We also achieve a few degrees of rotation errors. It can be seen that the estimation limit of the optimal target of width 0.96 meter with our \velodyne optimal is 30 meters. However, for a LiDAR with
a different number of beams or points, the estimation limit may be different.
Based on these results, we were motivated to build the target and run physical
experiments.


\section{Experimental Results}
\label{sec:OptimalShapeExperiment}
\begin{figure}[t]%
\centering
\includegraphics[trim=0 10 0 0,clip,width=0.8\columnwidth]{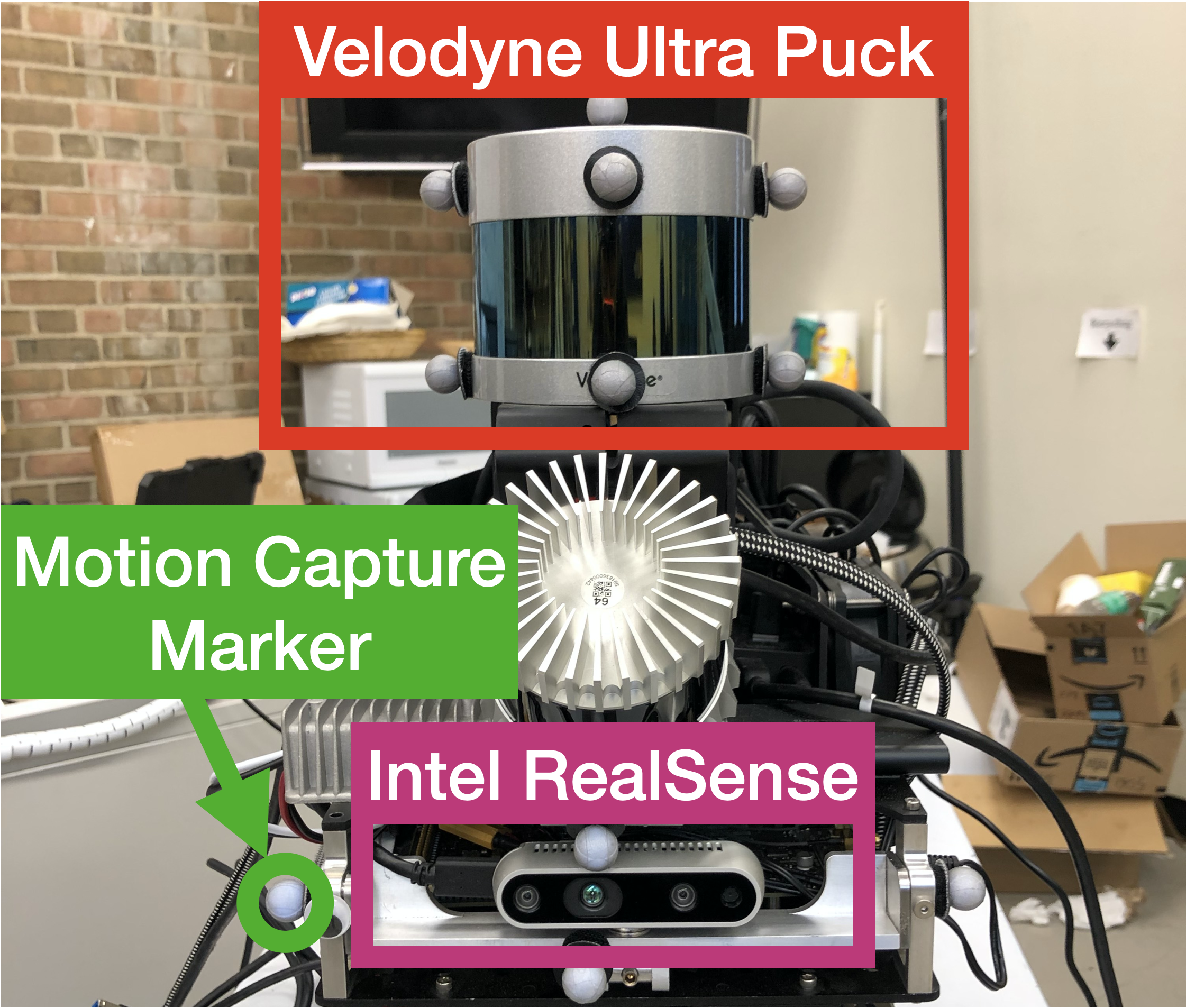}%
\caption[]{
The sensor suite consists of a \lidarN, a camera and several motion capture markers.
}%
\label{fig:torso}%
\end{figure}

We now present experimental evaluations of the pose and vertex estimation of the
optimal shape. All the experiments are conducted with a \velodyne and an Intel
RealSense camera rigidly attached to the torso of a Cassie-series bipedal robot, as shown in Fig.~\ref{fig:torso}. We use the Robot Operating System
(ROS)~\cite{ros} to communicate and synchronize between the sensors. Datasets are
collected in the atrium of the Ford Robotics Building at the University of Michigan,
and a spacious outdoor facility, M-Air~\cite{MAir}, equipped with a motion capture
system. 

\subsection{Quantitative Experimental Results in M-Air}
\begin{figure}[t]%
\centering
\includegraphics[trim=0 0 0 0,clip,width=1\columnwidth]{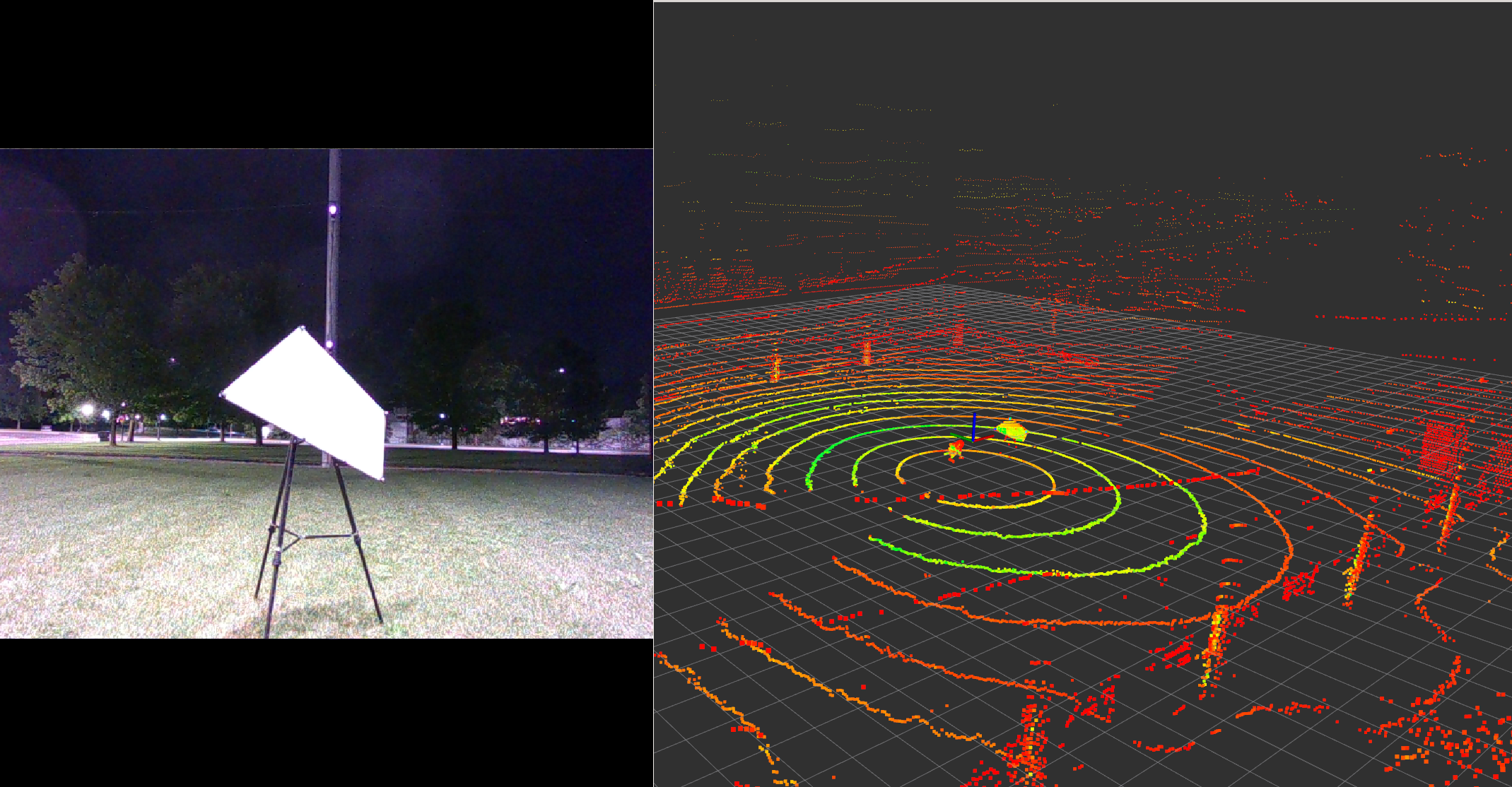}%
\caption[]{
The Experimental setup. The left shows passive markers are attached to the four corners of the optimized target shape and the right shows a \lidar scan in M-Air.
}%
\label{fig:MAirSetup}%
\end{figure}

The Qualisys motion capture system in M-Air is used as a proxy for ground truth
poses and vertices. The setup consists of 33 motion capture cameras with passive markers
attached to the target, the \lidar and the camera, as shown in Fig.~\ref{fig:torso} and Fig.~\ref{fig:MAirSetup}. Datasets
are collected at various distances and angles. Each of the datasets contains images
(20 Hz) and scans of point clouds (10 Hz). Similar to the simulation environment, the
optimal-shape target is placed at distances from 2 to 16 meters (maximum possible in M-Air) in 2 meter increments. At each distance, data is collected with a
target face-on to the \lidar and another dataset with the target roughly rotated by
the Euler angles (roll = $20^\circ$, pitch = $30^\circ$, yaw = $30^\circ$) as a
challenging case. The results are shown in Table~\ref{tab:ExpResults}. As expected,
the results are slightly worse (approximately one degree) than the simulator's due to the white noise of the
\lidar and many missing returns on the targets, as shown in Fig~\ref{fig:EXPResults}.

\begin{remark}
A Velodyne \lidar
return consists of the point's Cartesian coordinates, intensity, and ring number. For each ring, the first and the last
point are the edge points of the ring. Since we define a template at the \lidar
origin, we first center the target points by subtracting its centroid. Each
centered edge point is associated with the closest edge. Given the current association, we estimate the pose and then
redo edge-point-to-edge-line association. Therefore, the optimization process is an
alternating process. The optimization is terminated if $\|\Log(H_{k-1}H_k)\|$ is
smaller than $1e^{-5}$, where $\Log(\cdot)$ is the logarithm map.
\end{remark}

\subsection{Qualitative Experimental Results and Target Partial Illumination}
\begin{figure*}[!t]%
    \centering
    \begin{subfigure}{0.48\columnwidth}
        \centering
        \includegraphics[height=0.7\columnwidth, trim={0 0 0 0},clip]{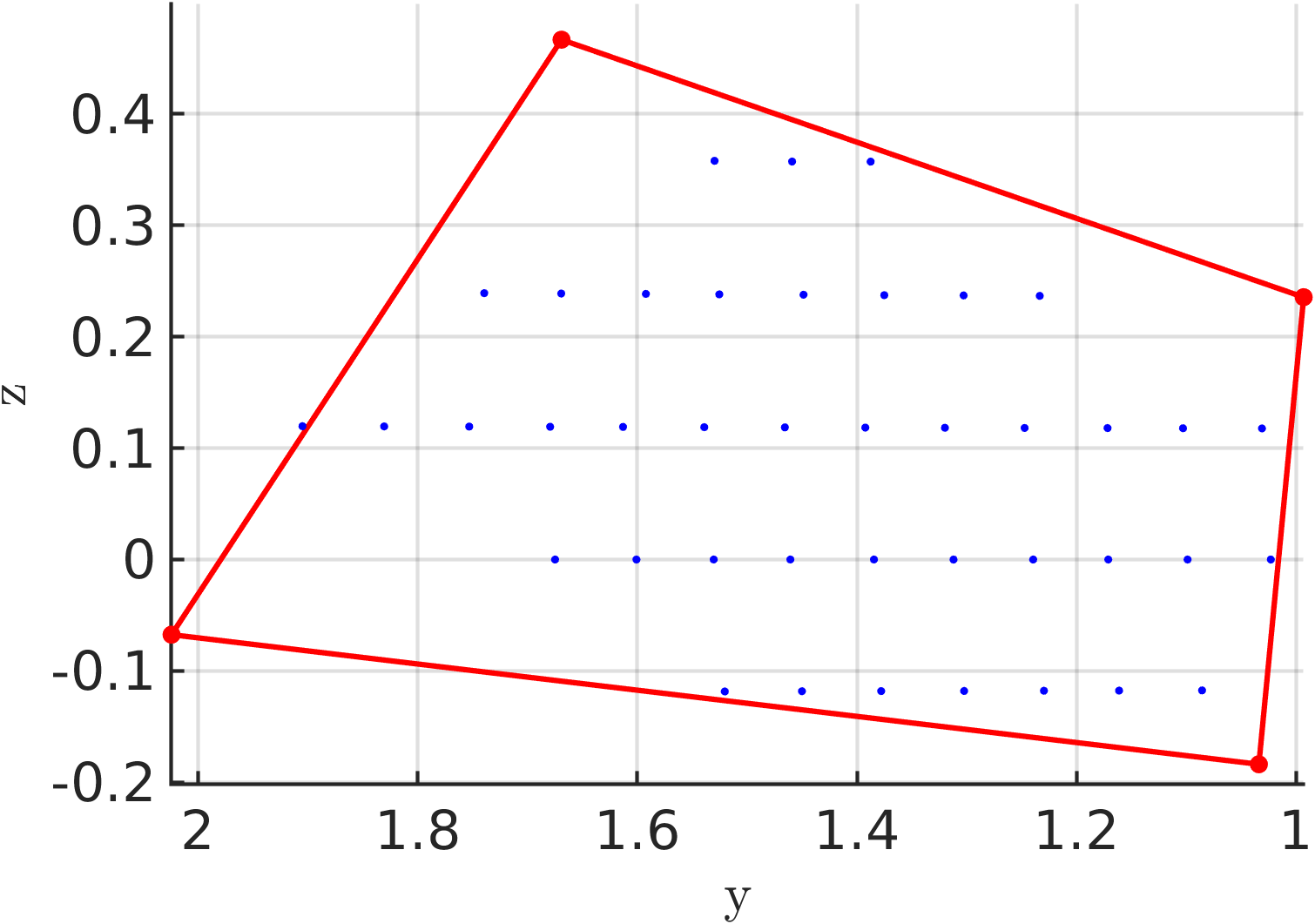}
    \end{subfigure}%
    \vspace{2pt}
    \begin{subfigure}{0.48\columnwidth}
        \centering
        \includegraphics[height=0.7\columnwidth, trim={0 0 0 0},clip]{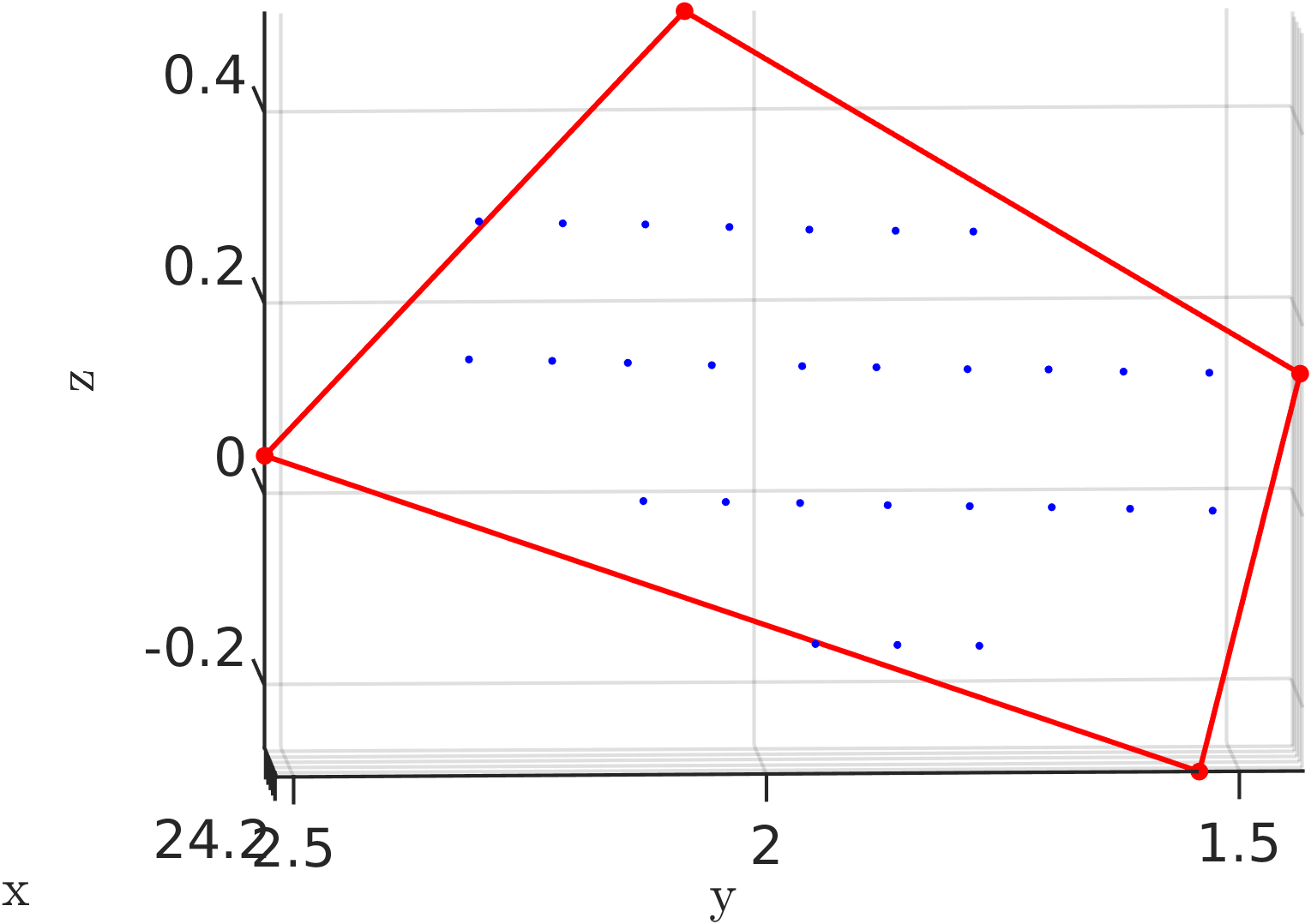}
    \end{subfigure}%
    \vspace{2pt}
    \begin{subfigure}{0.48\columnwidth}
        \centering
        \includegraphics[height=0.7\columnwidth, trim={0 0 0 0},clip]{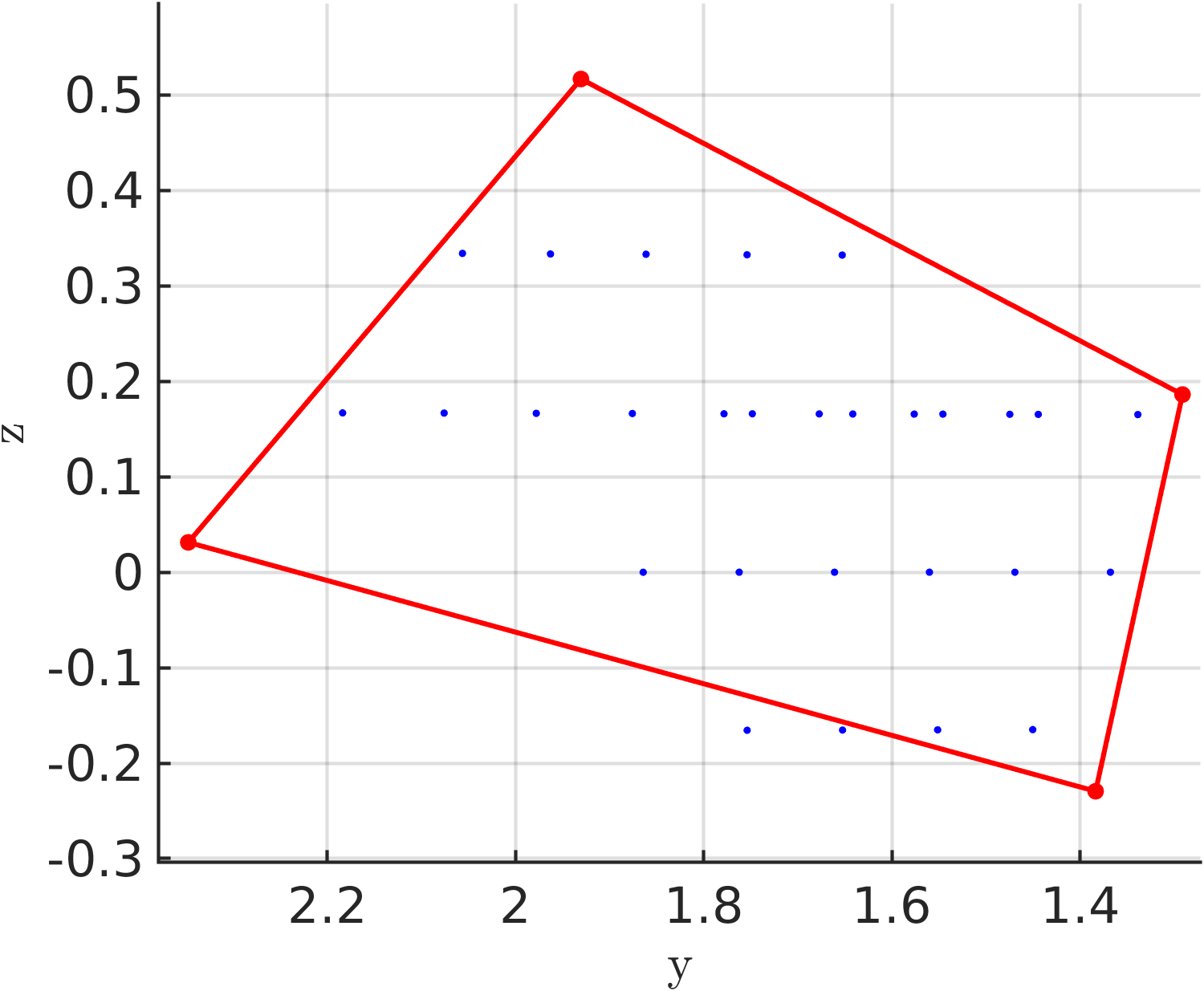}
    \end{subfigure}%
    \vspace{2pt}
    \begin{subfigure}{0.48\columnwidth}
        \centering
        \includegraphics[height=0.7\columnwidth, trim={0 0 0 0},clip]{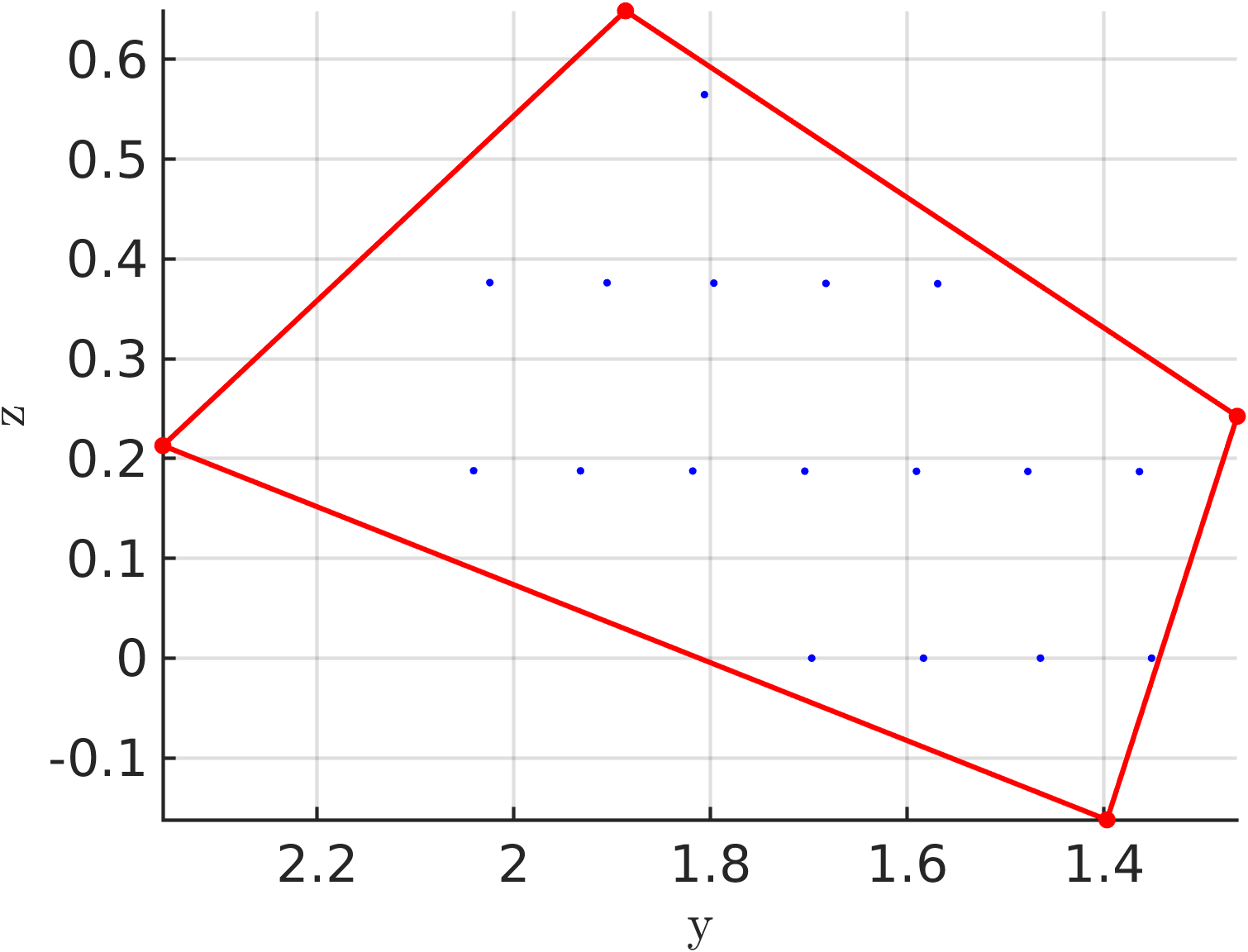}
    \end{subfigure}
    \vspace{2pt}
    \begin{subfigure}{0.48\columnwidth}
        \centering
        \includegraphics[height=1.1\columnwidth, trim={0 0 0 0},clip]{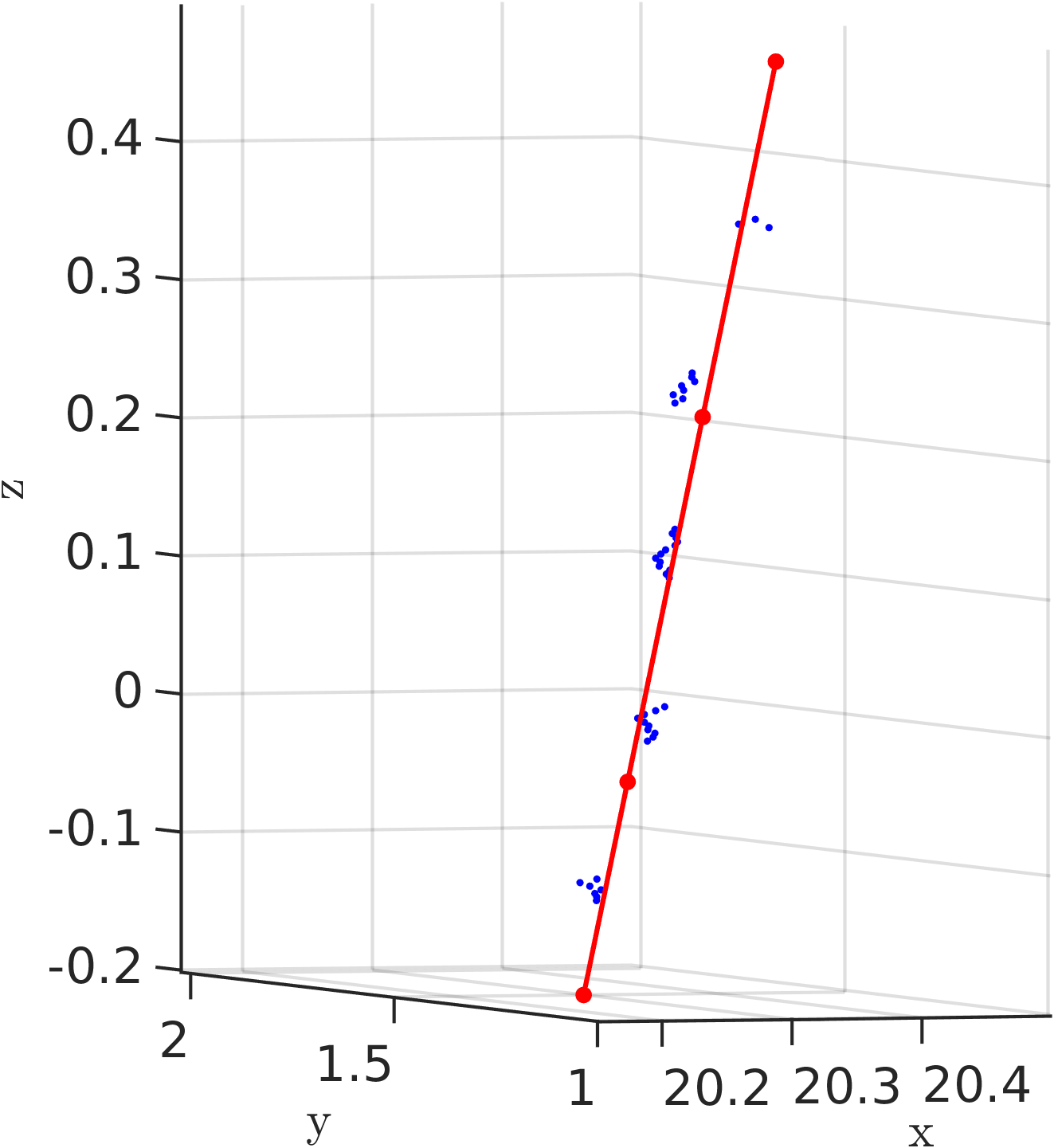}
    \end{subfigure}%
    \vspace{2pt}
    \begin{subfigure}{0.48\columnwidth}
        \centering
        \includegraphics[height=1.1\columnwidth, trim={0 0 0 0},clip]{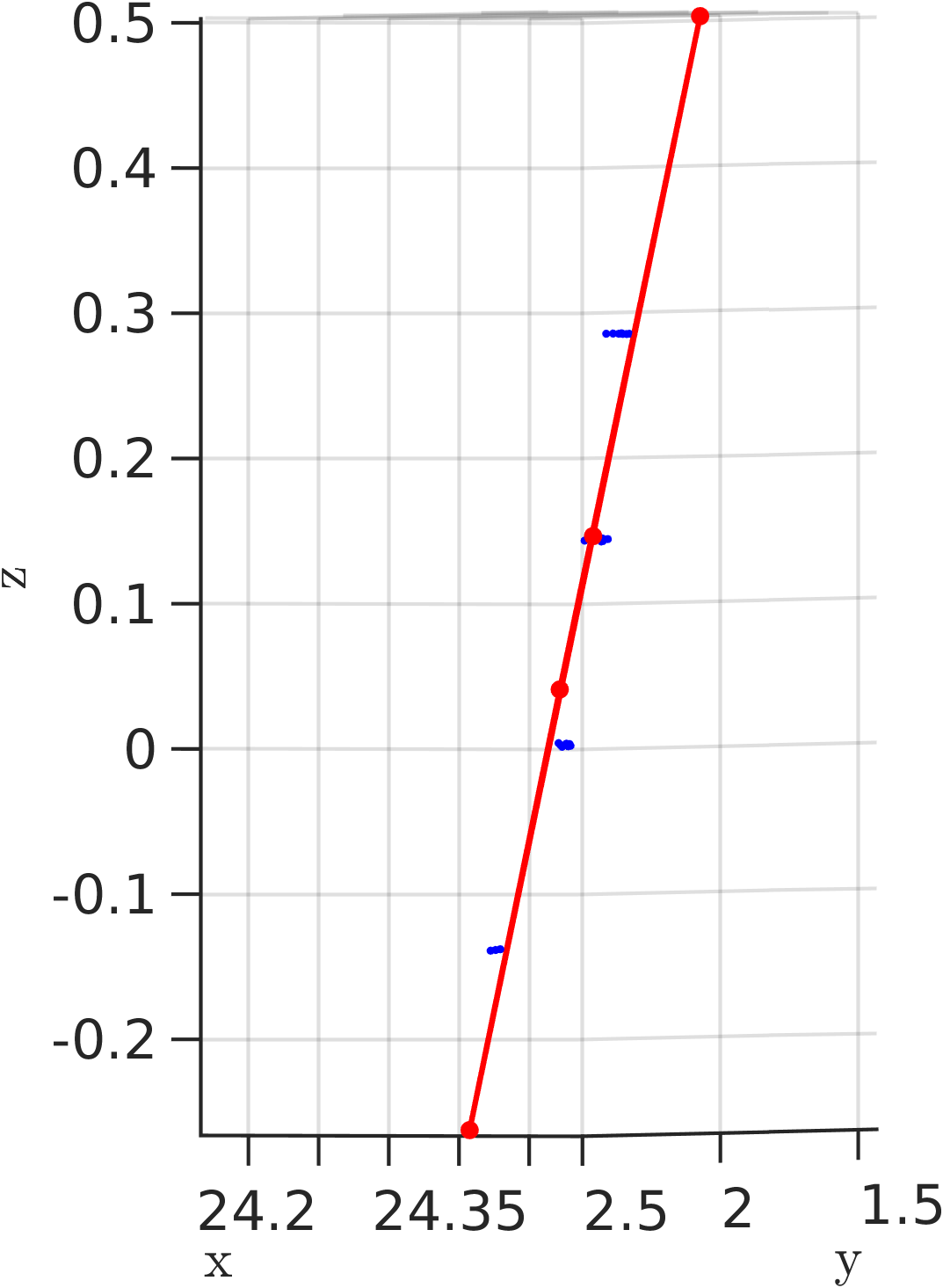}
    \end{subfigure}%
    \vspace{2pt}
    \begin{subfigure}{0.48\columnwidth}
        \centering
        \includegraphics[height=1.1\columnwidth, trim={0 0 0 0},clip]{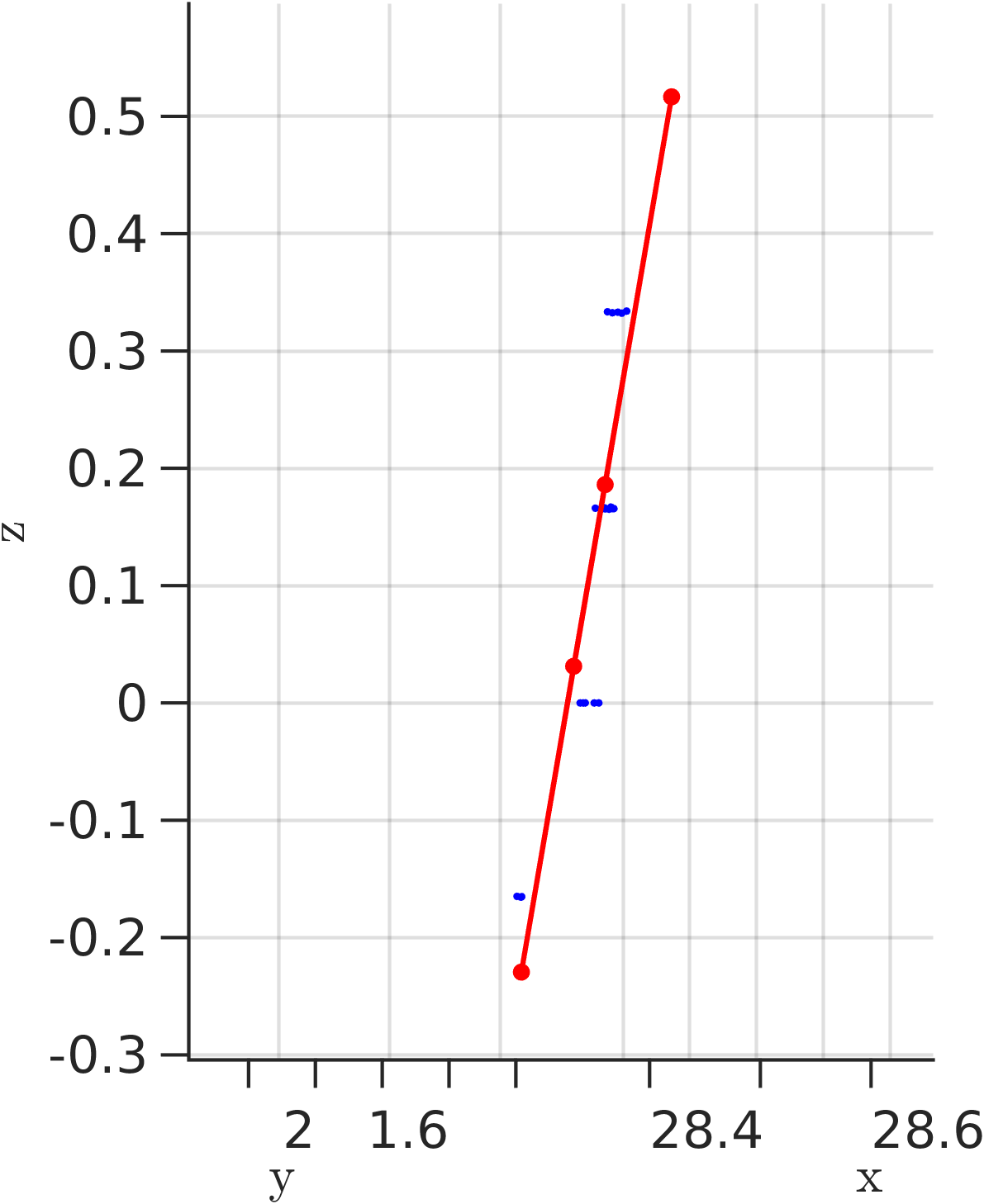}
    \end{subfigure}%
    \vspace{2pt}
    \begin{subfigure}{0.48\columnwidth}
        \centering
        \includegraphics[height=1.1\columnwidth, trim={0 0 0 0},clip]{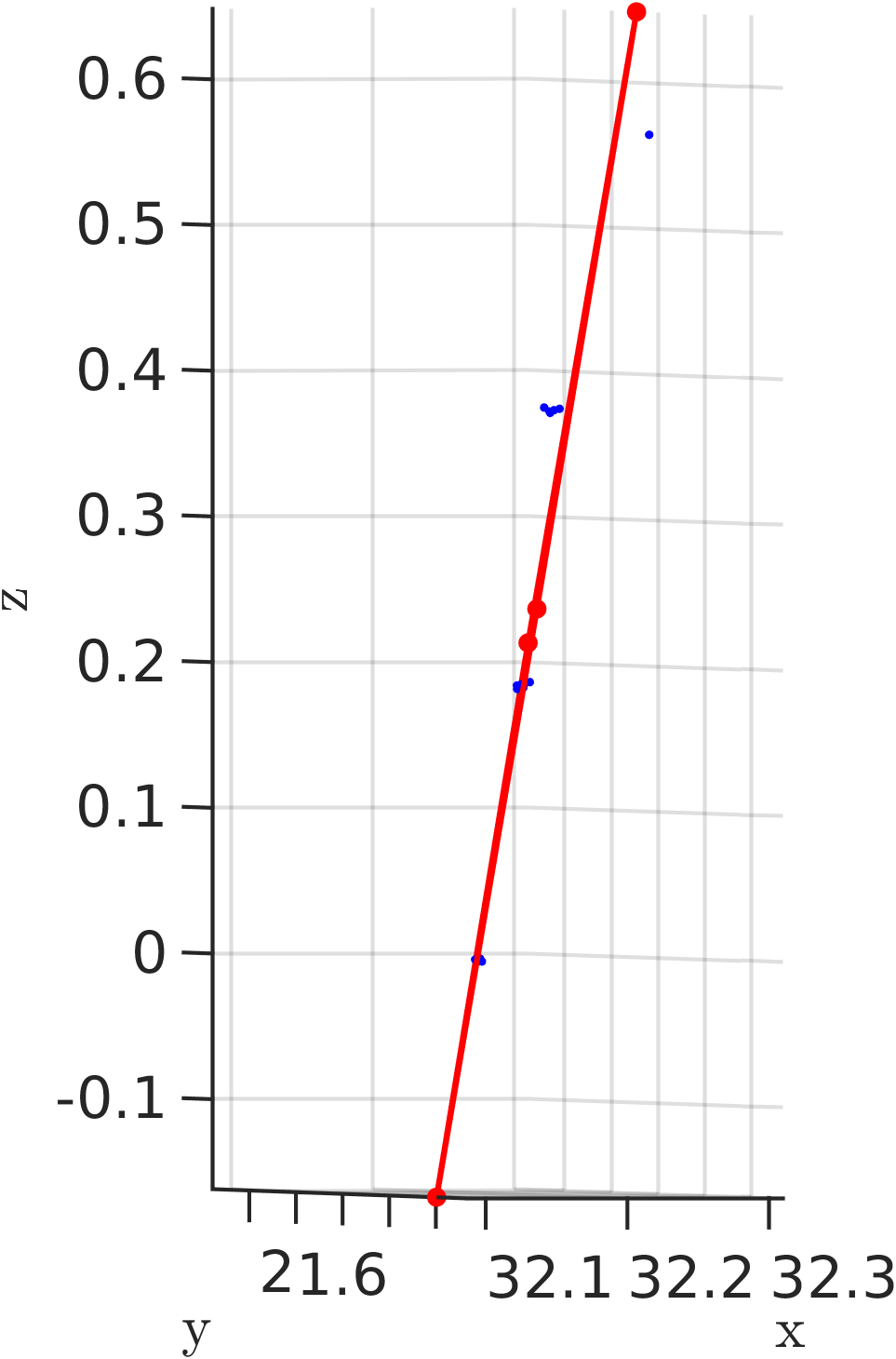}
    \end{subfigure}%
    \caption[]{Fitting results of the optimal shape at various distances (20, 24, 28,
        32 meters) in the atrium of the Ford Robotics Building at the University of
        Michigan. The blue dots are the \lidar returns on the target and the red
        frame are the fitting results. The top and bottom show the front view and a
    side view of the results, respectively.}
    \label{fig:EXPResults}%
\end{figure*}

\begin{figure*}[!t]%
    \centering
    \begin{subfigure}{0.48\columnwidth}
        \centering
        \includegraphics[height=0.7\columnwidth, trim={0 0 0 0},clip]{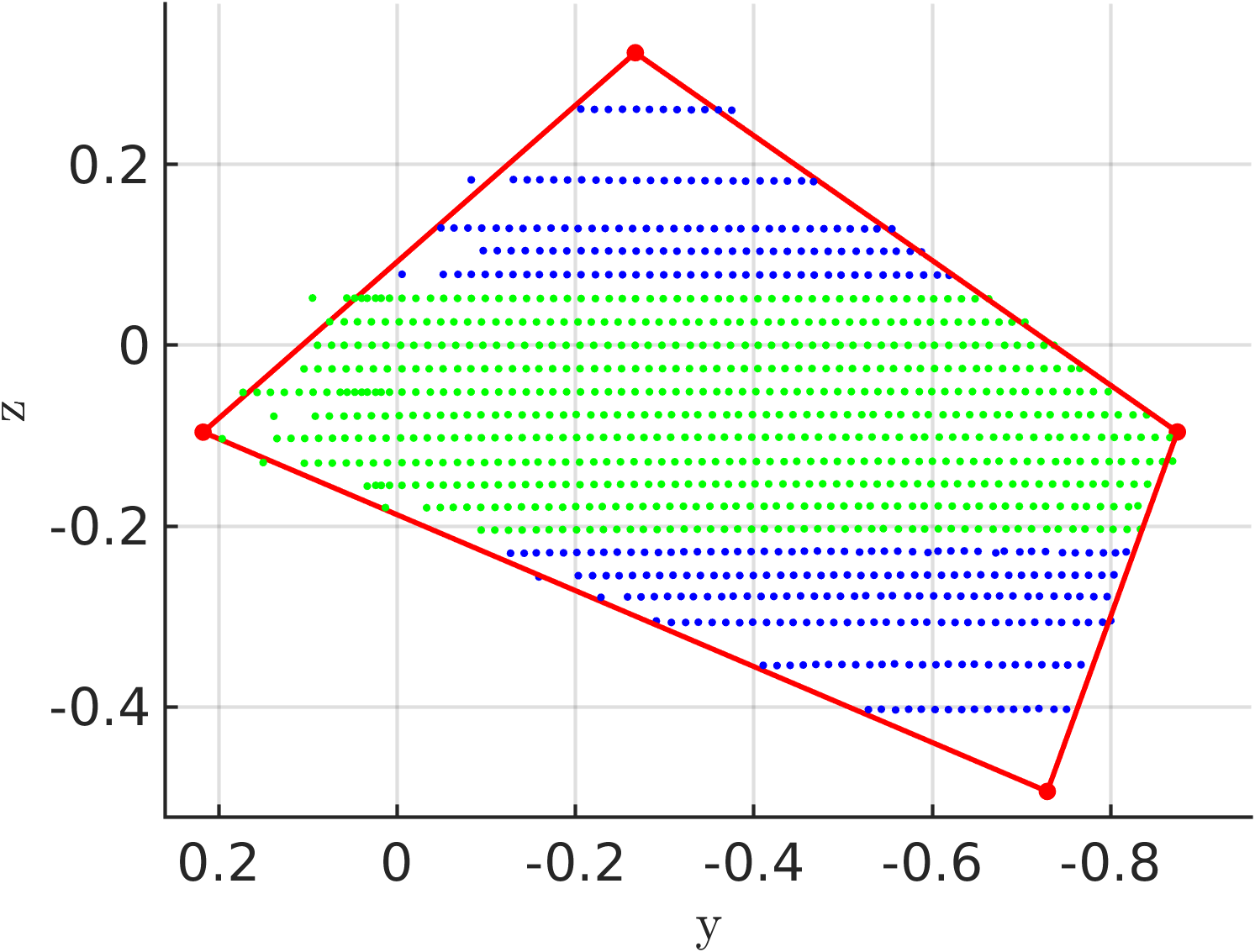}
    \end{subfigure}%
    \vspace{2pt}
    \begin{subfigure}{0.48\columnwidth}
        \centering
        \includegraphics[height=0.7\columnwidth, trim={0 0 0 0},clip]{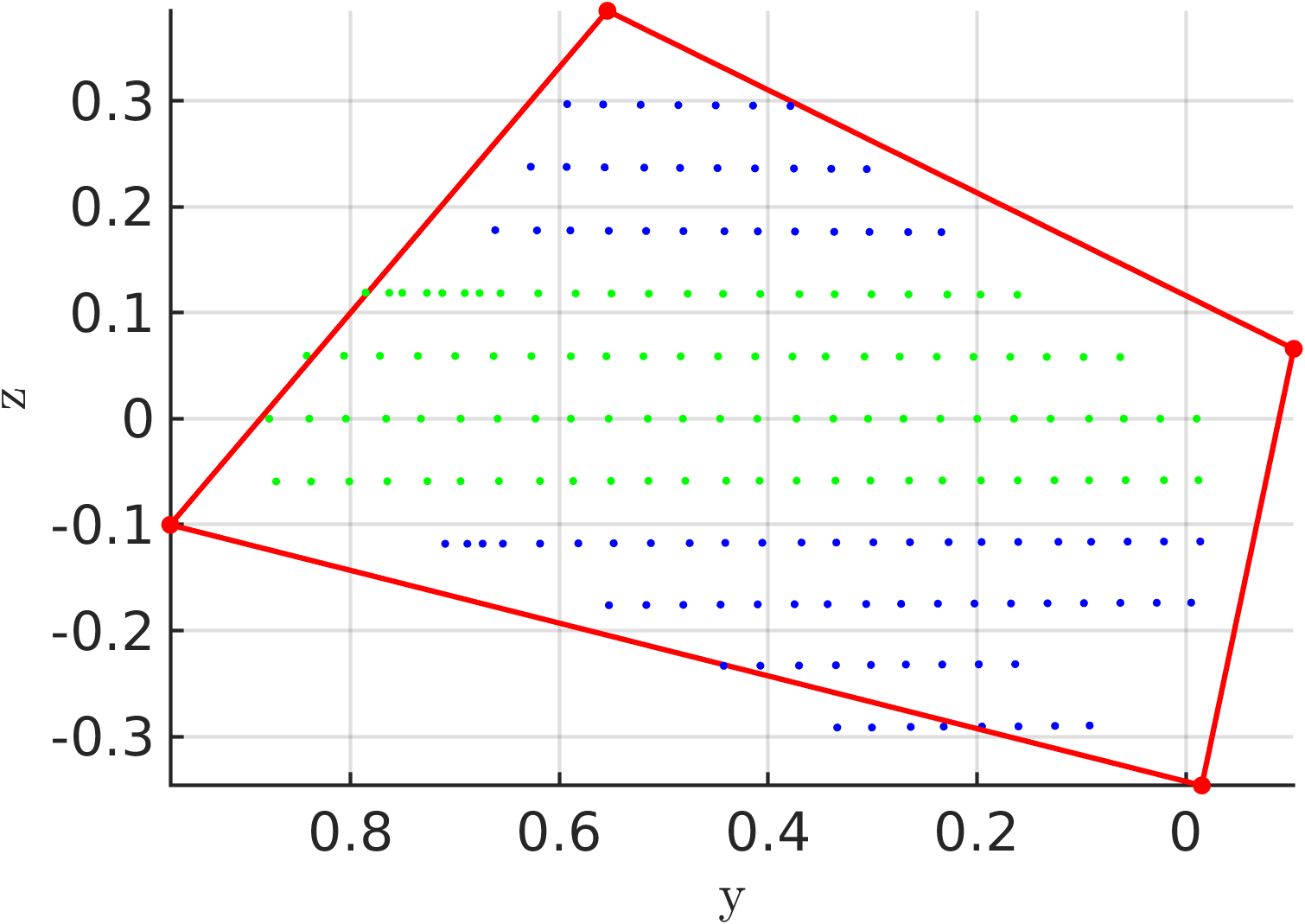}
    \end{subfigure}%
    \vspace{2pt}
    \begin{subfigure}{0.48\columnwidth}
        \centering
        \includegraphics[height=0.7\columnwidth, trim={0 0 0 0},clip]{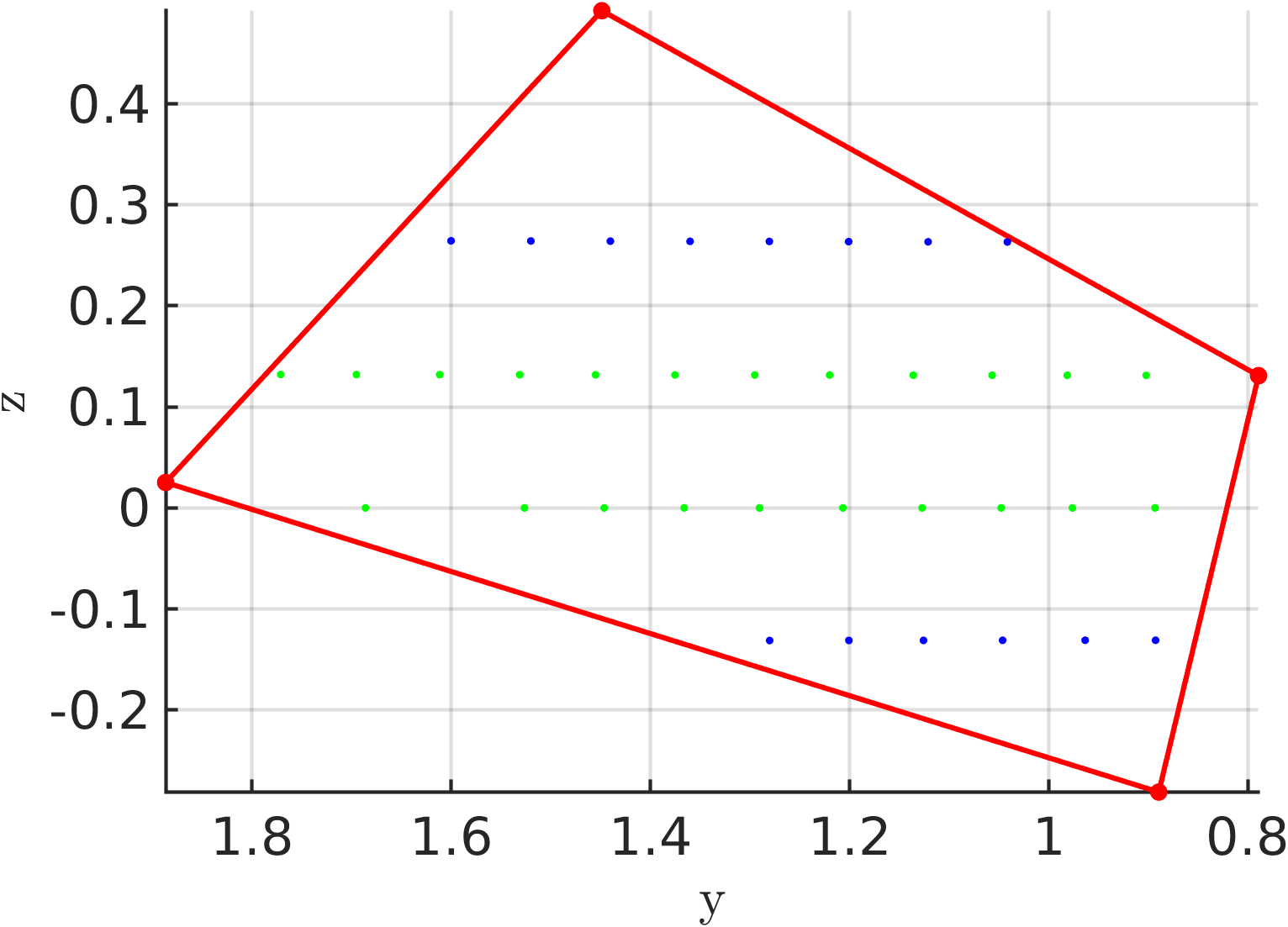}
    \end{subfigure}%
    \vspace{2pt}
    \begin{subfigure}{0.48\columnwidth}
        \centering
        \includegraphics[height=0.7\columnwidth, trim={0 0 0 0},clip]{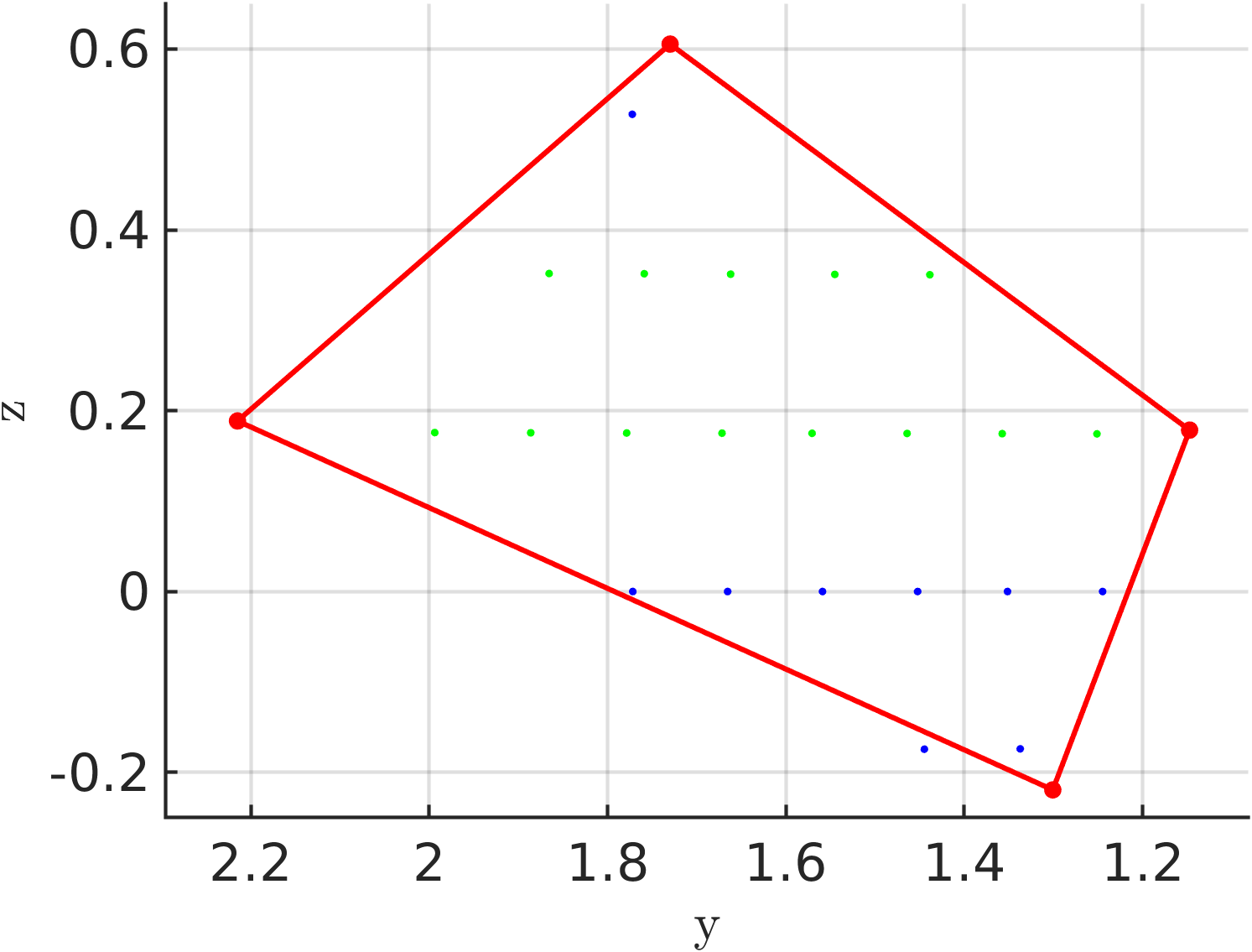}
    \end{subfigure}
    \vspace{2pt}
    \begin{subfigure}{0.48\columnwidth}
        \centering
        \includegraphics[height=1.1\columnwidth, trim={0 0 0 0},clip]{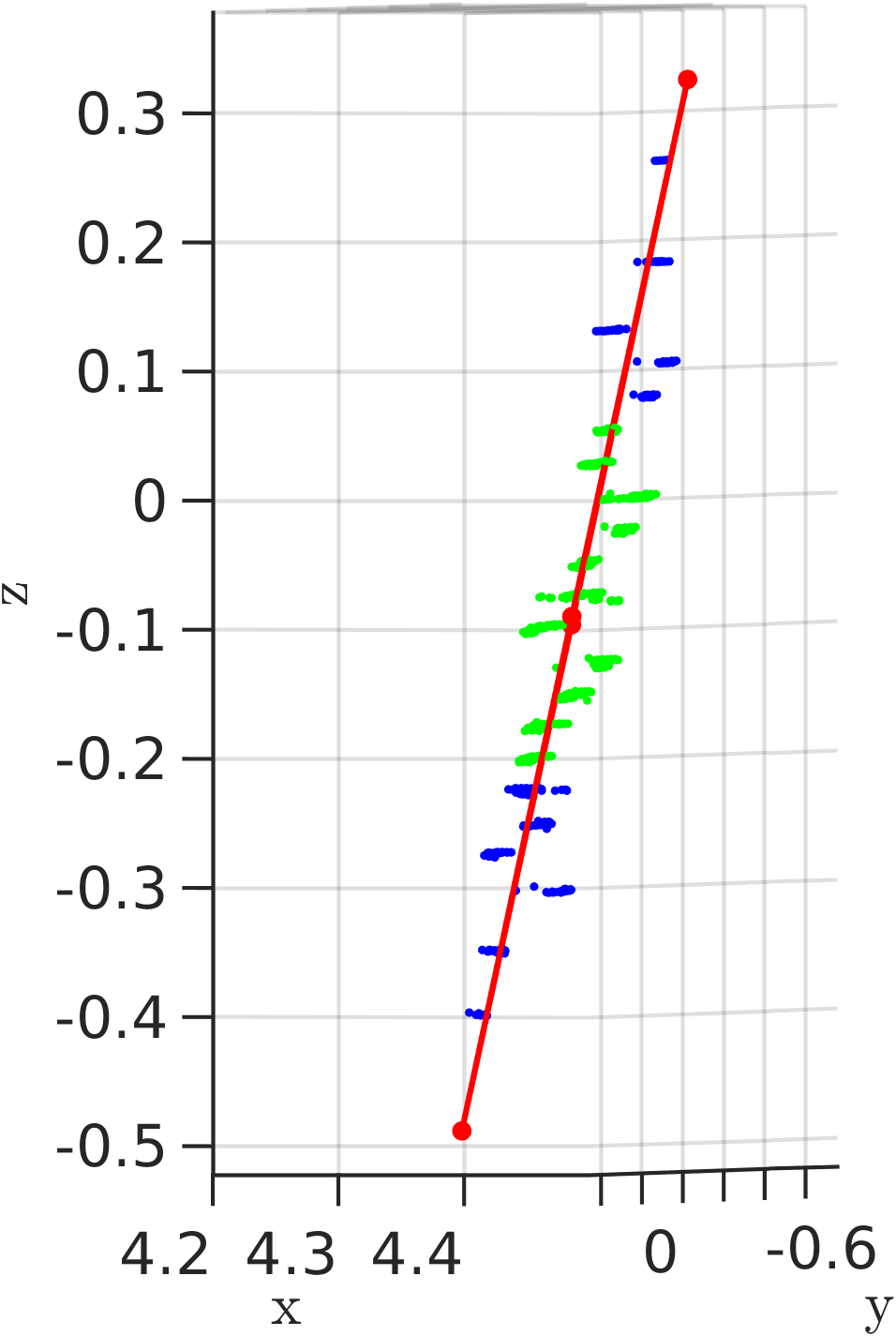}
    \end{subfigure}%
    \vspace{2pt}
    \begin{subfigure}{0.48\columnwidth}
        \centering
        \includegraphics[height=1.1\columnwidth, trim={0 0 0 0},clip]{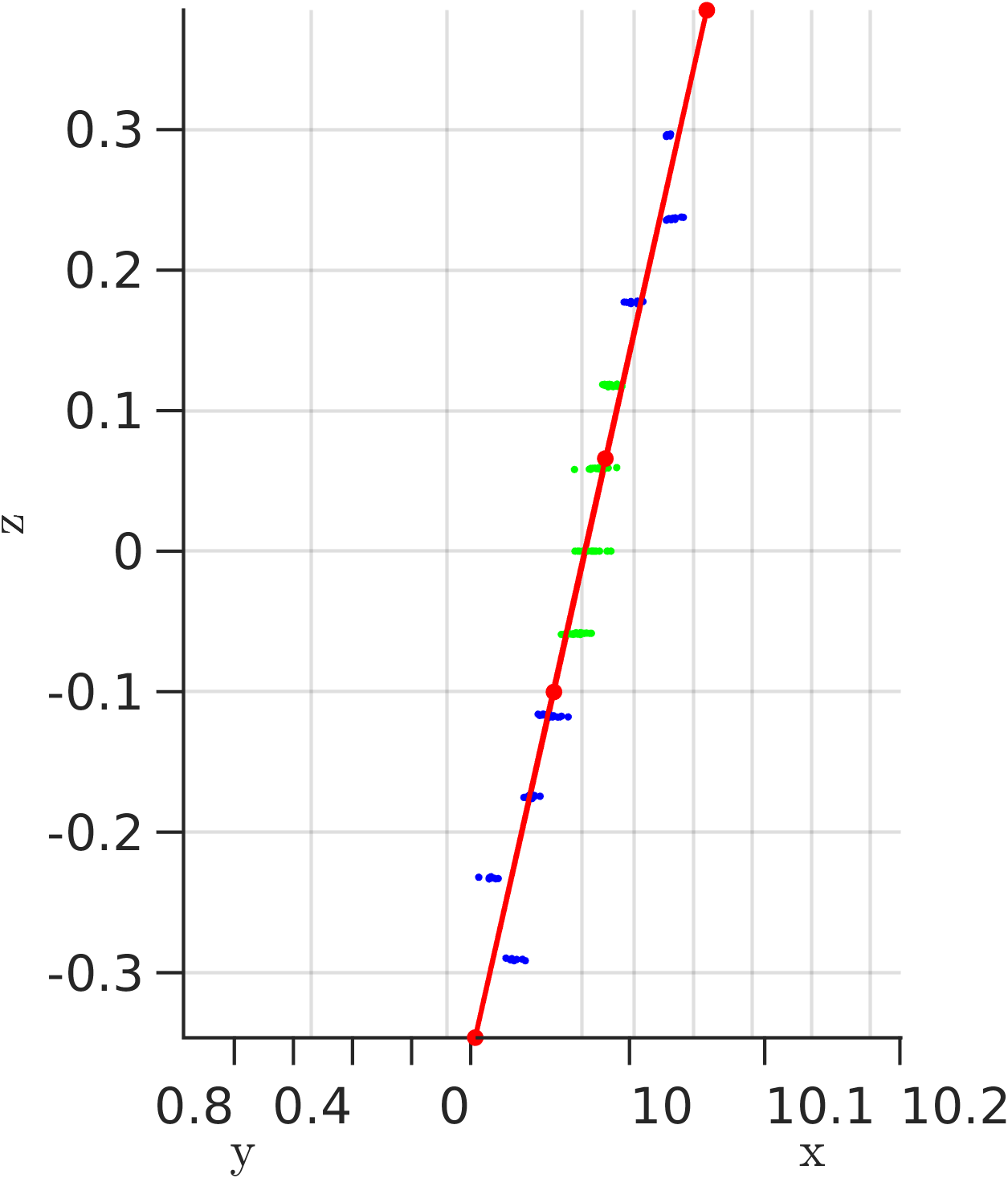}
    \end{subfigure}%
    \vspace{2pt}
    \begin{subfigure}{0.48\columnwidth}
        \centering
        \includegraphics[height=1.1\columnwidth, trim={0 0 0 0},clip]{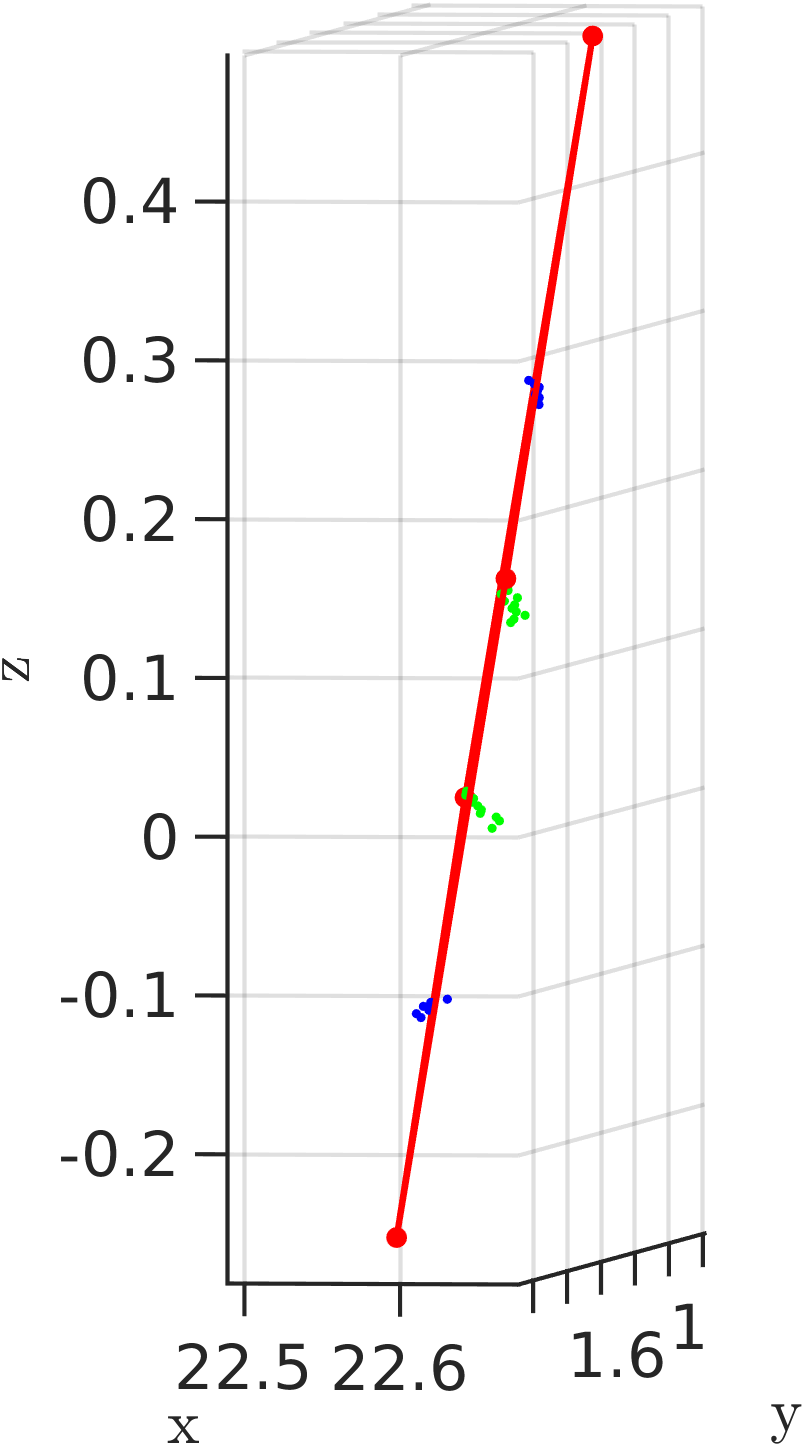}
    \end{subfigure}%
    \vspace{2pt}
    \begin{subfigure}{0.48\columnwidth}
        \centering
        \includegraphics[height=1.1\columnwidth, trim={0 0 0 0},clip]{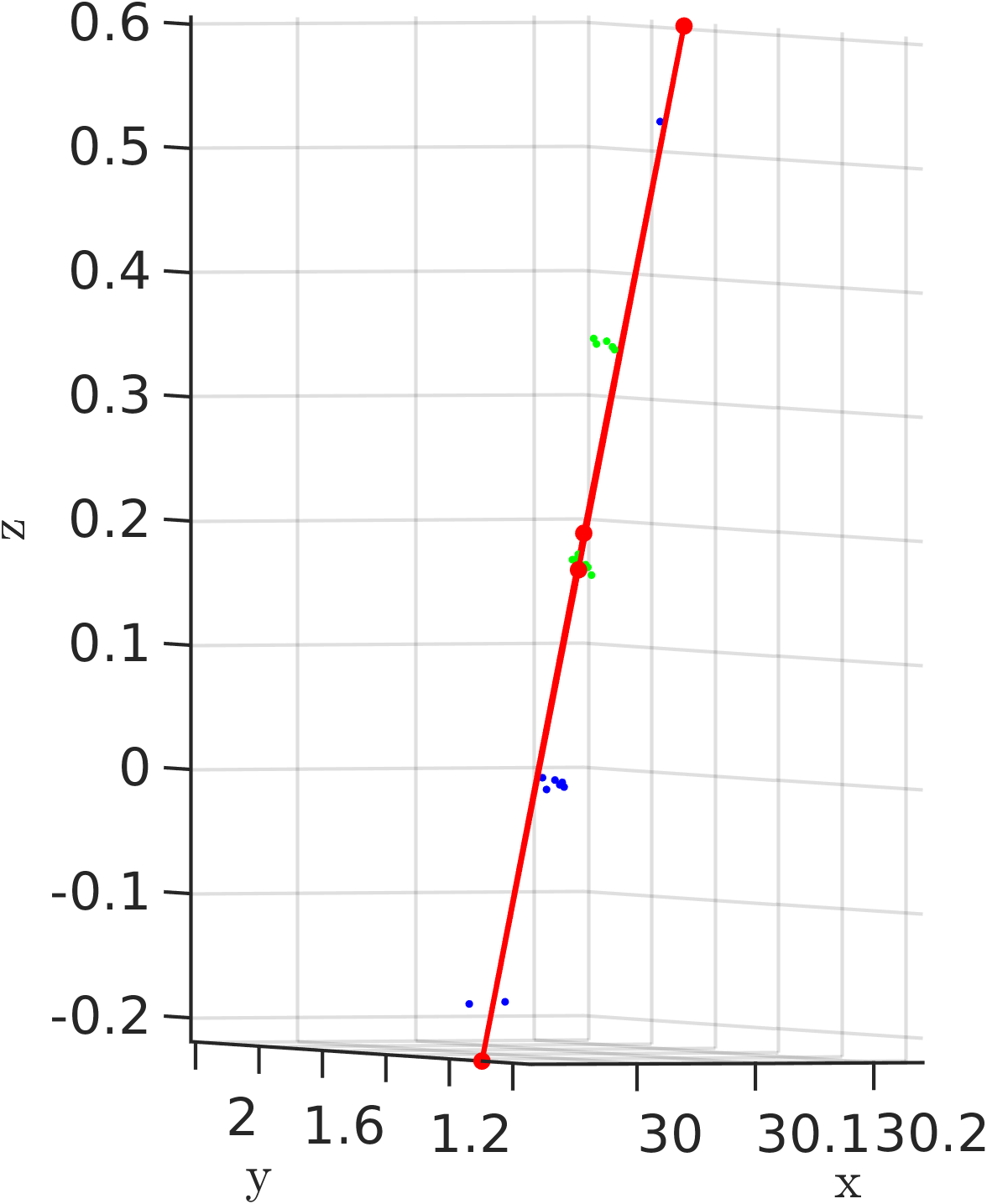}
    \end{subfigure}%
    \caption[]{Fitting results of the partially illuminated target at various distances (4, 10, 22,
        30 meters) in the atrium of the Ford Robotics Building at the University of
        Michigan. The selected distances are different from Fig.~\ref{fig:EXPResults}
        to show more results. The red frames are the fitting results. The blue dots
        are the \lidar returns on the targets while the green dots are considered
        missing. The top and bottom show the front view and a side view of the
        fitting results, respectively.}%
    \label{fig:OccludedEXPResults}%
\end{figure*}

For distances beyond 16 meters (the distance limit in M-Air), we present qualitative results from the atrium in Fig.~\ref{fig:EXPResults} to support the simulation-based analysis. The blue dots are \lidar measurements, and the red
frame is the fitting result. Figure~\ref{fig:OccludedEXPResults} illustrates a partially illuminated target (the green dots are assumed missing and only blue
dots are used for pose estimation); the resulting fitting results are the red frame.

\begin{table}[t]
\centering
\caption{Pose and vertex accuracy of the simulation results at various distances. The vertex estimation is computed by \eqref{eq:e_RMSE}. The quantization error $e_q$
is the distance between two adjacent points on the same ring. 
The translation error $e_t$ in meters and rotation error $e_r$ in degrees are
computed by \eqref{eq:PoseError}.
} 
\label{tab:SimNoiseFreeResults}
\scalebox{0.85}{
\begin{tabular}{|c|c|c|c|c|c|c|c|}
\hline
\multicolumn{8}{|c|}{\textbf{Face-on LiDAR}} \\ \hline
\begin{tabular}[c]{@{}c@{}}Distance\\ {[}m{]}\end{tabular} & \begin{tabular}[c]{@{}c@{}}No.  \\ Points\end{tabular} & \begin{tabular}[c]{@{}c@{}}$e_q$\\ {[}m{]}\end{tabular} & \begin{tabular}[c]{@{}c@{}}RMSE\\ {[}m{]}\end{tabular} & \begin{tabular}[c]{@{}c@{}}RMSE\\ {[}\%{]}\end{tabular} & \begin{tabular}[c]{@{}c@{}}$e_t$ \\ {[}m{]}\end{tabular} & \begin{tabular}[c]{@{}c@{}}$e_t$\\  {[}\%{]}\end{tabular} & \begin{tabular}[c]{@{}c@{}}$e_r$ \\ {[}deg{]}\end{tabular} \\ \hline
2.00 & 2530& 0.01 & 0.01 & 0.49 & 0.002 & 0.11 & 0.71 \\ \hline
4.00 & 1015& 0.03 & 0.03 & 0.83 & 0.005 & 0.12 & 2.45 \\ \hline
6.00 & 548 & 0.04  &0.02 & 0.40 & 0.01 & 0.12 & 1.77 \\ \hline
8.00 & 338 & 0.06  &0.02 & 0.28 & 0.01 & 0.15 & 1.58 \\ \hline
16.00 & 96  & 0.11 &0.04 & 0.22 & 0.03 & 0.21 & 0.68 \\ \hline
30.00 & 28  & 0.21 &0.03 & 0.10 & 0.03 & 0.08 & 1.09 \\ \hline
32.00 & 26  & 0.22 &0.06 & 0.17 & 0.05 & 0.17 & 1.20 \\ \hline
\multicolumn{8}{|c|}{\textbf{Extreme Angle (20, 30, 30)}} \\ \hline
\begin{tabular}[c]{@{}c@{}}Distance\\ {[}m{]}\end{tabular} & \begin{tabular}[c]{@{}c@{}}No.  \\ Points\end{tabular} & \begin{tabular}[c]{@{}c@{}}$e_q$\\ {[}m{]}\end{tabular} & \begin{tabular}[c]{@{}c@{}}RMSE\\ {[}m{]}\end{tabular} & \begin{tabular}[c]{@{}c@{}}RMSE\\ {[}\%{]}\end{tabular} & \begin{tabular}[c]{@{}c@{}}$e_t$ \\ {[}m{]}\end{tabular} & \begin{tabular}[c]{@{}c@{}}$e_t$\\  {[}\%{]}\end{tabular} & \begin{tabular}[c]{@{}c@{}}$e_r$ \\ {[}deg{]}\end{tabular} \\ \hline
1.94 & 1734& 0.01  & 0.01 & 0.62 & 0.01 & 0.37 & 0.71 \\ \hline
3.93 & 716 & 0.03 & 0.03 & 0.71 & 0.03 & 0.64 & 0.95 \\ \hline
7.93 & 258 & 0.06 & 0.04 & 0.46 & 0.03 & 0.44 & 0.96 \\ \hline
15.93 & 70 & 0.11& 0.05 & 0.31 & 0.05 & 0.29 & 1.37 \\ \hline
29.93 & 21 & 0.21& 0.12 & 0.39 & 0.09 & 0.30 & 5.65 \\ \hline
31.93 & 17  & 0.22& 0.47 & 1.46 & 0.06 & 0.18 & 34.72 \\ \hline
\end{tabular}
}
\end{table}

\begin{table}[ht]
    \centering
\caption{Pose and vertex accuracy of the experimental results. The ground truth is
    provided by a motion capture system with 33 cameras. The quantization error $e_q$
is the distance between two adjacent points on the same ring. The
    translation error $e_t$ in meters and rotation error $e_r$ in degrees are
    computed by \eqref{eq:PoseError}.
}
\label{tab:ExpResults}
\scalebox{0.81}{
\begin{tabular}{|c|c|c|c|c|c|c|c|}
\hline
\multicolumn{8}{|c|}{\textbf{Face-on LiDAR}} \\ \hline
\begin{tabular}[c]{@{}c@{}}Distance\\ {[}m{]}\end{tabular} & \begin{tabular}[c]{@{}c@{}}No.  \\ Points\end{tabular} & \begin{tabular}[c]{@{}c@{}}$e_q$\\ {[}m{]}\end{tabular} & \begin{tabular}[c]{@{}c@{}}RMSE\\ {[}m{]}\end{tabular} & \begin{tabular}[c]{@{}c@{}}RMSE\\ {[}\%{]}\end{tabular} & \begin{tabular}[c]{@{}c@{}}$e_t$ \\ {[}m{]}\end{tabular} & \begin{tabular}[c]{@{}c@{}}$e_t$\\  {[}\%{]}\end{tabular} & \begin{tabular}[c]{@{}c@{}}$e_r$ \\ {[}deg{]}\end{tabular} \\ \hline
2.1487 & 3009  & 0.015 & 0.026 & 1.193 & 0.024 & 1.133 & 0.619 \\ \hline
4.0363 & 1120  & 0.028 & 0.029 & 0.718 & 0.024 & 0.603 & 1.906 \\ \hline
6.0877 & 527  & 0.042 & 0.033 & 0.544 & 0.03 & 0.5 & 1.426 \\ \hline
7.9145 & 589  & 0.055 & 0.033 & 0.413 & 0.03 & 0.384 & 1.765 \\ \hline
10.093 & 197  & 0.07 & 0.034 & 0.334 & 0.03 & 0.294 & 1.846 \\ \hline
12.016 & 136  & 0.084 & 0.035 & 0.288 & 0.03 & 0.247 & 2.081 \\ \hline
13.987 & 192  & 0.098  & 0.03 & 0.216 & 0.028 & 0.201 & 1.737 \\ \hline
15.981 & 155  & 0.112 & 0.03 & 0.187 & 0.028 & 0.176 & 1.237 \\ \hline
17.971 & 119  & 0.125 & 0.031 & 0.173 & 0.028 & 0.156 & 1.387 \\ \hline
\multicolumn{8}{|c|}{\textbf{Extreme Angle (20, 30, 30)}} \\ \hline
\begin{tabular}[c]{@{}c@{}}Distance\\ {[}m{]}\end{tabular} & \begin{tabular}[c]{@{}c@{}}No.  \\ Points\end{tabular} & \begin{tabular}[c]{@{}c@{}}$e_q$\\ {[}m{]}\end{tabular} & \begin{tabular}[c]{@{}c@{}}RMSE\\ {[}m{]}\end{tabular} & \begin{tabular}[c]{@{}c@{}}RMSE\\ {[}\%{]}\end{tabular} & \begin{tabular}[c]{@{}c@{}}$e_t$ \\ {[}m{]}\end{tabular} & \begin{tabular}[c]{@{}c@{}}$e_t$\\  {[}\%{]}\end{tabular} & \begin{tabular}[c]{@{}c@{}}$e_r$ \\ {[}deg{]}\end{tabular} \\ \hline
2.0448& 2708& 0.014  & 0.048 & 2.364 & 0.027 & 1.304 & 4.577 \\ \hline
4.1225& 898 & 0.029  & 0.04 & 0.971 & 0.034 & 0.835 & 2.641 \\ \hline
5.966 & 504 & 0.042   & 0.033 & 0.549 & 0.029 & 0.486 & 1.826 \\ \hline
7.9452& 276 & 0.055 & 0.052 & 0.656 & 0.031 & 0.395 & 4.633 \\ \hline
9.9484& 155 & 0.069 & 0.041 & 0.412 & 0.034 & 0.345 & 3.52 \\ \hline
12.006& 105 & 0.084 & 0.06 & 0.498 & 0.036 & 0.299 & 6.064 \\ \hline
14.106& 75  & 0.098 & 0.051 & 0.364 & 0.045 & 0.318 & 2.714 \\ \hline
16.05 & 54  & 0.112   & 0.057 & 0.355 & 0.041 & 0.253 & 4.634 \\ \hline
18.087& 48  & 0.126 & 0.055 & 0.305 & 0.044 & 0.241 & 3.593 \\ \hline
\end{tabular}
}
\end{table}

\section{Conclusion}
\label{sec:OptimalShapeConclusion}
We presented the concept of optimizing target shape to enhance pose estimation for \lidar point clouds. 
We formulated the problem in terms of choosing a target shape that induces large gradients at edge points under translation and
rotation so as to mitigate the effects of quantization uncertainty associated with point cloud
sparseness. For additional robustness, the cost function or score for a candidate target was defined to be the minimum score under a set of relative rotations of the edge points; this had the effect of breaking symmetry in the candidate target, which also removes pose ambiguity. 

For a given target, we
used the target's geometry to jointly estimate the target's
pose and its vertices. The estimation problem was formulated so
that an existing semi-definite programming global solver could be modified to
globally and efficiently compute the pose and vertices of the target. A \lidar 
simulator generated synthetic ground truth of the target pose and vertices. We validated that the combination of the
optimal shape with the global solver achieved centimeter error in vertex estimation,
centimeter error in translation, and a few degrees off in rotation in pose estimation
when a partially illuminated target was placed 30 meters from the \lidar.
In experiments, when compared to
ground truth data collected by a motion capture system with 33 cameras, we achieved results similar to those of the simulations.

In the future, we shall establish a system to automatically detect the shape in both
images and \lidar point clouds. If a dataset has been collected and labeled,
automatic detection using deep-learning architectures is also an exciting future
direction. Currently, the proposed algorithm assumes the point cloud has been motion
compensated; how to adopt motion distortion into the algorithm is another direction
for future work. Applying it as a fiducial marker system or as an automatic
calibration system also offers another interesting area for further research. Furthermore, applying the proposed algorithm to 3D target shape
fitting and generating shapes with more sides provide interesting research
directions.


\section*{Acknowledgment}
\small{ 
Toyota Research Institute provided funds to support this
work. Funding for J. Grizzle was in part provided by NSF
Award No. 1808051 and 2118818. The first author thanks Wonhui Kim for
useful conversations.}


\bibliographystyle{bib/IEEEtran}
\bibliography{DefinesBib/bib_all/strings-abrv,DefinesBib/bib_all/ieee-abrv,DefinesBib/bib_all/BipedLab.bib,DefinesBib/bib_all/Books.bib,DefinesBib/bib_all/Bruce.bib,DefinesBib/bib_all/ComputerVision.bib,DefinesBib/bib_all/ComputerVisionNN.bib,DefinesBib/bib_all/IntrinsicCal.bib,DefinesBib/bib_all/L2C.bib,DefinesBib/bib_all/LibsNSoftwares.bib,DefinesBib/bib_all/ML.bib,DefinesBib/bib_all/OptimizationNMath.bib,DefinesBib/bib_all/Other.bib,DefinesBib/bib_all/StateEstimationSLAM.bib,DefinesBib/bib_all/MotionPlanning.bib,DefinesBib/bib_all/Mapping.bib,DefinesBib/bib_all/TrajectoriesOptimization.bib,DefinesBib/bib_all/Controls.bib}

\begin{appendices}
\end{appendices}
\end{document}